\documentclass[a4paper]{cas-sc}

\usepackage[authoryear,longnamesfirst]{natbib}

\def\tsc#1{\csdef{#1}{\textsc{\lowercase{#1}}\xspace}}
\tsc{WGM}
\tsc{QE}
\tsc{EP}
\tsc{PMS}
\tsc{BEC}
\tsc{DE}

\usepackage{amsmath,amsfonts}

\usepackage{array}
\usepackage{subcaption}
\usepackage{textcomp}
\usepackage{url}
\usepackage{verbatim}
\usepackage{graphicx}
\usepackage{graphicx}

\usepackage{bm}

\usepackage{footnote}

\usepackage{amssymb}
\usepackage{booktabs}

\usepackage{float}

\usepackage{dsfont}
\usepackage{subcaption}

\usepackage[ruled,vlined]{algorithm2e} 
\usepackage[noend]{algpseudocode} 
\usepackage{tikz}
\usepackage{caption}

\usepackage{amssymb}  

\usepackage{caption}
\usepackage{graphicx}

\usepackage{nicefrac}

\usepackage{enumitem}
\usepackage{colortbl}
\usepackage{multirow}

\usepackage[export]{adjustbox}

\usepackage{pifont}

\definecolor{sh_gray}{rgb}{0.84,0.84,0.84}
\definecolor{sh_gray2}{rgb}{1,0.89,0.75}
\definecolor{color3}{rgb}{0.95,0.95,0.95}
\definecolor{color4}{rgb}{0.94,0.94,1}
\definecolor{color5}{rgb}{1,0.96,0.88}
\definecolor{cvprblue}{rgb}{0.21,0.49,0.74}

\usepackage{colortbl} 
\usepackage{xcolor} 

\definecolor{best}{rgb}{1, 0.88, 0.88} 
\definecolor{secondbest}{rgb}{0.8, 0.88, 1} 
\definecolor{best}{rgb}{1, 0.88, 0.88} 
\definecolor{secondbest}{rgb}{0.8, 0.88, 1} 

\newlength \g
\begin{document}
\let\WriteBookmarks\relax
\def\floatpagepagefraction{1}
\def\textpagefraction{.001}
\shorttitle{Leveraging social media news}
\shortauthors{Han Hu et~al.}

\title [mode = title]{VAMamba: An Efficient Visual Adaptive Mamba for Image Restoration}                      



\author[1]{Han Hu}[type=editor]


\affiliation[1]{organization={Shandong Normal University},
                postcode={250358}, 
                state={Jinan},
                country={China}}

\author[2]{Zhuoran Zheng}[style=chinese]
\cormark[1]

\author[3]{Liang Li}[style=chinese]


\affiliation[2]{organization={Sun Yat-sen University},
                postcode={510080}, 
                state={Guangzhou},
                country={China}}

\author[1]{Chen Lyu}
\cormark[1]

\affiliation[3]{organization={School of Transportation and Logistics Engineering},
                postcode={250357}, 
                state={Jinan}, 
                country={China}}

\cortext[cor1]{Corresponding author}


\begin{abstract}
Recent Mamba-based image restoration methods have achieved promising results but remain limited by fixed scanning patterns and inefficient feature utilization. Conventional Mamba architectures rely on predetermined paths that cannot adapt to diverse degradations, constraining both restoration performance and computational efficiency.
To overcome these limitations, we propose VAMamba, a Visual Adaptive Mamba framework with two key innovations. First, QCLAM (Queue-based Cache Low-rank Adaptive Memory) enhances feature learning through a FIFO cache that stores historical representations. Similarity between current LoRA-adapted and cached features guides intelligent fusion, enabling dynamic reuse while effectively controlling memory growth.
Second, GPS-SS2D (Greedy Path Scan SS2D) introduces adaptive scanning. A Vision Transformer generates score maps to estimate pixel importance, and a greedy strategy determines optimal forward and backward scanning paths. These learned trajectories replace rigid patterns, enabling SS2D to perform targeted feature extraction.
The integration of QCLAM and GPS-SS2D allows VAMamba to adaptively focus on degraded regions while maintaining high computational efficiency. Extensive experiments across diverse restoration tasks demonstrate that VAMamba consistently outperforms existing approaches in both restoration quality and efficiency, establishing new benchmarks for adaptive image restoration. Our code is available at \href{https://github.com/WaterHQH/VAMamba}{\textcolor{red}{https://github.com/WaterHQH/VAMamba}}.

\end{abstract}



\begin{keywords}
Image Restoration \sep State-Space Models \sep Low-Rank Adaptation \sep Greedy Path Scan
\end{keywords}

\maketitle

\section{Introduction}

Image restoration is a fundamental problem in computer vision, encompassing a range of critical applications including denoising \cite{zhang2017beyond},\cite{guo2019toward},\cite{guo2015efficient-TCSVT}, deblurring \cite{kupyn2018deblurgan}\cite{nah2017deep}, low-light enhancement \cite{chen2018learning},\cite{zhang2019kindling},\cite{he2023low-TCSVT}, dehazing \cite{ren2016single}\cite{li2017aod}, rain removal \cite{zhang2018density},\cite{cheng2024fdce-TCSVT},\cite{zhang2023underwater-TCSVT}, and defocus blur recovery \cite{karaali2017edge},\cite{lee2019deep}. The objective is to recover high-quality images from degraded or low-quality inputs, which are often plagued by complex artifacts such as noise, blurring, low-light conditions, or atmospheric disturbances like haze and rain. Despite advances in deep learning-based image restoration methods, fundamental challenges remain, especially in balancing high-fidelity restoration with computational efficiency across diverse degradation types.

\begin{figure}[!t]
\captionsetup{justification=justified,singlelinecheck=false}
\centering
\begin{subfigure}[t]{0.48\linewidth}
\captionsetup{width=0.48\linewidth}
  \includegraphics[width=\linewidth]{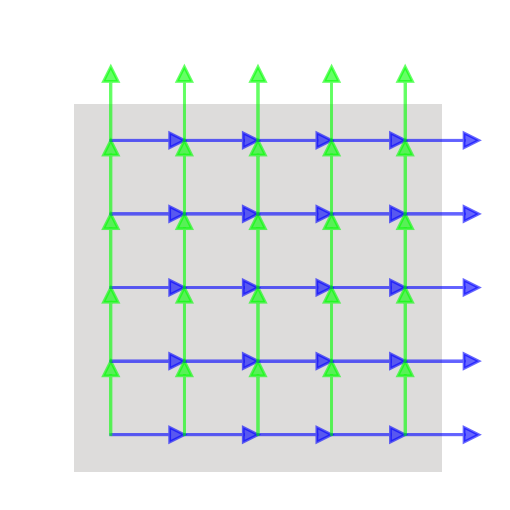}
  \caption{Fixed scanning path}
\end{subfigure}\hfill
\begin{subfigure}[t]{0.48\linewidth}
\captionsetup{width=0.48\linewidth}
  \includegraphics[width=\linewidth]{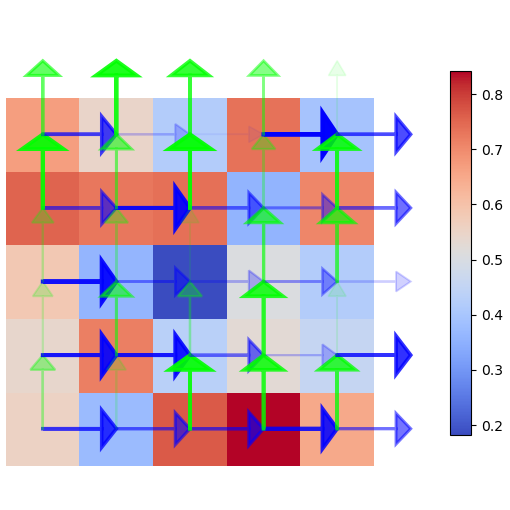}
  \caption{Adaptive scanning path}
\end{subfigure}
\caption{Comparison between fixed and adaptive scanning paths in SS2D models. (a) The fixed scanning path lacks adaptability, limiting performance across diverse degradation types. (b) Our adaptive scanning path dynamically adjusts to the input, enhancing restoration accuracy and generalization.}
\label{fig:scan_path_comparison}
\vspace{-0.7em}
\end{figure}


\par While CNN-based methods have demonstrated success in capturing local patterns, they inherently struggle to model long-range dependencies due to limited receptive fields \cite{zhang2017beyond, guo2019toward}. Transformer architectures, on the other hand, address this limitation through self-attention mechanisms \cite{chen2021pre, liang2021swinir}. However, the quadratic complexity of attention mechanisms increases computation significantly, especially for high-resolution images, which restricts their scalability in real-time applications. Furthermore, both CNN and Transformer models are often designed with task-specific architectures\cite{zhao2023tufusion-TCSVT}, limiting their adaptability across various degradation scenarios.

\par Recently, State Space Models (SSMs), particularly those based on the Mamba framework, have emerged as promising alternatives for image restoration due to their efficiency in modeling long-range dependencies with linear complexity \cite{gu2023mamba, guo2024mambair}. Mamba-based models such as MambaIR \cite{guo2024mambair}, VMambaIR \cite{shi2024vmambair}, and MambaCSR \cite{ren2024mamabcsr} leverage the SS2D mechanism to efficiently capture both spatial and channel dependencies with linear computational complexity. These models offer significant computational advantages over Transformer architectures while maintaining effective long-range dependency modeling capabilities. However, current Mamba-based approaches rely on predetermined fixed scanning paths, including horizontal, vertical, and various backward scanning patterns.

\par The reliance on fixed scanning strategies creates fundamental limitations that constrain restoration performance. As illustrated in Fig.~\ref{fig:scan_path_comparison}(a), fixed-path methods lack the flexibility to dynamically adjust attention allocation based on local image structure and specific degradation characteristics. This rigidity becomes particularly problematic when processing heterogeneous degradation patterns that require adaptive feature extraction strategies. Moreover, fixed scanning paths in SS2D models often result in redundant computations and inefficient memory utilization, reducing overall model adaptability across various degradation types.

\par To address these critical limitations, we propose VAMamba, a Visual Adaptive Mamba architecture that introduces two key innovations for enhanced image restoration. Our approach extends the SS2D mechanism through QCLAM and GPS-SS2D components. The QCLAM module combines dynamic caching with low-rank adaptation using a FIFO queue-based similarity matching mechanism. Meanwhile, GPS-SS2D employs Vision Transformer-generated score maps and greedy path planning to create adaptive scanning trajectories that respond to image content and degradation patterns(as shown in Fig.~\ref{fig:scan_path_comparison}(b)).

\par The main contributions of this work are summarized below:
\begin{itemize}
    \item we introduce QCLAM, a novel module that integrates queue-based caching with LoRA adaptation to achieve parameter-efficient learning while optimizing memory usage through intelligent feature reuse.
    
    \item we propose GPS-SS2D, an adaptive scanning strategy that dynamically generates optimal scanning paths based on image content analysis, enabling targeted feature extraction according to degradation characteristics.
    
    \item we demonstrate through extensive experiments that VAMamba achieves superior performance across multiple restoration tasks, establishing new benchmarks for both restoration quality and computational efficiency.
\end{itemize}

\section{Related Work}

\subsection{Advances in Image Restoration Methods}
Image restoration has been extensively explored with convolutional neural networks (CNNs) and Transformer-based architectures. CNNs, such as DnCNN \cite{zhang2017beyond} and RIDNet \cite{anwar2019real}, have achieved notable results by capturing local features effectively. However, they are inherently limited in modeling long-range dependencies, which are crucial for handling complex degradation. Transformer-based methods, including IPT \cite{chen2021pre} and SwinIR \cite{liang2021swinir}, address this limitation with self-attention mechanisms, allowing them to capture global dependencies. Yet, the quadratic complexity of attention poses challenges for real-time high-resolution processing \cite{zamir2022restormer}. Although hybrid approaches like Restormer attempt to balance attention and efficiency, the adaptability of these models to diverse degradation types remains limited. Diffusion Models (DMs) \cite{wang2023diffusion} have also been explored, but their high computational cost limits practicality. Therefore, a more adaptive and efficient solution is essential for versatile image restoration.

\subsection{State Space Models in Image Restoration}
SSMs have gained significant attention in image restoration for their ability to efficiently model long-range dependencies with linear complexity, making them well-suited for high-dimensional tasks like image restoration \cite{gu2023mamba}. Unlike Transformers, which suffer from high computational costs due to the quadratic complexity of self-attention, SSMs can achieve comparable results with reduced computational overhead, positioning them as an efficient alternative for real-time applications.

Mamba-based models, such as MambaIR \cite{guo2024mambair} and its variant VmambaIR \cite{shi2024vmambair}, incorporate the SS2D mechanism to handle spatial and channel dependencies. While MambaIR introduced a fixed scanning mechanism to capture dependencies, VmambaIR improved upon this by adding the Omni Selective Scan (OSS) mechanism, expanding the scanning to cover multiple directions and enhancing spatial-channel interactions. However, both methods rely on predefined, fixed scanning paths, limiting their adaptability and effectiveness when applied to complex, heterogeneous degradation patterns often found in real-world data. Fixed paths can lead to redundant feature extraction and an inability to dynamically adjust based on local degradation characteristics, making these models less flexible in scenarios requiring finer adaptivity.

In contrast, our approach introduces an adaptive scanning strategy within the SS2D framework, enabling dynamic adjustment of feature extraction based on input characteristics (see Fig.~\ref{fig:scan_path_comparison}(b)). This enhancement allows our model to focus selectively on more informative regions without being constrained by fixed paths, thus overcoming the adaptability challenges inherent in previous SSM-based methods.

\section{Method}
\subsection{Network Architecture}

\begin{figure*}[!htbp]
    \centering
    \includegraphics[width=1\linewidth]{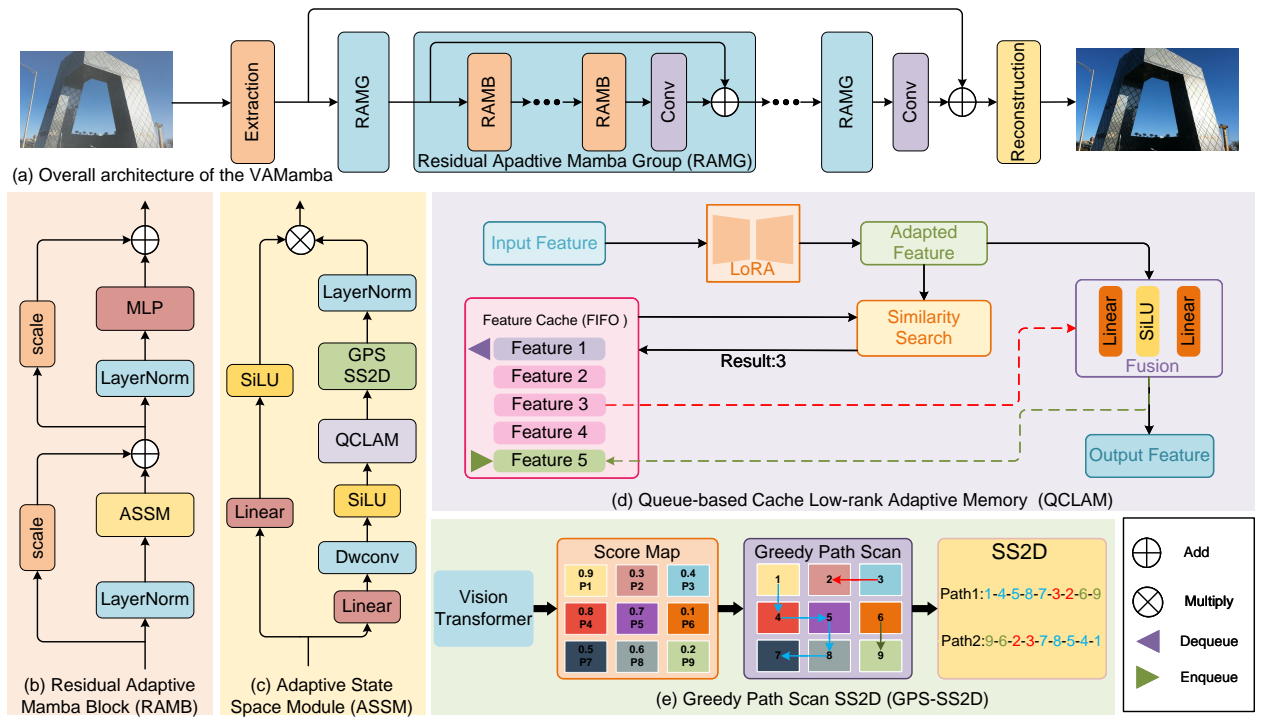}
   \caption{The overall network architecture of our VAMamba, including (a) the complete processing pipeline, (b) Residual Adaptive Mamba Block (RAMB), (c) Adaptive State Space Module (ASSM), (d) Queue-based Cache Low-rank Adaptive Memory (QCLAM), and (e) Greedy Path Scan SS2D (GPS-SS2D).}
    \label{fig:framework}
\end{figure*}

As shown in Fig.~\ref{fig:framework}, our VAMamba mainly consists of three parts: shallow feature extraction,deep feature extraction and image reconstruction. Given a low quality (LQ)  input ${I_{LQ}}\in\mathbb{R}^{H\times W\times 3}$, we extract the shallow feature from ${I_{LQ}}$ with a 3×3 convolutional layer:
\begin{equation}
    F_s = \text{Conv}_{3\times3}(I_{{LQ}}),
\end{equation}

 where $ {F_{s}}\in\mathbb{R}^{H\times W\times C}$ represents the extracted shallow features that capture initial spatial information. Subsequently, we process these shallow features through a sequence of Residual Adaptive Mamba Groups (RAMGs) to extract hierarchical deep features. Let $\mathcal{B}_{i}$ denote the transformation of the $i$-th RAMG, where the intermediate features are computed recursively as:

\begin{equation}
\centering
F_i = \mathcal{B}_i(F_{i-1}), \quad i = 1, 2, \ldots, K
\end{equation}

where $F_0=F_s$ and $K$ represents the total number of RAMG blocks. Each RAMG contains multiple Residual Adaptive Mamba Blocks (RAMBs) that incorporate our proposed QCLAM and GPS-SS2D mechanisms for adaptive feature learning.

To enhance feature aggregation and introduce beneficial inductive bias, we apply a 3×3 convolutional layer to the output of the final RAMG:
\begin{equation}
F_d = f_{conv}(F_K),
\end{equation}

where $f_{conv}(\cdot)$ is the final convolutional transformation layer.

For tasks such as image denoising or deblurring, which do not involve upsampling, we adopt a residual learning approach. Rather than directly generating HQ images from LQ inputs, we estimate the residual between the two. Specifically, the final output is formulated as:
\begin{equation}
I_{HQ} = f_{rec}(f_{conv}(F_d)) + F_s,
\end{equation}

where  $f_{\text{rec}}(\cdot)$ denotes the reconstruction function and $f_{conv}(F_d)$ is the deep feature obtained from the feature extraction module. This formulation enables precise removal of noise and artifacts by focusing on residual detail recovery.

\subsection{Residual Adaptive Mamba Block }
The Residual Adaptive Mamba Block serves as the fundamental building unit within each RAMG, designed to integrate adaptive state space modeling with efficient residual learning. As illustrated in Fig.~\ref{fig:framework}(b), RAMB processes input features through two sequential stages with learnable scaling factors.

Given input features $X$, the first stage applies layer normalization followed by the Adaptive State Space Module with scaling:

\begin{equation}X_{stage1}=\alpha\cdot\mathrm{ASSM}(\mathrm{LayerNorm}(X)).\end{equation}

The second stage takes the output from the first stage, applies layer normalization and MLP processing with scaling:

\begin{equation}X_{stage2}=\beta\cdot\mathrm{MLP}(\mathrm{LayerNorm}(X_{stage1})).\end{equation}

The final output combines the original input with both processed stages through residual connections:

\begin{equation}X_{out}=X_{stage2}.\end{equation}

This sequential two-stage design enables RAMB to first capture adaptive state space dependencies, then apply traditional feature transformation, while maintaining gradient flow through residual connections and learnable scaling factors.

\subsection{Adaptive State Space Module}
The Adaptive State Space Module represents the core innovation that enables content-aware feature processing through our proposed adaptive mechanisms. As shown in Fig.~\ref{fig:framework}(c), ASSM processes features through two sequential stages before final integration.

Given input features $X$, the first branch processes the features through a sequence of  modules:
Linear$\rightarrow$Dwconv \\ $\rightarrow$SiLU$\rightarrow$QCLAM$\rightarrow$GPS-SS2D$\rightarrow$LayerNorm, resulting in the final output $X_{branch1}$

The second branch applies SiLU activation  and layer normalization:

\begin{equation}X_{branch2}=\mathrm{LayerNorm}(\mathrm{SiLU}(X)).\end{equation}

The final ASSM output combines both branches through element-wise multiplication:

\begin{equation}X_{assm}=X_{branch1}\odot X_{branch2}.\end{equation}

This dual-branch design enables ASSM to simultaneously leverage local spatial relationships and parameter-efficient adaptation through the first branch, while capturing content-aware global dependencies through the second branch. The multiplicative fusion allows for effective feature modulation, where each branch can gate and enhance information from the other, maximizing both computational efficiency and restoration quality.











\subsection{Queue-based Cache Low-rank Adaptive Memory}

The QCLAM module addresses the challenge of efficient feature adaptation and memory utilization through a novel combination of Low-Rank Adaptation (LoRA) and intelligent feature caching. As depicted in Fig.~\ref{fig:framework}(d), QCLAM employs a FIFO queue-based cache system with similarity-guided feature fusion to enhance learning efficiency while reducing computational overhead.
Given input features $X_{input}$, QCLAM first applies LoRA adaptation to generate parameter-efficient feature transformations:

\begin{equation}
X_{adapted}=\mathrm{LoRA}(X_{input})=X_{input}+\Delta W\cdot X_{input},
\end{equation}

where $\Delta W=W_{down}\cdot W_{up}$ represents the low-rank weight update with $W_{down}\in\mathbb{R}^{C\times r}$ and $W_{up}\in\mathbb{R}^{r\times C}$, and $r\ll C$ is the rank parameter.

The adapted features are then compared against the feature cache $\mathcal{C}=\{F_1,F_2,\ldots,F_N\}$ using cosine similarity:

\begin{equation}s_i=\frac{X_{adapted}\cdot F_i}{||X_{adapted}||\cdot||F_i||}, i=1,2,\ldots,N\end{equation}

The most similar cached feature  $F_{bset}$ is selected based on the highest similarity score:

\begin{equation}k=\arg\max s_i, F_{best}=F_k\end{equation}

Feature fusion is performed through a learned linear combination weighted by the similarity score:

\begin{align}
X_{fused} &= \mathrm{Fusion}(X_{adapted}, F_{best}) \nonumber \\
          &= \gamma \cdot X_{adapted} + (1 - \gamma) \cdot F_{best}.
\end{align}

where $\gamma$ is computed based on the similarity score to balance between current and cached features.

The fused feature is then enqueued into the cache while the oldest feature is dequeued to maintain the FIFO constraint:

\begin{equation}\mathcal{C}\leftarrow\mathrm{Enqueue}(X_{fused},\mathrm{Dequeue}(\mathcal{C}))\end{equation},

where $\mathcal{C}$ represents a queue storing features, $X_{fused}$ is the newly fused feature, 
$\mathrm{Dequeue}(\mathcal{C}))$ removes and returns the oldest feature from the queue,  and $\mathrm{Enqueue}(X_{fused}$ inserts the new feature while updating the queue.

\subsection{Greedy Path Scan SS2D}

The GPS-SS2D mechanism revolutionizes traditional state space scanning by introducing content-aware path planning that adapts to image degradation patterns through intelligent patch-level analysis. As illustrated in Fig.~\ref{fig:framework}(e), GPS-SS2D employs a Vision Transformer to generate patch-wise importance scores, followed by greedy path planning to determine optimal scanning trajectories for the SS2D operation.

Given input features $X\in\mathbb{R}^{H\times W\times C}$, GPS-SS2D first applies patch partitioning through a Vision Transformer. The ViT automatically divides the input into $n$ non-overlapping patches and generates importance scores for each patch, forming a score map $S$:

\begin{equation}S=\mathrm{ViT}(X), S\in\mathbb{R}^{\sqrt{n}\times\sqrt{n}},\end{equation}

where each position $S(i,j)$ in $S$ directly corresponds to the importance score of one patch. The score for the $k$-th patch is: $score_k=S_{patch_k}$

To ensure proper probability distribution for path planning, we apply softmax normalization to all patch scores:

\begin{equation}P_k=\frac{\exp(score_k)}{\sum_{j=1}^n\exp(score_j)}\end{equation}

The greedy path planning algorithm then generates optimal scanning trajectories based on these normalized patch scores  Algorithm~\ref{alg:gps_ss2d}.

\begin{algorithm}
\caption{Greedy Path Planning for GPS-SS2D}
\label{alg:gps_ss2d}
\KwIn{Patch scores $P = \{P_1, P_2, \ldots, P_n\}$}
\KwOut{ $\text{Path}_{\text{forward}}$,  $\text{Path}_{\text{backward}}$}

$\text{visited} \leftarrow \emptyset$\;
$\text{Path}_{\text{forward}} \leftarrow [\,]$\;
$\text{current\_patch} \leftarrow \arg\max_{i \in \{1,2,\ldots,n\}} P_i$\;

\While{$|\text{visited}| < n$}{
    Add $\text{current\_patch}$ to $\text{Path}_{\text{forward}}$\;
    Add $\text{current\_patch}$ to $\text{visited}$\;
    
    $\text{neighbors} \leftarrow \text{get\_neighbors}(\text{current\_patch})$ \tcp{up, down, left, right}
    $\text{unvisited\_neighbors} \leftarrow \text{neighbors} \setminus \text{visited}$\;
    
    \eIf{$\text{unvisited\_neighbors} \neq \emptyset$}{
        $\text{next\_patch} \leftarrow \arg\max_{j \in \text{unvisited\_neighbors}} P_j$\;
    }{
        $\text{unvisited\_patches} \leftarrow \{1,2,\ldots,n\} \setminus \text{visited}$\;
        \eIf{$\text{unvisited\_patches} \neq \emptyset$}{
            $\text{next\_patch} \leftarrow \arg\max_{k \in \text{unvisited\_patches}} P_k$\;
        }{
            \textbf{break} \tcp{All patches have been visited}
        }
    }
    
    $\text{current\_patch} \leftarrow \text{next\_patch}$\;
}

$\text{Path}_{\text{backward}} \leftarrow \text{reverse}(\text{Path}_{\text{forward}})$\;

\Return{$\text{Path}_{\text{forward}}$, $\text{Path}_{\text{backward}}$}
\end{algorithm}

The SS2D operation is then performed along these learned paths:

\begin{equation}X_{ss2d}=\mathrm{SS2D}(X,\mathrm{Path}_{forward},\mathrm{Path}_{backward})\end{equation}

This adaptive scanning strategy enables GPS-SS2D to prioritize patches with higher importance scores, ensuring that the state space model focuses computational resources on the most informative regions. The greedy algorithm guarantees sequential processing of neighboring high-importance patches when possible, maintaining spatial locality while adapting to content-specific degradation patterns. When all neighbors of a patch have been visited, the algorithm jumps to the highest-scoring position among the remaining unvisited patches, ensuring globally optimal scanning order. The bidirectional scanning paths provide complementary information flow, allowing the model to capture dependencies from different scanning perspectives and achieve superior restoration performance.

\section{Experiments and Analysis}
\subsection{Experimental Settings}
\noindent\textbf{Datasets.} We conduct experiments on multiple standard image restoration benchmarks to evaluate the performance of the proposed VAMamba model. For denoising, we use BSD400~\cite{martin2001database} for training and BSD68~\cite{roth2009fields} for testing. For the deraining task, we use Rain13k~\cite{yang2017deep, hu2019depth} as the training dataset, and Rain100H, Rain100L~\cite{yang2017deep}, Test1200~\cite{zhang2018density}, and Test2800~\cite{fu2017removing} as testing datasets.  For motion deblurring, we train the model on the GoPro dataset~\cite{nah2017deep} and evaluate it on GoPro, HIDE~\cite{shen2019humanaware}, RealBlur-J, and RealBlur-R~\cite{rim2020realblur}.  For defocus blur removal, we adopt DPDD(train)~\cite{abuolaim2020defocus} as the training set and DPDD(test)~\cite{abuolaim2020defocus} for evaluation. For dehazing, we employ the RESIDE SOTS (Outdoor) dataset~\cite{li2018benchmarking} for both training and testing. Finally, for low-light enhancement, we train and test on LOL datasets~\cite{wei2018deep}.

\noindent\textbf{Comparison Methods.} 
For all tasks, we compare our method to the prevailing approaches in
their respective fields and we adopt PSNR~\cite{PSNR_thu} , SSIM~\cite{SSIM} and  LPIPS~\cite{zhang2018unreasonable} as evaluation metrics.

\noindent\textbf{Training Details.} 
The VAMamba is trained for 200k iterations with an initial learning rate of \(2 \times 10^{-4}\) using the AdamW optimizer. 
A cosine annealing strategy is applied to decay the learning rate. The batch size is set to $16$, and training is conducted on two NVIDIA RTX 4090 GPUs. To stabilize training, we apply data augmentation, including random cropping, horizontal and vertical flipping, and rotation. The input image patches are cropped to \(256 \times 256\). The loss function combines \(L_1\) loss and FFT-based frequency-domain loss, defined as:
\begin{equation}
    \mathcal{L} = \| I_{\text{HR}} - I_{\text{GT}} \|_1 + \lambda \| \mathcal{F}(I_{\text{HR}}) - \mathcal{F}(I_{\text{GT}}) \|_1,
\end{equation}
where \(\lambda\) is empirically set to $0.05$. This hybrid loss helps balance pixel-level fidelity and high-frequency detail preservation.

\noindent\textbf{Hyperparameter Settings.} 
we set the LoRA rank to 16, the cache size in QCLAM to 5, and the patch size for GPS-SS2D to 64×64. These values are selected through preliminary experiments to achieve a good balance between accuracy and efficiency.

\subsection{Main Results}

\noindent \textbf{Image Denoising Results.}We further include the gaussian color image denoising task for additional validation. VAMamba demonstrates excellent performance on the denoising task, particularly excelling at noise level 15 in Table.~\ref{tab:denoising_results}. As shown in Fig.~\ref{fig: denoise}, VAMamba can recover more texture details even for some very thin edges. Our adaptive scanning mechanism GPS-SS2D can intelligently identify noise patterns and real image content, avoiding over-smoothing important image details during the denoising process. The QCLAM module reuses relevant denoising patterns through similarity matching, improving denoising quality while maintaining computational efficiency.

\noindent\textbf{Image Deraining Results.}Tab.~\ref{tab: derain} demonstrates that VAMamba achieves consistently superior performance across all evaluation metrics and datasets. On Rain100H dataset, VAMamba achieves 31.81 dB PSNR, surpassing FourierMamba (31.79 dB) while maintaining identical SSIM (0.913) and achieving better perceptual quality with LPIPS of 0.210 vs 0.212. The improvement becomes more pronounced on Rain100L dataset, where VAMamba achieves 39.90 dB PSNR compared to FourierMamba's 39.73 dB, representing a solid 0.17 dB improvement. Notably, VAMamba achieves significantly better perceptual quality with LPIPS of 0.112 compared to FourierMamba's 0.115, despite FourierMamba having a marginally higher SSIM (0.986 vs 0.981).

\begin{figure*}[!htb]
    \centering
    \begin{adjustbox}{max width=\textwidth}
    \begin{tabular}{cccccccc}
        \includegraphics[width=0.26\textwidth]{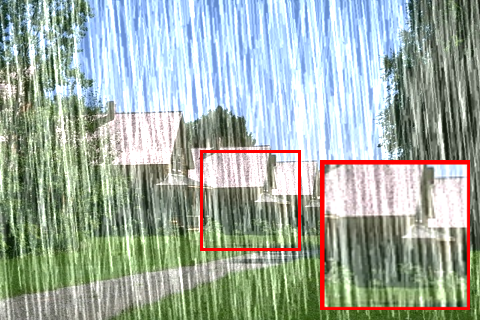} &
        \includegraphics[width=0.26\textwidth]{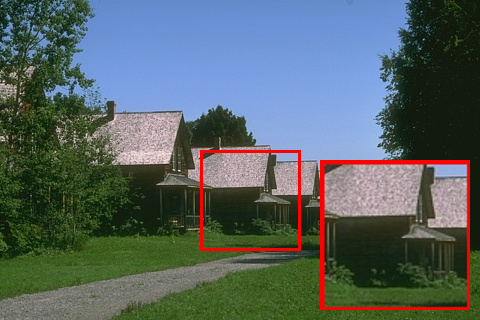} &
        \includegraphics[width=0.26\textwidth]{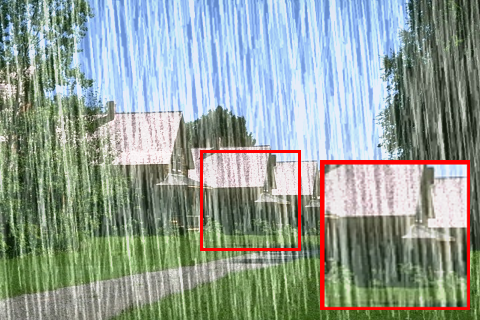} &
        \includegraphics[width=0.26\textwidth]{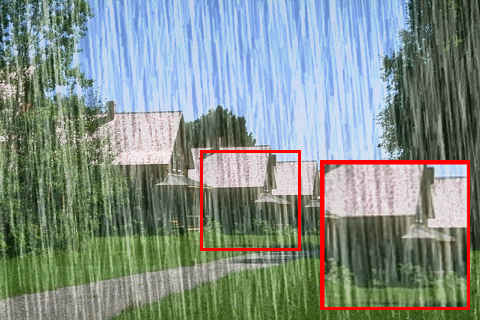} &
        \includegraphics[width=0.26\textwidth]{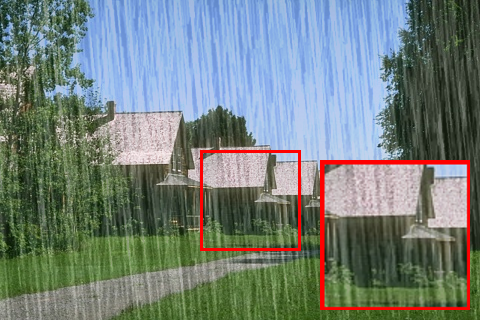} &
        \includegraphics[width=0.26\textwidth]{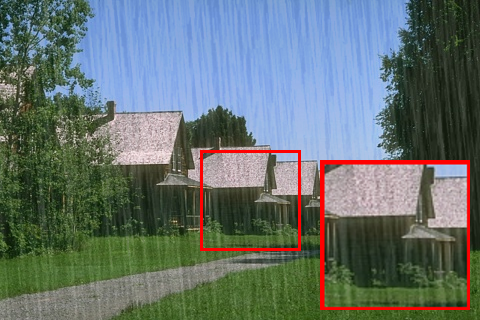} &
        \includegraphics[width=0.26\textwidth]{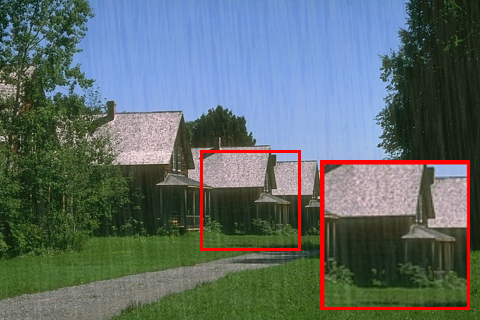} &
         \\
        \includegraphics[width=0.26\textwidth]{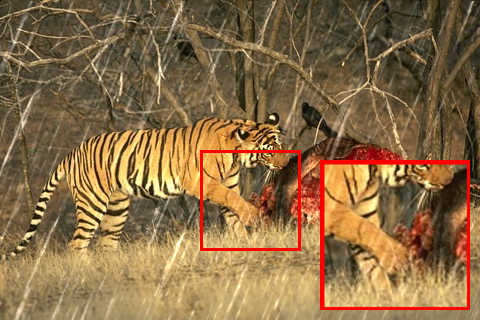} &
        \includegraphics[width=0.26\textwidth]{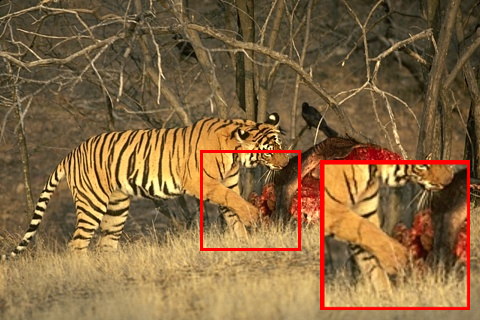} &
        \includegraphics[width=0.26\textwidth]{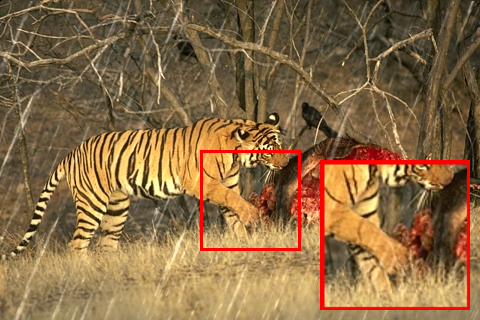} &
        \includegraphics[width=0.26\textwidth]{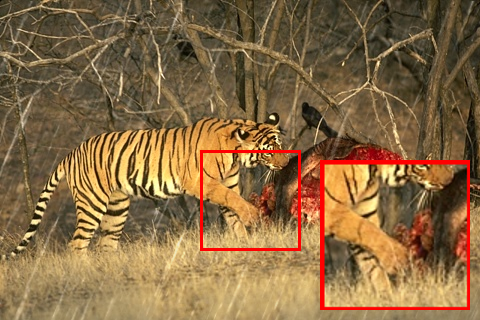} &
        \includegraphics[width=0.26\textwidth]{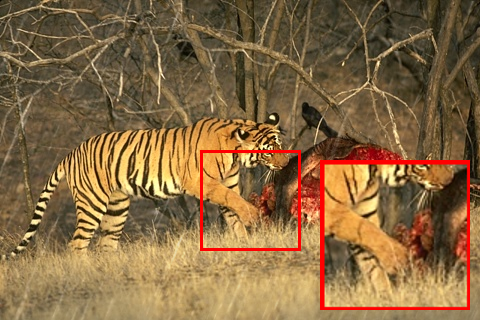} &
        \includegraphics[width=0.26\textwidth]{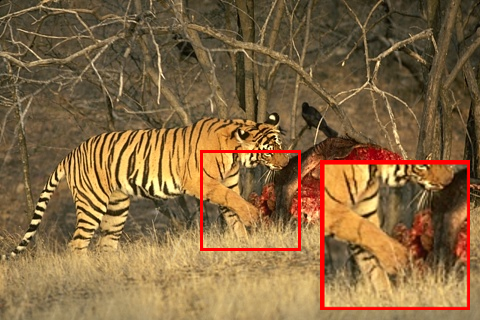} &
        \includegraphics[width=0.26\textwidth]{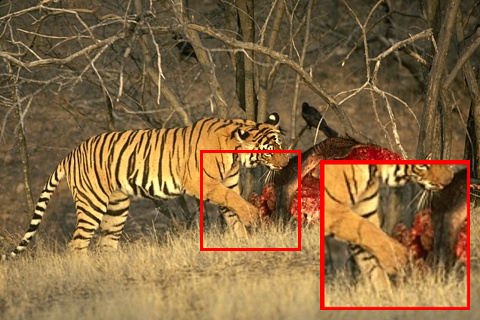}\\
        \includegraphics[width=0.26\textwidth]{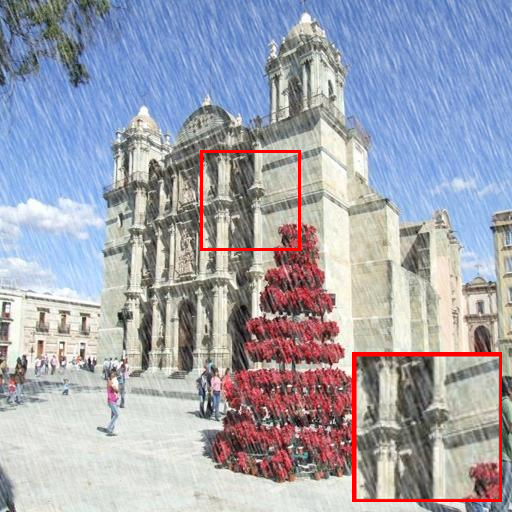} &
        \includegraphics[width=0.26\textwidth]{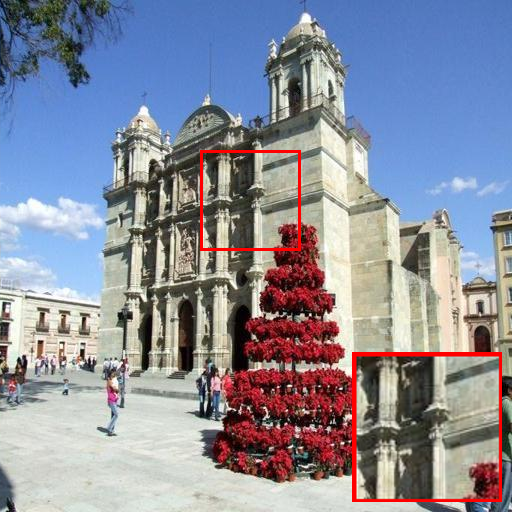} &
        \includegraphics[width=0.26\textwidth]{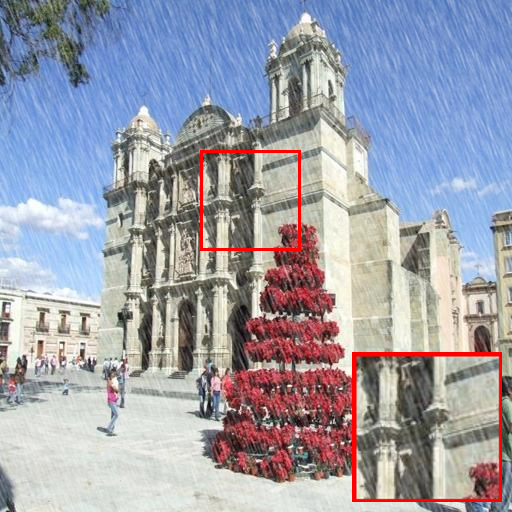} &
        \includegraphics[width=0.26\textwidth]{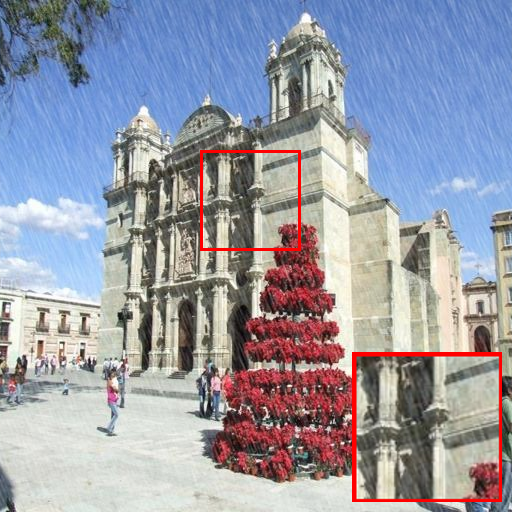} &
        \includegraphics[width=0.26\textwidth]{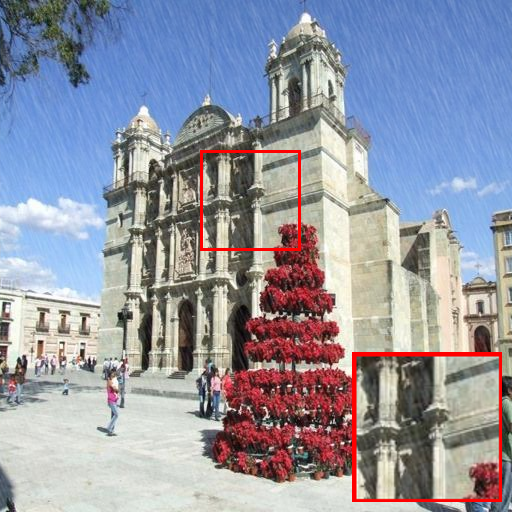} &
        \includegraphics[width=0.26\textwidth]{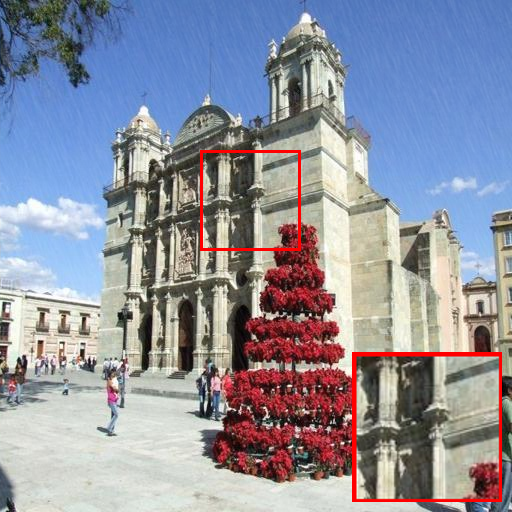} &
        \includegraphics[width=0.26\textwidth]{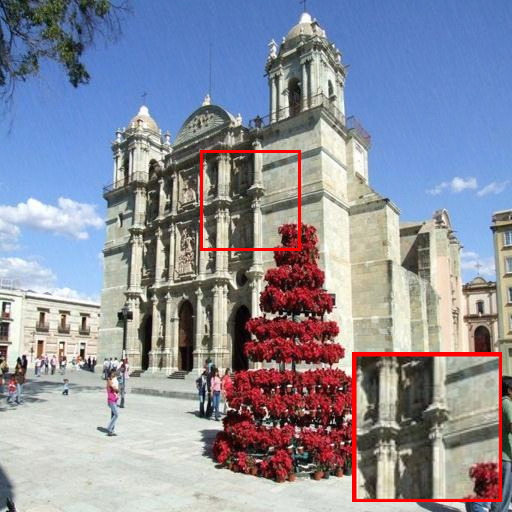}\\
        \includegraphics[width=0.26\textwidth]{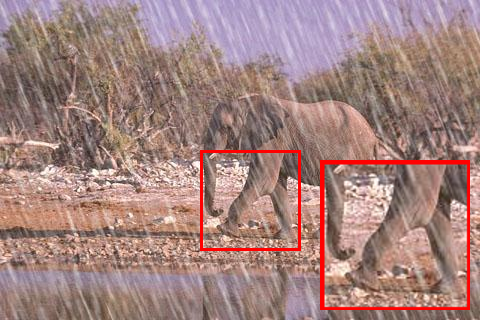} &
        \includegraphics[width=0.26\textwidth]{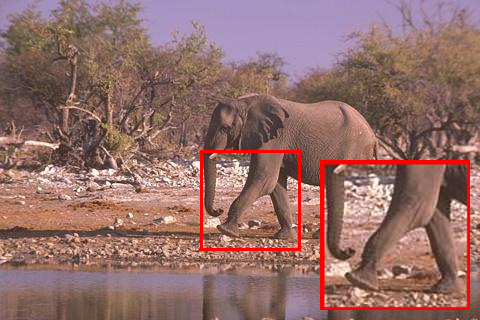} &
        \includegraphics[width=0.26\textwidth]{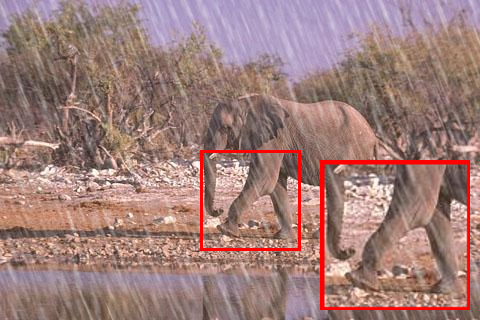} &
        \includegraphics[width=0.26\textwidth]{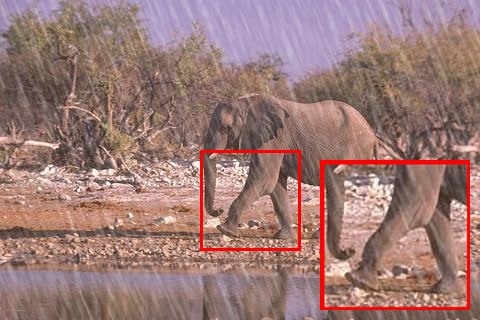} &
        \includegraphics[width=0.26\textwidth]{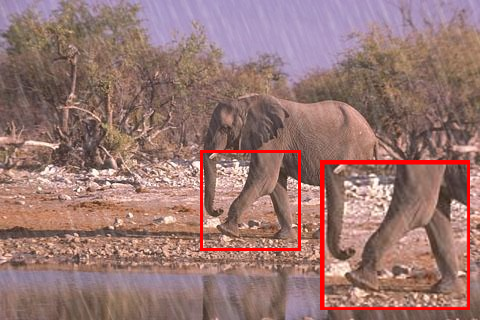} &
        \includegraphics[width=0.26\textwidth]{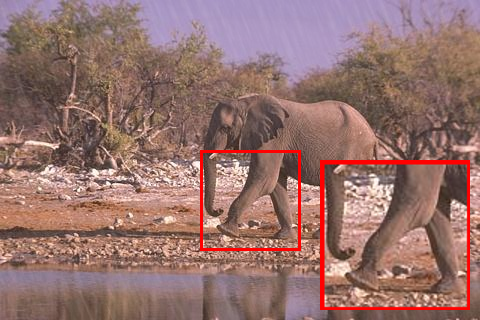} &
        \includegraphics[width=0.26\textwidth]{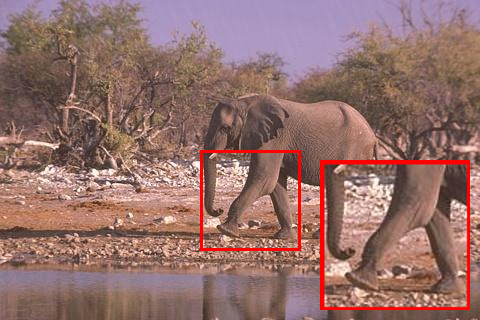}\\
       \huge Input &\huge GT & \huge MambaIR &\huge VMambaIR & \huge FreqMamba & \huge FourierMamba & \huge Ours \\
    \end{tabular}
    \end{adjustbox}
    \caption{Visual comparison of image deraining results on four standard datasets: Rain100H, Rain100L, Test1200, and Test2800 (from top to bottom). The proposed VAMamba achieves superior restoration quality, generating clearer and more visually appealing results compared to other methods. Red boxes highlight important details, which are zoomed in and displayed in the bottom-right corner of each image for a better view. }
    \label{fig: derain}
\end{figure*}

The performance advantages are consistent across challenging test datasets: on Test2800, while FourierMamba achieves slightly higher PSNR (34.23 vs 34.10), VAMamba demonstrates superior structural preservation (SSIM 0.951 vs 0.949) and notably better perceptual quality (LPIPS 0.162 vs 0.168). The images in Fig.~\ref{fig: derain} are heavily obstructed by bright, linear rain streaks, which pose a significant challenge for traditional Mamba methods. Other methods often struggle to capture sharp and heavy patterns. In contrast, our approach effectively removes these obstructions, demonstrating notable improvements over state-of-the-art methods. VAMamba excels in removing complex rain streaks while preserving fine details—the tiger's texture patterns remain sharp and natural, architectural details are cleanly restored, and background elements maintain their original clarity without artifacts.

\begin{figure*}[!htb]
    \centering
    \begin{adjustbox}{max width=\textwidth}
    \begin{tabular}{cccccccc}
        \includegraphics[width=0.26\textwidth]{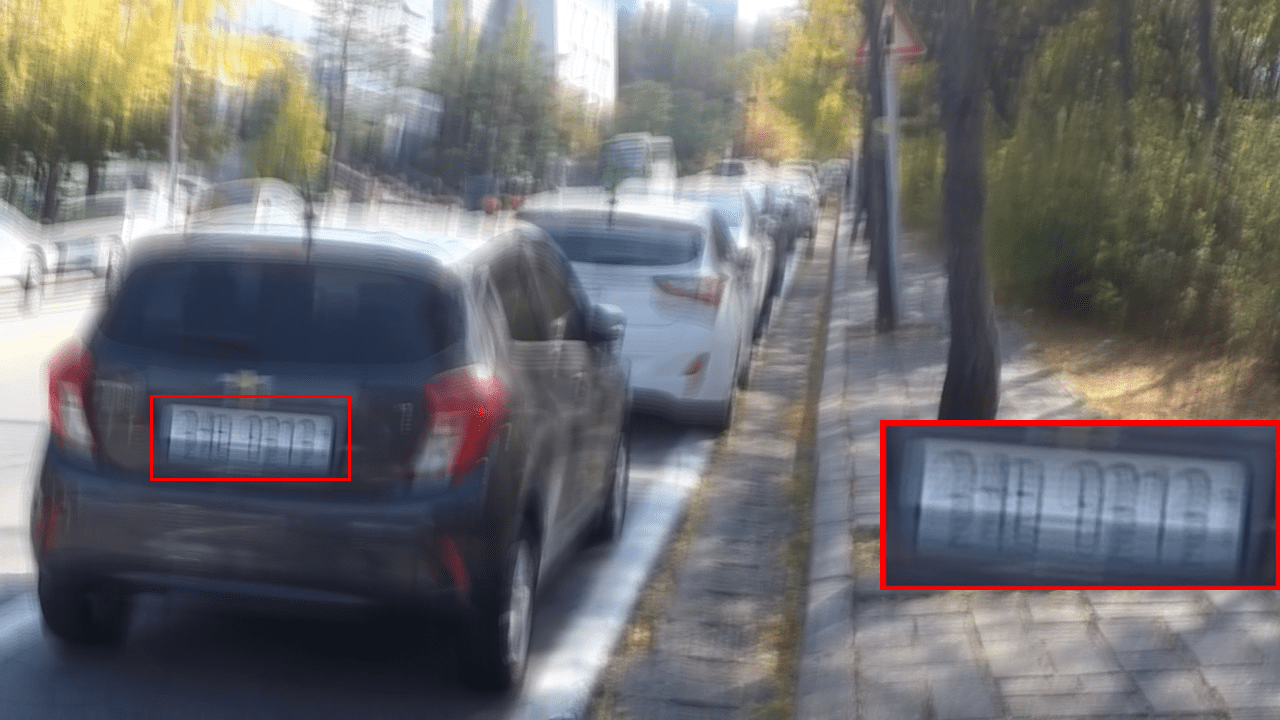} &
        \includegraphics[width=0.26\textwidth]{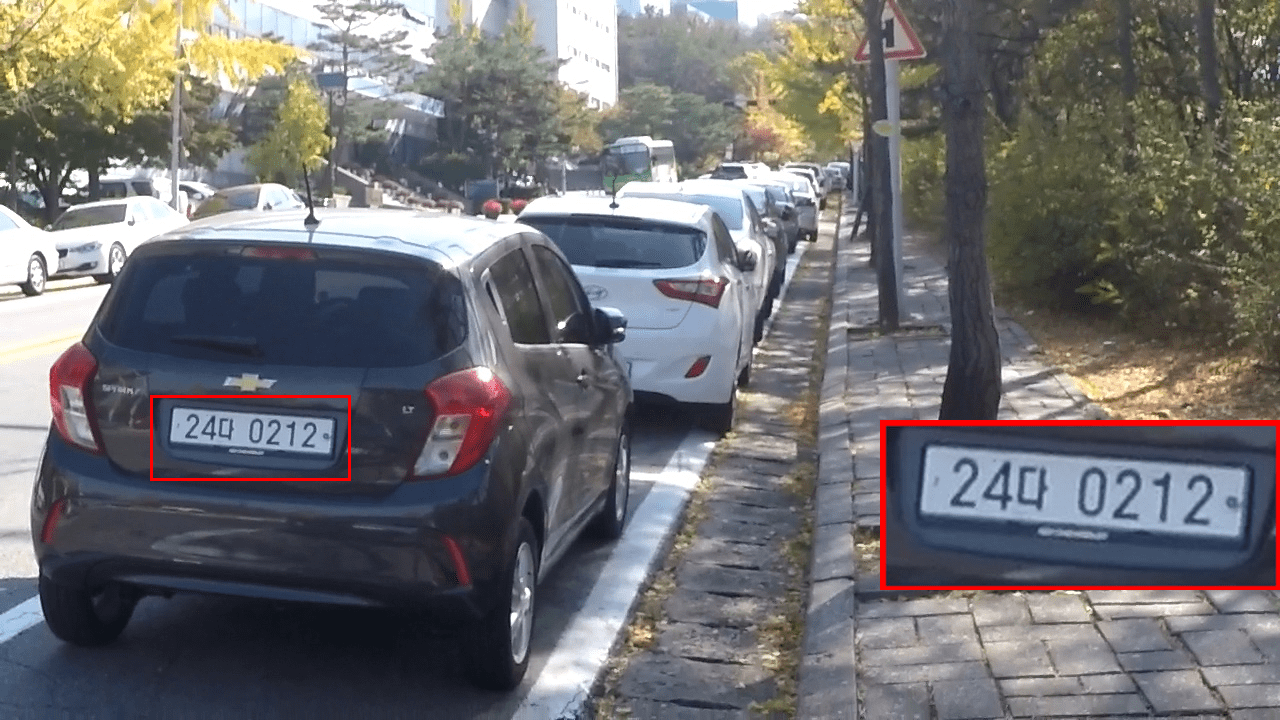} &
        \includegraphics[width=0.26\textwidth]{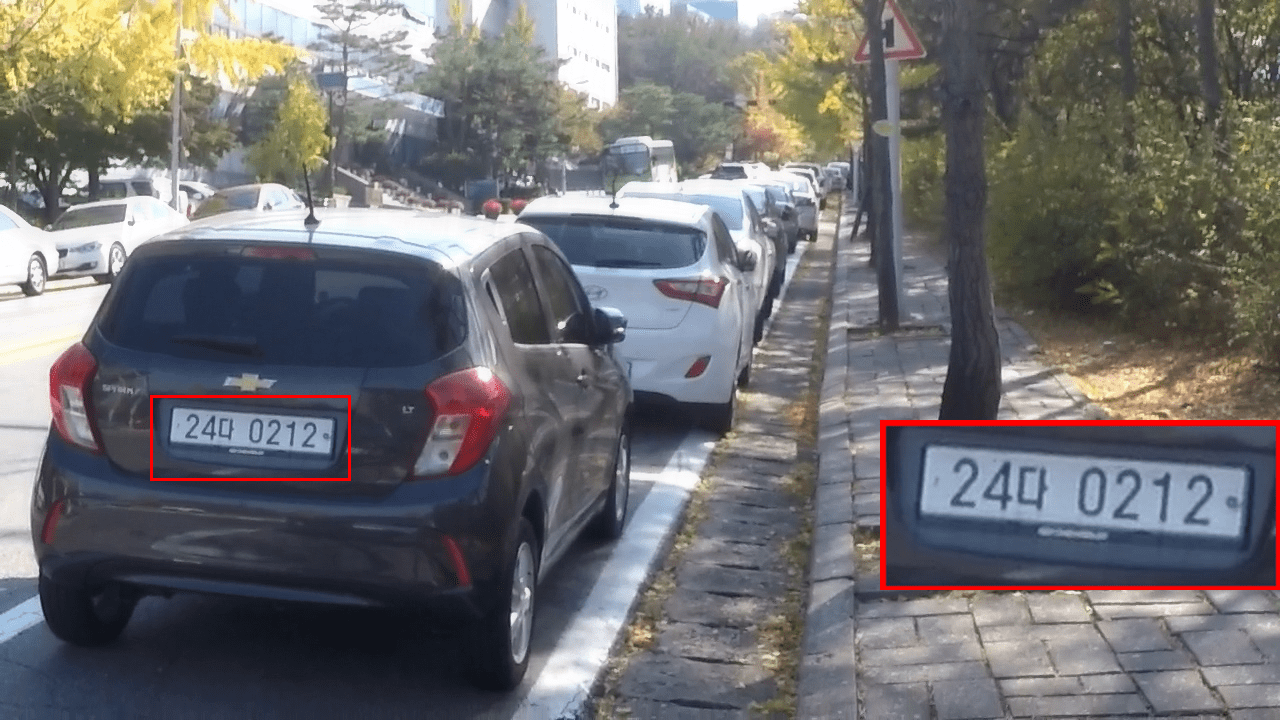} &
        \includegraphics[width=0.26\textwidth]{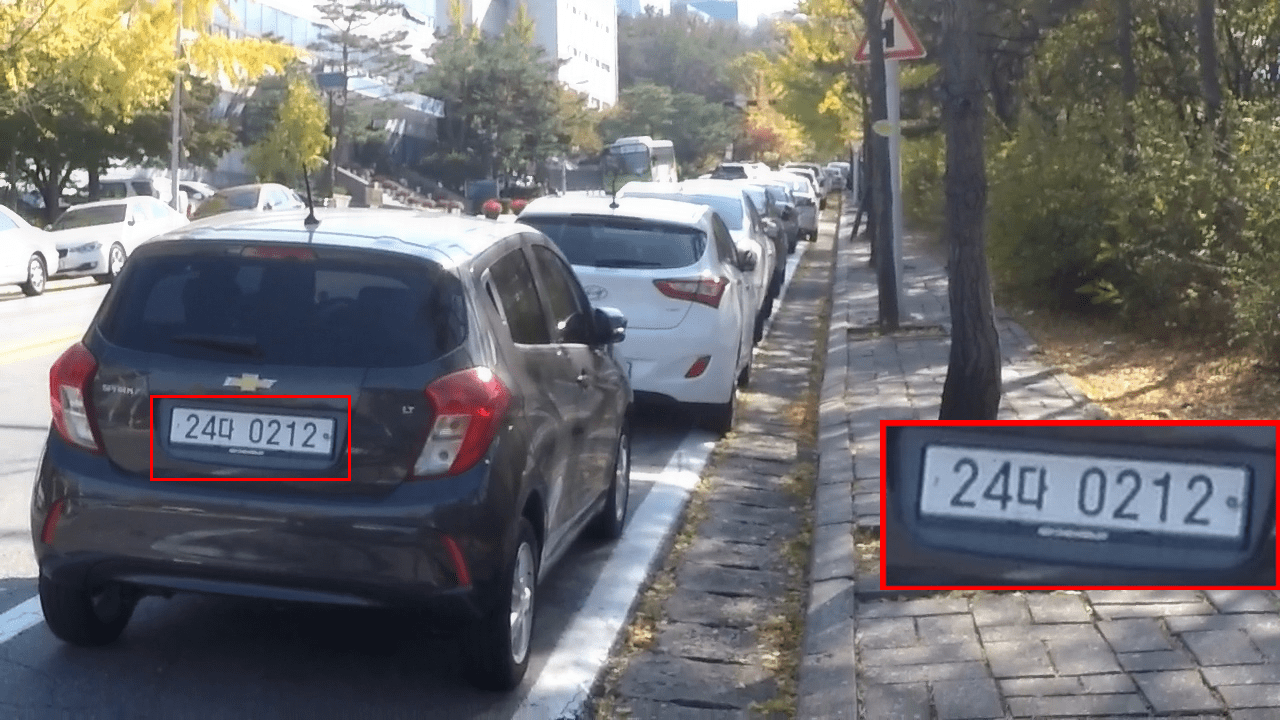} &
        \includegraphics[width=0.26\textwidth]{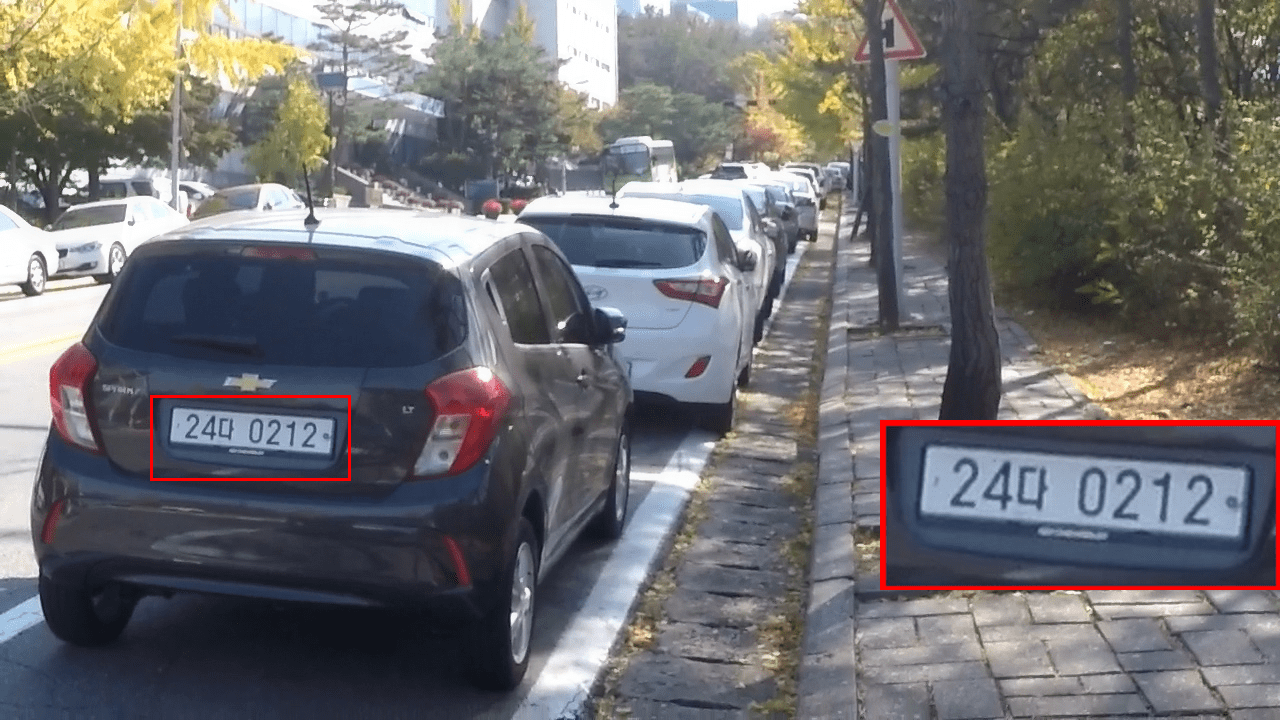} &
        \includegraphics[width=0.26\textwidth]{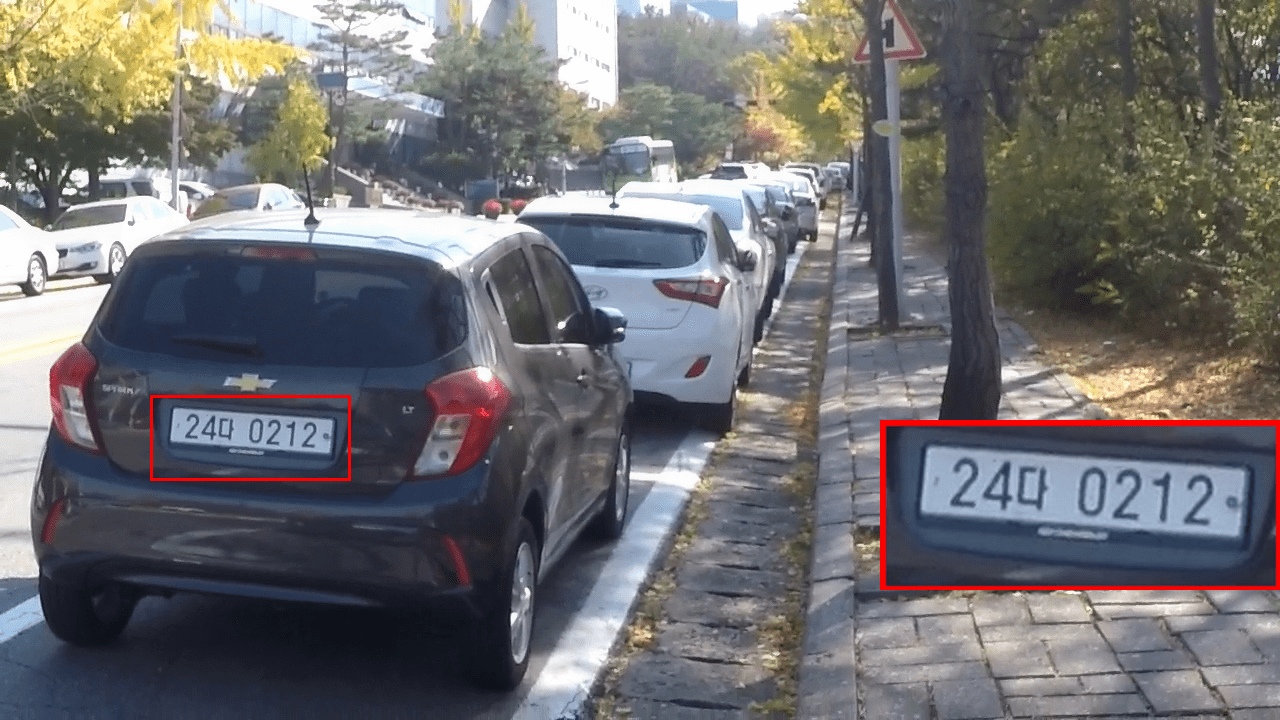} &
        \includegraphics[width=0.26\textwidth]{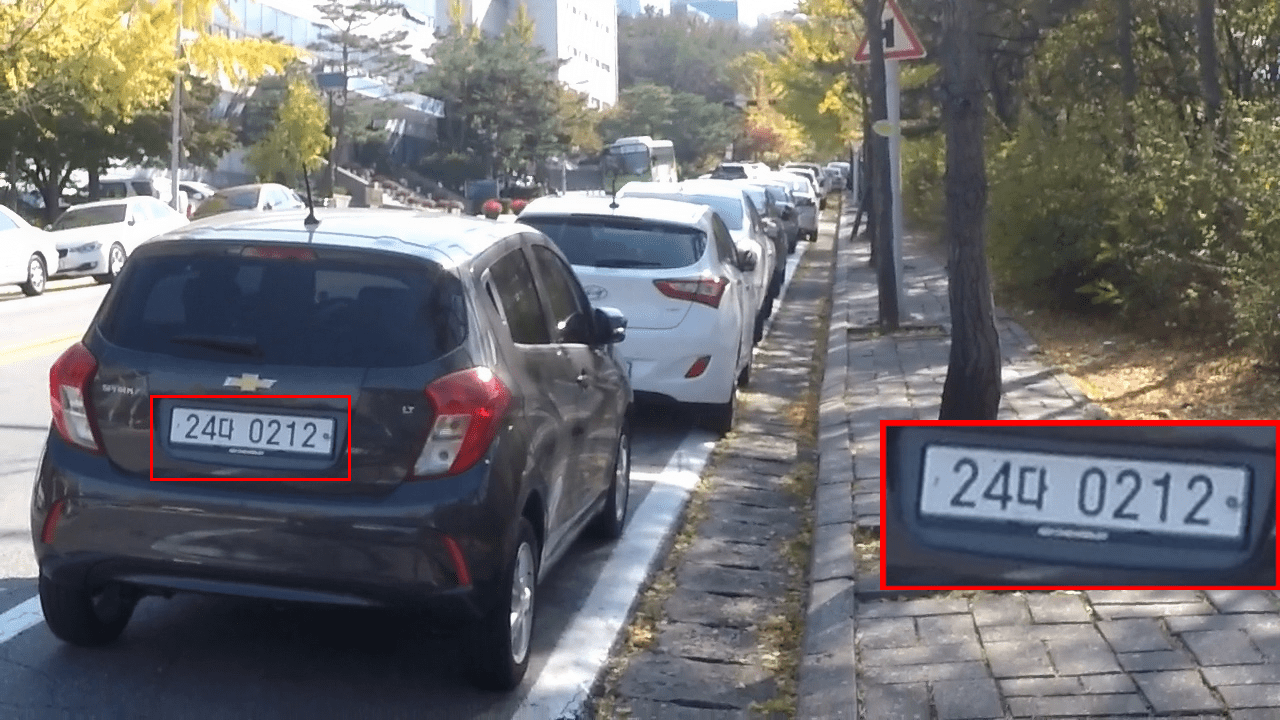} &
         \\
        \includegraphics[width=0.26\textwidth]{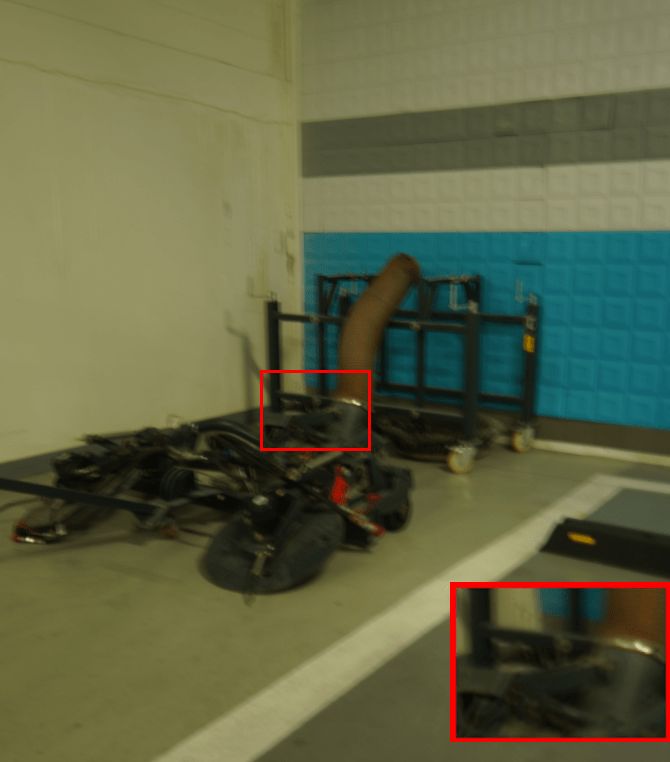} &
        \includegraphics[width=0.26\textwidth]{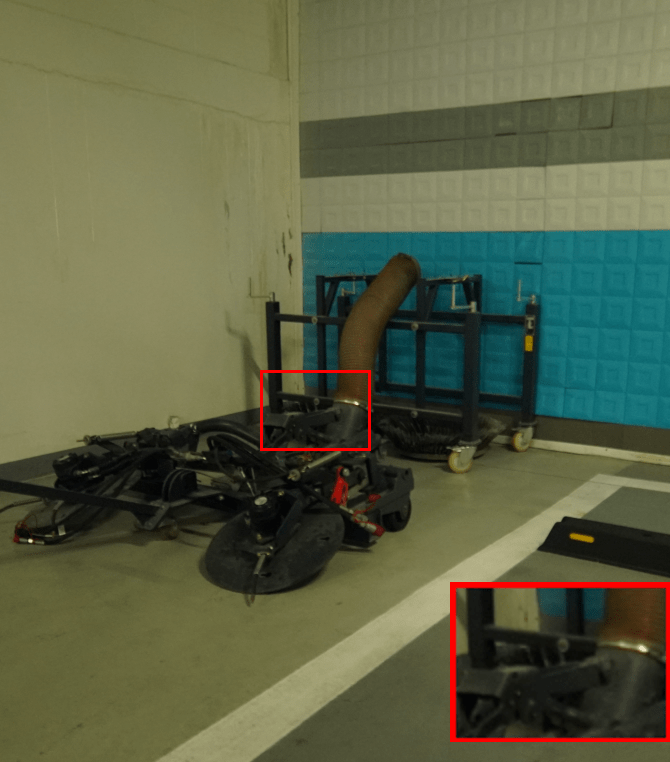} &
        \includegraphics[width=0.26\textwidth]{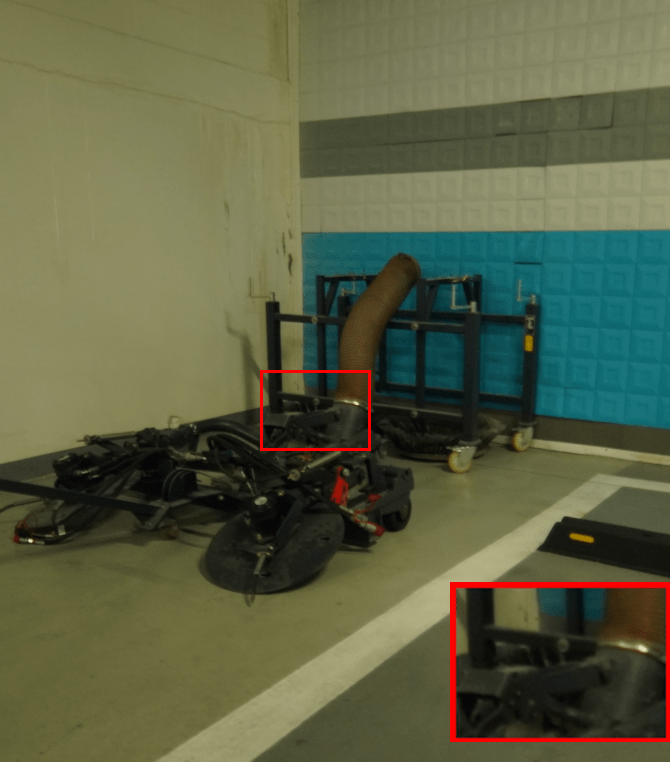} &
        \includegraphics[width=0.26\textwidth]{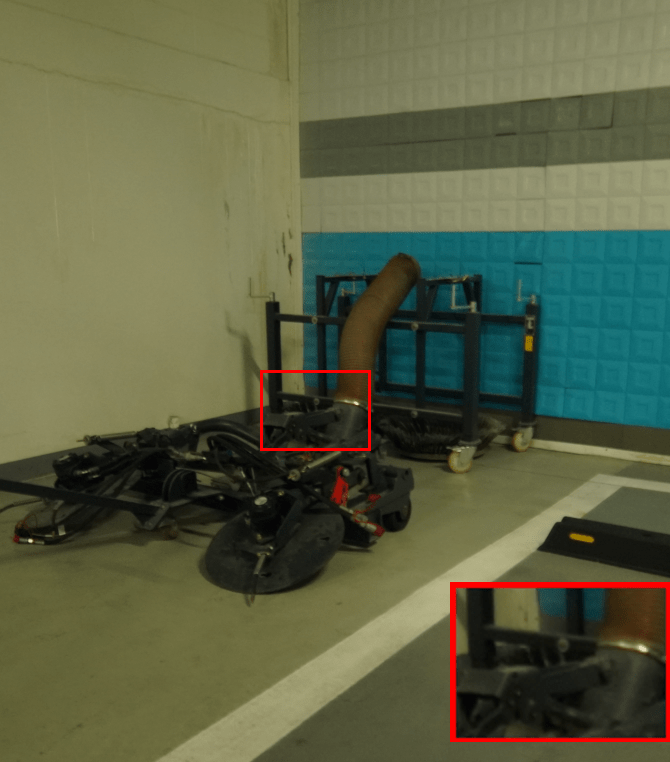} &
        \includegraphics[width=0.26\textwidth]{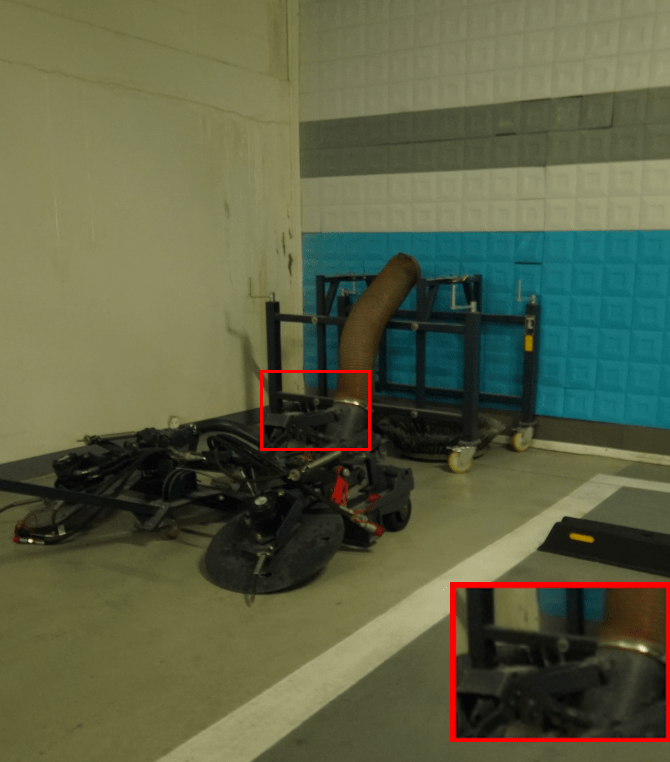} &
        \includegraphics[width=0.26\textwidth]{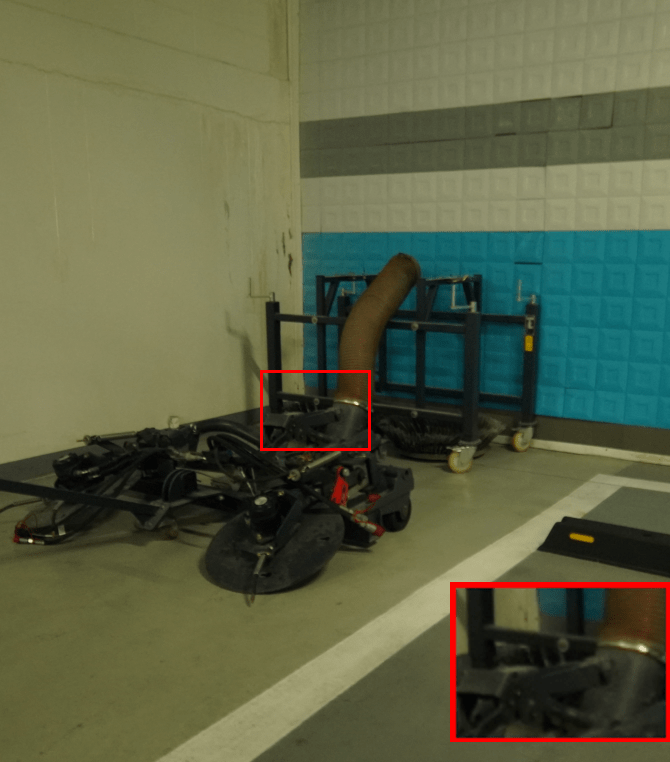} &
        \includegraphics[width=0.26\textwidth]{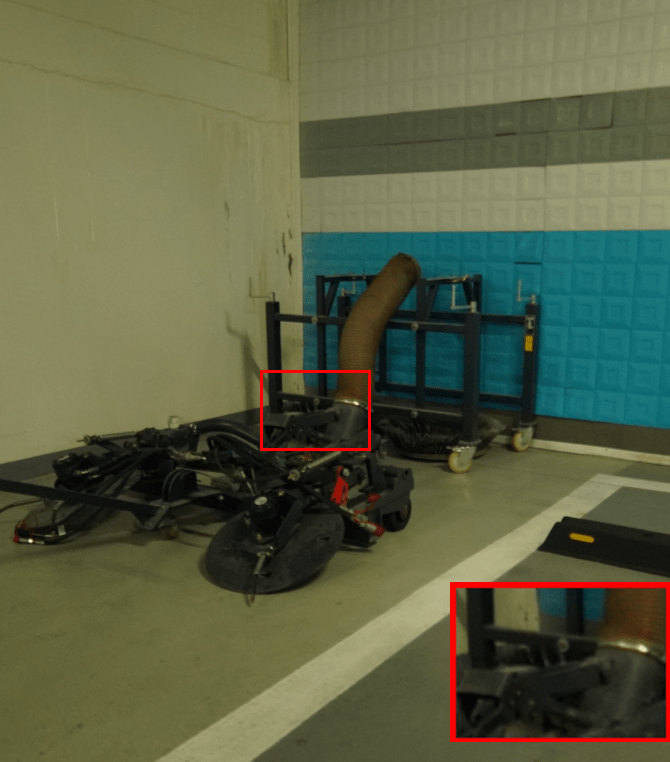} &
         \\
        \includegraphics[width=0.26\textwidth]{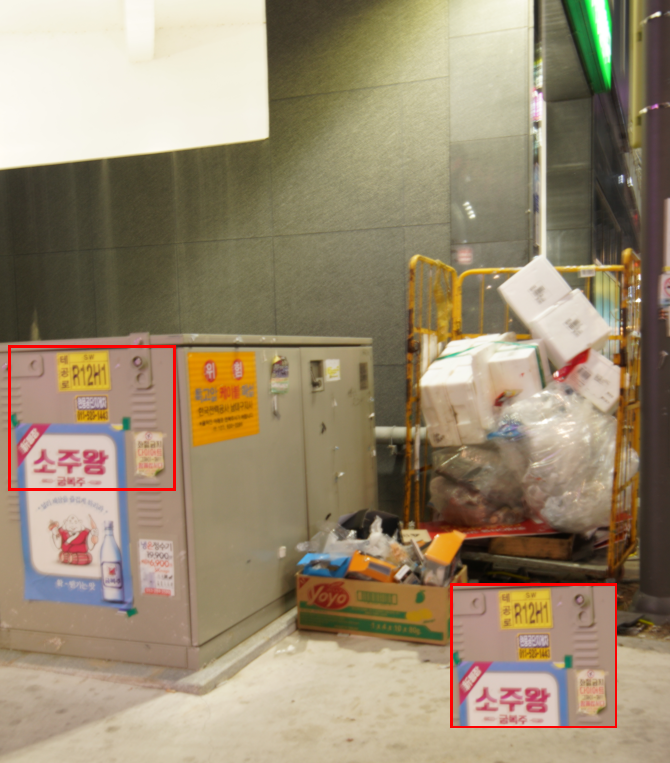} &
        \includegraphics[width=0.26\textwidth]{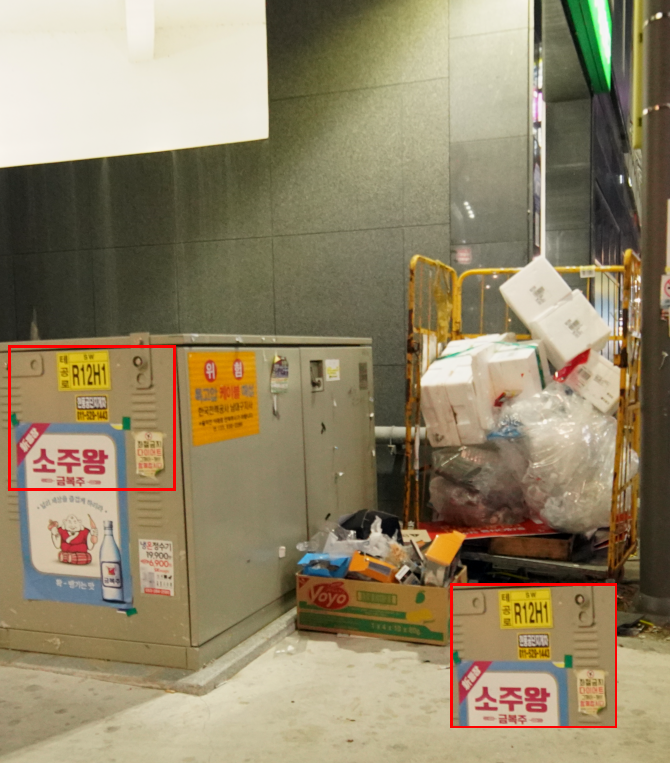} &
        \includegraphics[width=0.26\textwidth]{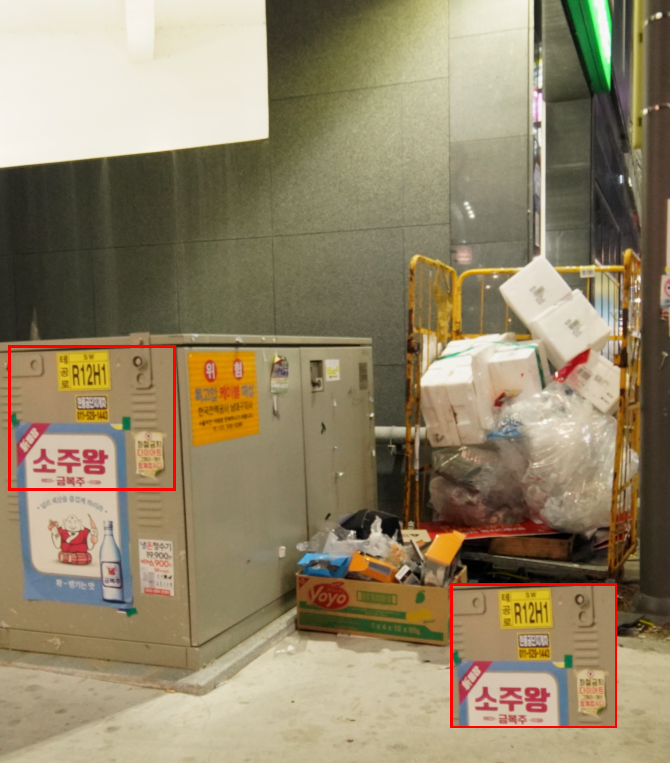} &
        \includegraphics[width=0.26\textwidth]{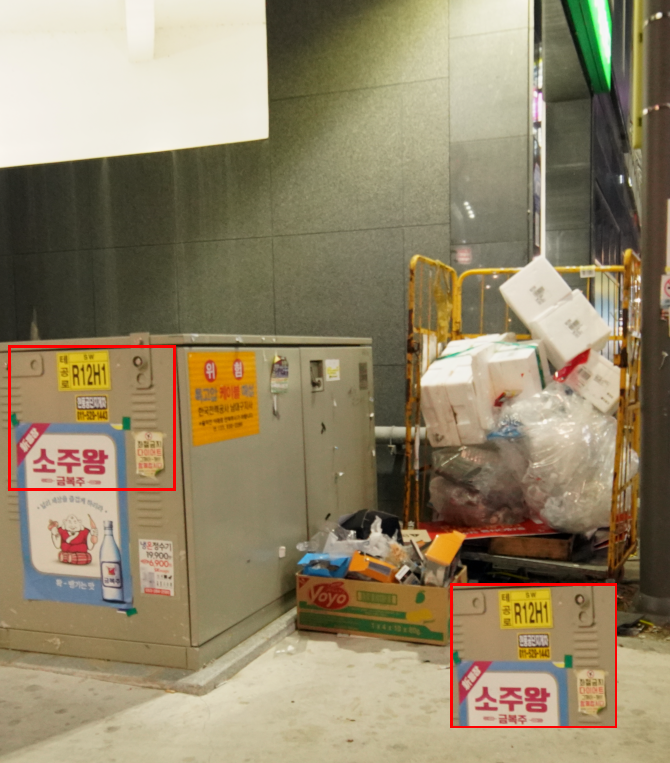} &
        \includegraphics[width=0.26\textwidth]{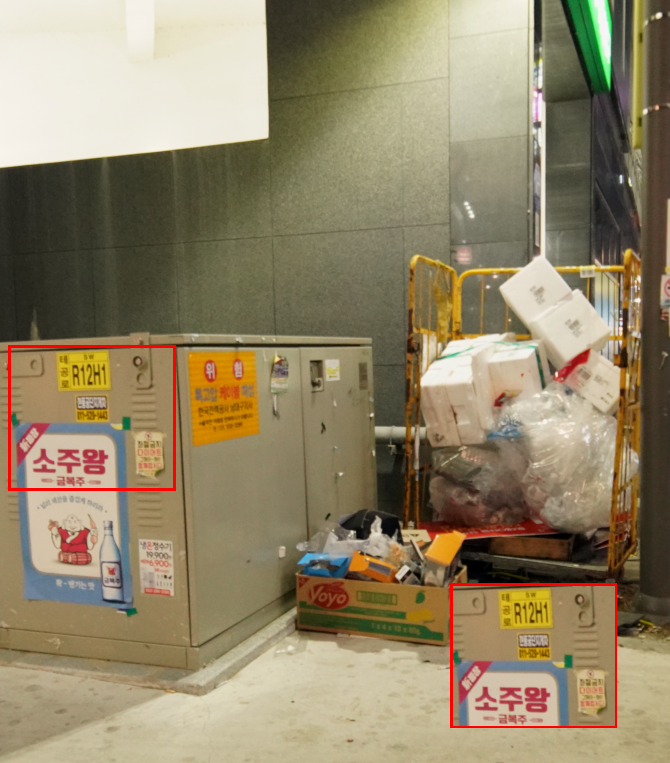} &
        \includegraphics[width=0.26\textwidth]{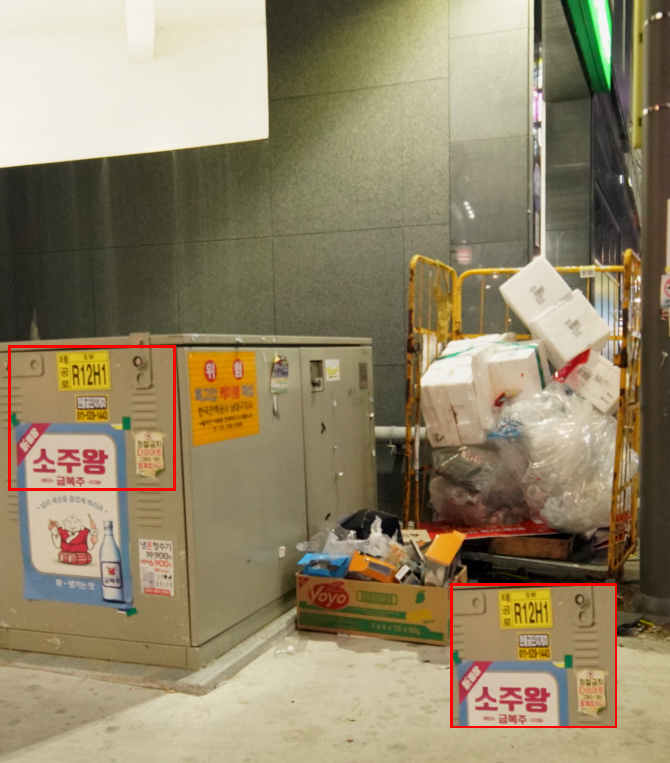} &
        \includegraphics[width=0.26\textwidth]{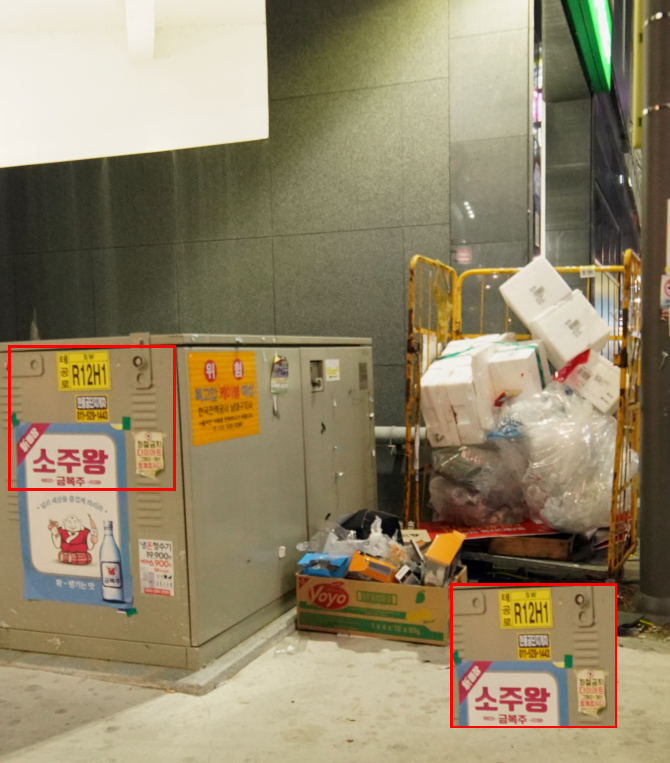} &\\
        
       \huge Input &\huge GT & \huge MPRNet &\huge Restormer & \huge Stripformer & \huge FFTformer & \huge Ours \\
    \end{tabular}
    \end{adjustbox}
    \caption{Single image motion deblurring results. Our VAMamba generates sharper and visually-faithful details.}
    \label{fig: deblur}\vspace{-5mm}
\end{figure*}

\noindent\textbf{Single-image Motion Deblurring Results.}Table.~ \ref{tab: deblur} demonstrates VAMamba's exceptional cross-dataset generalization capabilities. Although trained exclusively on the GoPro dataset, our method outperforms other approaches on all four benchmark datasets. Fig.~\ref{fig: deblur} demonstrates that the images generated by our method are sharper and visually more aligned with the ground truth compared to those produced by other algorithms.

\begin{figure*}[!htb]
\centering
\begin{tabular}{ccccc}
    \includegraphics[width=0.16\linewidth]{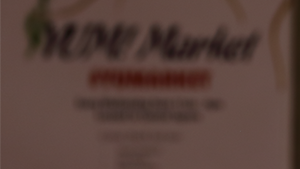} &
    \includegraphics[width=0.16\linewidth]{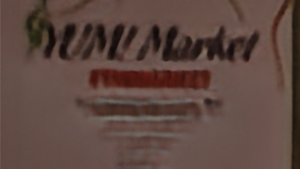} &
    \includegraphics[width=0.16\linewidth]{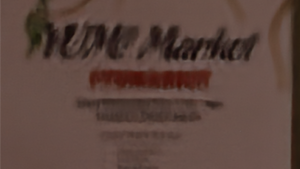} &
    \includegraphics[width=0.16\linewidth]{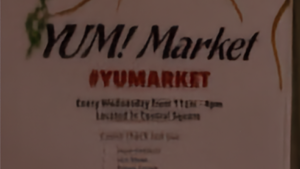} &
    \includegraphics[width=0.16\linewidth]{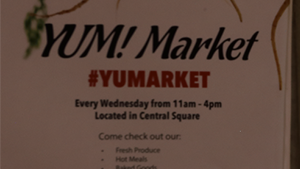} \\
    \small Input &
    \small IFAN   &
    \small Restormer  &
    \small Ours &
    \small GT
    \\
    \small 27.49~dB & \small 28.64~dB&\small 29.46~dB & \small\textbf{31.32~dB} &\small PSNR\\
    \includegraphics[width=0.16\linewidth]{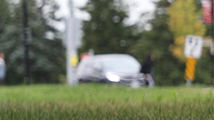} &
    \includegraphics[width=0.16\linewidth]{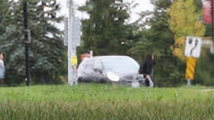} &
    \includegraphics[width=0.16\linewidth]{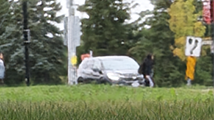} &
    \includegraphics[width=0.16\linewidth]{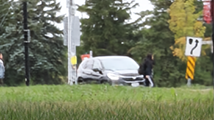} &
    \includegraphics[width=0.16\linewidth]{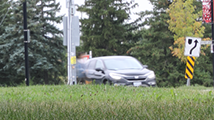} \\
    \small Input  &
    \small IFAN  &
    \small Restormer &
    \small Ours &
    \small GT\\  
    \small 21.02~dB & \small 22.37~dB &\small23.23~dB & \small \textbf{24.12~dB} & \small PSNR\\
\end{tabular}\vspace{-2mm}
\caption{Visual comparisons with state-of-the-art methods for defocus blur removal.}\vspace{-2mm}
\label{fig: dpdd}
\end{figure*}

\noindent \textbf{Defocus Deblurring Results.}Following the approach of Restormer~\cite{zamir2022restormer}, we evaluate our method on the DPDD dataset~\cite{abuolaim2020defocus}. Table.~\ref{tab: dpdd} presents the image fidelity scores, where our proposed model achieves significant improvements across all scene categories, outperforming state-of-the-art schemes. Fig.~\ref{fig: dpdd} further illustrates that our method effectively removes spatially varying defocus blur.

\begin{figure*}[!htb]
	\centering
	\includegraphics[width=1\linewidth]{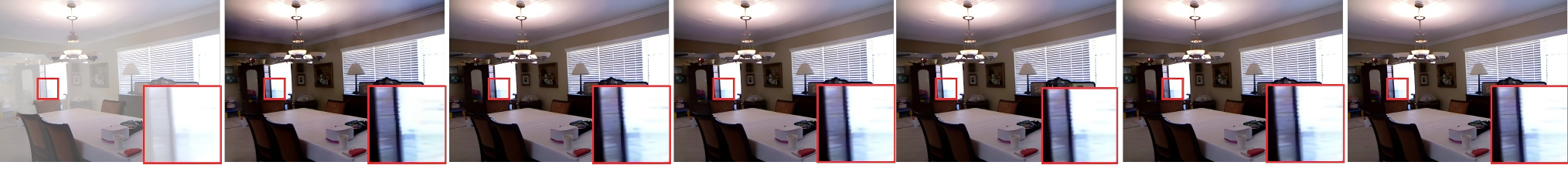}\\
	\includegraphics[width=1\linewidth]{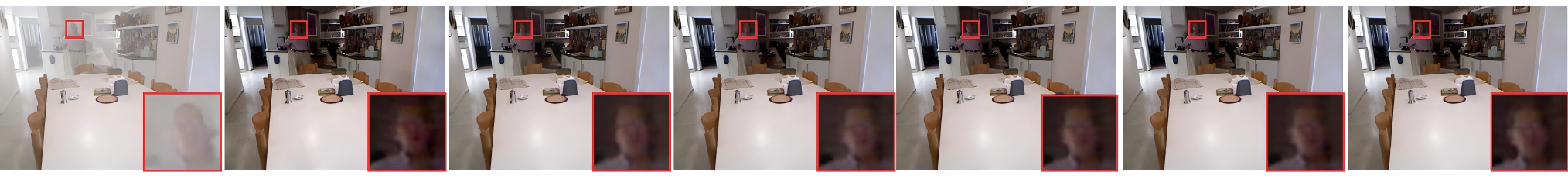}
	\begin{tabular}{ccccccc}
    \makebox[2.5cm]{Input} & \makebox[2.5cm]{GridDehazeNet} & \makebox[2.5cm]{FFA-Net} & \makebox[2.5cm]{DehazeFormer} & \makebox[2.5cm]{MambaIR} & \makebox[2.5cm]{Ours} & \makebox[2.5cm]{GT}\\
	\end{tabular}
	\caption{Visual comparisons on a synthetic hazy image sampled from SOTS-Indoor dataset.}
	\label{fig:ohaze_comparison_indoor}
\end{figure*}

\noindent \textbf{Image  Dehazing Results.}Table.~\ref{tab:dehaze} presents the image dehazing results on the SOTS-Indoor, SOTS-Outdoor, and Haze4K datasets, where VAMamba outperforms other model methods. Fig.~\ref{fig:ohaze_comparison_outdoor} and Fig.~\ref{fig:ohaze_comparison_indoor} show visual comparisons for the SOTS-Outdoor and SOTS-Indoor datasets, respectively. It can be observed that our proposed model achieves superior dehazing performance.

\noindent \textbf{Low Light Image Enhancement Results.}For the low-light image enhancement task, we employed three evaluation metrics: PSNR, SSIM, and LPIPS.
As shown in Table.~\ref{tab:lol_enhancement}, VAMamba achieves the highest PSNR and LPIPS scores while securing the second-highest SSIM on both the LOLv1~\cite{we2018deep} and LOL-v2-real~\cite{yang2021sparse} datasets, demonstrating competitive performance across all metrics.
Evidently, as shown Fig.~\ref{fig:lowlight_comparison} prior methods exhibit inferior performance in noise suppression and often produce results with noticeable color distortion, as seen in the enhanced outputs of KinD, LLFlow, and Retformer.

\begin{table}[!htb]
\centering
\caption{\underline{\textbf{Gaussian Color Image Denoising}} results ($\sigma = 15$). 
We report PSNR on different datasets. \colorbox{best}{best} and \colorbox{secondbest}{secondbest} are highlighted.}
\label{tab:denoising_results}
\resizebox{\linewidth}{!}{
\begin{tabular}{l|c|cccc}
\toprule[0.15em]
\textbf{Method} & \textbf{Venue} & \textbf{CBSD68} & \textbf{Kodak24} & \textbf{McMaster} & \textbf{Urban100} \\
\midrule[0.1em]
SwinIR       & ICCV'21  & 34.42 & 35.34 & 35.61 & 34.96 \\              
Restormer    & CVPR'22  & 34.40 & 35.35 & 35.61 & 35.11 \\
Xformer      & CVPR'23  & 34.43 & 35.39 & 35.68 & 35.29 \\
MambaIR      & ECCV'24  & 34.48 & 35.42 & 35.70 & 35.37 \\
MambaIRv2    & arXiv'24 & \cellcolor{secondbest}\underline{34.48} 
                        & \cellcolor{secondbest}\underline{35.43} 
                        & \cellcolor{best}\textbf{35.73} 
                        & \cellcolor{secondbest}\underline{35.42} \\
\rowcolor{color4}
\textbf{VAMamba (Ours)} 
             & --       & \cellcolor{best}\textbf{34.50} 
                        & \cellcolor{best}\textbf{35.44} 
                        & \cellcolor{secondbest}\underline{35.72} 
                        & \cellcolor{best}\textbf{35.44} \\
\bottomrule[0.15em]
\end{tabular}
}
\vspace{-2mm}
\end{table}


\begin{table*}[!htb]
\centering
\caption{ \underline{\textbf{Image Deraining}} results across multiple datasets.}
\label{tab: derain}
\setlength{\tabcolsep}{6pt}
\vspace{-2mm}
   \resizebox{0.95\linewidth}{!}{
    \begin{tabular}{c|c|ccc|ccc|ccc|ccc}
    \toprule[0.15em]
    \multirow{2}{*}{\textbf{Method}}&\multirow{2}{*}{\textbf{Venue}}& \multicolumn{3}{c|}{\textbf{Rain100H}} & \multicolumn{3}{c|}{\textbf{Rain100L}} & \multicolumn{3}{c|}{\textbf{Test2800}} & \multicolumn{3}{c}{\textbf{Test1200}}\\
         &   & PSNR & SSIM & LPIPS & PSNR & SSIM & LPIPS & PSNR & SSIM & LPIPS & PSNR & SSIM & LPIPS\\
    \midrule
    DerainNet &TIP'17& 14.92 & 0.592 & 0.458 & 27.03 & 0.884 & 0.287 & 24.31 & 0.861 & 0.315 & 23.38 & 0.835 & 0.342\\
    UMRL &CVPR'19 &  26.01 & 0.832 & 0.298 & 29.18 & 0.923 & 0.218 & 29.97 & 0.905 & 0.235 & 30.55 & 0.910 & 0.225\\
    RESCAN  &ECCV'18& 26.36 & 0.786 & 0.312 & 29.80 & 0.881 & 0.235 & 31.29 & 0.904 & 0.228 & 30.51 & 0.882 & 0.248\\
    PreNet &CVPR'19& 26.77 & 0.858 & 0.285 & 32.44 & 0.950 & 0.182 & 31.75 & 0.916 & 0.215 & 31.36 & 0.911 & 0.218\\
    MSPFN  &CVPR'20 & 28.66 &  0.860 & 0.268 &  32.40 & 0.933 & 0.195 & 32.82 & 0.930 & 0.198 & 32.39 & 0.916 & 0.208\\
    SPAIR  &ICCV'21 & 30.95 & 0.892 & 0.235 & 36.93 & 0.969 & 0.145 & 33.34 & 0.936 & 0.185 & 33.04 & 0.922 & 0.195\\
    MPRNet &CVPR'21& 30.41 & 0.890 & 0.242 & 36.40 & 0.965 & 0.152 & 33.64 & 0.938 & 0.178 & 32.91 & 0.916 & 0.202\\
    
    Restormer  &CVPR'22 & 31.46 &  0.904 & 0.225 & {38.99} & 0.978 & 0.125 & 34.18 & 0.944 & 0.168 & 33.19 & 0.926 & 0.188\\
    Fourmer  &ICML'23 & 30.76 &   0.896 & 0.238 & 37.47 & 0.970 & 0.142 & - & - & - & 33.05 & 0.921 & 0.192\\
    IR-SDE  &ICML'23 & 31.65 &   0.904 & 0.222 & 38.30 & {0.980} & 0.132 & 30.42 & 0.891 & 0.245 & - & - & -\\
    MambaIR  & ECCV'24 & 30.62 &   0.893 & 0.245 & 38.78 & 0.977 & 0.128 & 33.58 & 0.927 & 0.195 & 32.56 & 0.923 & 0.198\\
    VMambaIR  &arxiv'24 & {31.66} &  {0.909} & 0.218 & {39.09} & 0.979 & 0.122 & 34.01 & 0.944 & 0.172 & {33.33} & 0.926 & 0.185\\
    FreqMamba & arxiv'24 & 31.74 & 0.912 & 0.215 & 39.18 & 0.981& 0.118 & \cellcolor{best}34.25 & \cellcolor{secondbest}0.950 & \cellcolor{secondbest}0.165 & 33.36 & 0.931 & 0.182\\
    FourierMamba & arxiv'24 & \cellcolor{secondbest}31.79 & \cellcolor{best}0.913 & \cellcolor{secondbest}0.212 & \cellcolor{secondbest}39.73 &\cellcolor{best}0.986 & \cellcolor{secondbest}0.115 & \cellcolor{secondbest}34.23 & 0.949 & 0.168 & \cellcolor{secondbest}34.76 & \cellcolor{best}0.938 & \cellcolor{secondbest}0.175\\
    \midrule[0.1em]
    \textbf{VAMamba (Ours)} & - & \cellcolor{best}31.81 & \cellcolor{best}0.913 & \cellcolor{best}0.210 & \cellcolor{best}39.90 & \cellcolor{secondbest}0.981 & \cellcolor{best}0.112 & 34.10 & \cellcolor{best}0.951 & \cellcolor{best}0.162 & \cellcolor{best}34.80 & \cellcolor{secondbest}0.934 & \cellcolor{best}0.172\\
    \bottomrule[0.15em]
    \end{tabular}%
    }
  \vspace{-0mm}
\end{table*}

\begin{table*}[!htb]
\begin{center}
\caption{ \underline{\textbf{Single-image motion deblurring}} results. Our VAMamba is trained only on the GoPro dataset~\cite{gopro2017} and directly applied to the HIDE~\cite{shen2019human} and RealBlur~\cite{rim_2020_realblur} benchmark datasets.}
\label{tab: deblur}
\vspace{-1mm}
\setlength{\tabcolsep}{4.5pt}
\scalebox{0.95}{
\begin{tabular}{l |c |c c c | c c c | c c c | c c c }
\toprule[0.15em]
 && \multicolumn{3}{c|}{\textbf{GoPro}} & \multicolumn{3}{c|}{\textbf{HIDE}} & \multicolumn{3}{c|}{\textbf{RealBlur-R}} & \multicolumn{3}{c}{\textbf{\textbf{RealBlur-J}}} \\
 \textbf{Method} & \textbf{Venue} & PSNR & SSIM & LPIPS & PSNR & SSIM & LPIPS & PSNR & SSIM & LPIPS & PSNR & SSIM & LPIPS\\
\midrule[0.15em]
Nah et al.  & CVPR'17 & 29.08 & 0.914 & 0.285 & 25.73 & 0.874 & 0.324 &  32.51 & 0.841 & 0.298 &  27.87 & 0.827 & 0.335\\
MIMO-UNet+ &ICCV'21 & 32.45 & 0.957 & 0.182 & 29.99 & 0.930 & 0.215 & 35.54 & 0.947 & 0.165 & 27.63 & 0.837 & 0.318\\
MPRNet &CVPR'21 & 32.66 & 0.959 & 0.175 &30.96 & 0.939 & 0.198 & 35.99 & 0.952 & 0.158 & 28.70 & 0.873 & 0.285\\
Restormer&CVPR'22 & 32.92 & 0.961 & 0.168 & 31.22 & 0.942 & 0.188 & 36.19 & 0.957 & 0.152 & 28.96 & 0.879 & 0.275\\
Stripformer &ECCV'22&33.08 & 0.962 & 0.162& 31.03 & 0.940 & 0.192&39.84 & 0.974 & \cellcolor{best}0.098&32.48 & 0.929 & \cellcolor{secondbest}0.155\\
FFTformer &CVPR'23&\cellcolor{secondbest}34.21 & \cellcolor{secondbest}0.969 & \cellcolor{secondbest}0.145&\cellcolor{secondbest}31.62 & \cellcolor{secondbest}0.945 & \cellcolor{secondbest}0.178&\cellcolor{secondbest}40.11 & \cellcolor{secondbest}0.973 & \cellcolor{secondbest}0.102& \cellcolor{secondbest}32.62 & \cellcolor{secondbest}0.932 & \cellcolor{best}0.148\\
MambaIR& ECCV'24&30.12 & 0.945 & 0.235 & 29.22& 0.924 & 0.248 & 35.91 & 0.949 & 0.172 & 26.99 & 0.856 & 0.298\\
\bottomrule[0.1em]
\textbf{VAMamba (Ours)} &-& \cellcolor{best}33.95& \cellcolor{best}0.970 & \cellcolor{best}0.138 & \cellcolor{best}31.38 & \cellcolor{best}0.948 & \cellcolor{best}0.172 & \cellcolor{best}39.95 & \cellcolor{best}0.960 & 0.105 & \cellcolor{best}32.35 & 0.903 & 0.162\\
\bottomrule[0.1em]
\end{tabular}}
\end{center}\vspace{0mm}
\end{table*}

\begin{table*}[!t]
\begin{center}
\caption{ \underline{\textbf{Defocus deblurring}} comparisons on the DPDD testset (containing 37 indoor and 39 outdoor scenes).
}
\label{tab: dpdd}
\setlength{\tabcolsep}{6pt}
\scalebox{0.92}{
\begin{tabular}{l |c | c c c | c c c | c c c }
\toprule[0.15em]
   & & \multicolumn{3}{c|}{\textbf{Indoor Scenes}} & \multicolumn{3}{c|}{\textbf{Outdoor Scenes}} & \multicolumn{3}{c}{\textbf{Combined}} \\
   \textbf{Method} & \textbf{Venue} & PSNR & SSIM & LPIPS & PSNR & SSIM & LPIPS & PSNR & SSIM & LPIPS \\
\midrule[0.15em]
DMENet & ICCV'19 & 25.50 & 0.788 & 0.298 & 21.43 & 0.644 & 0.397 & 23.41 & 0.714 & 0.349 \\
DPDNet & CVPR'20 & 26.54 & 0.816 & 0.239 & 22.25 & 0.682 & 0.313 & 24.34 & 0.747 & 0.277\\
KPAC$_S$ & CVPR'21 & 27.97 & 0.852 & 0.182 & 22.62 & 0.701 & 0.269 & 25.22 & 0.774 & 0.227 \\
IFAN & CVPR'21 & 28.11 & 0.861 & 0.179 & 22.76 & 0.720 & 0.254 & 25.37 & 0.789 & 0.217\\
Restormer& CVPR'22 & \cellcolor{secondbest}28.87 & \cellcolor{secondbest}0.882 & \cellcolor{secondbest}0.145 & \cellcolor{secondbest}23.24 & \cellcolor{secondbest}0.743 & \cellcolor{secondbest}0.209 & \cellcolor{secondbest}25.98 & \cellcolor{secondbest}0.811 & \cellcolor{secondbest}0.178   \\
\midrule[0.15em]
\textbf{VAMamba (Ours)} & - & \cellcolor{best}29.17 & \cellcolor{best}0.912 & \cellcolor{best}0.115 & \cellcolor{best}25.14 & \cellcolor{best}0.803 & \cellcolor{best}0.181 & \cellcolor{best}27.16 & \cellcolor{best}0.858 & \cellcolor{best}0.148\\
\bottomrule[0.1em]
\end{tabular}}
\end{center}
\vspace{0mm}
\end{table*}

\begin{table*}[!htb]
\begin{center}
\caption{ \underline{\textbf{Image dehazing results}} on the SOTS-Indoor, SOTS-Outdoor, and Haze4K datasets. VAMamba achieves competitive performance across all tasks.}
\label{tab:dehaze}
\vspace{-1mm}
\setlength{\tabcolsep}{6pt}
\scalebox{0.92}{
\begin{tabular}{l|c|ccc|ccc|ccc}
\toprule[0.15em]
 & & \multicolumn{3}{c|}{\textbf{SOTS-Indoor}} & \multicolumn{3}{c|}{\textbf{SOTS-Outdoor}} & \multicolumn{3}{c}{\textbf{Haze4K}} \\
\textbf{Method} & \textbf{Venue} & PSNR & SSIM & LPIPS & PSNR & SSIM & LPIPS & PSNR & SSIM & LPIPS \\
\midrule[0.15em]
DCP & CVPR'09 & 16.62 & 0.818 & 0.485 & 24.75 & 0.927 & 0.298 & 14.01 & 0.760 & 0.542 \\
GridDehazeNet & ICCV'19 & 32.16 & 0.984 & 0.245 & 30.08 & 0.982 & 0.182 & 23.29 & 0.930 & 0.285 \\
FFA-Net & AAAI'20 & 36.39 & 0.989 & 0.165 & 33.57 & 0.984 & 0.142 & 27.06 & 0.950 & 0.195 \\
DeHamer & CVPR'22 & 36.63 & 0.988 & 0.162 & 35.18 & 0.983 & 0.135 & 28.52 & 0.947 & 0.188 \\
DehazeFormer-L & TIP'23 & 38.46 & 0.996 & 0.128 & 36.19 & 0.988 & 0.118 & 30.89 & 0.977 & 0.142 \\
MambaIR & ECCV'24 & \cellcolor{secondbest}40.58 & 0.995 & \cellcolor{secondbest}0.112 & 35.41 & 0.986 & 0.125 & \cellcolor{secondbest}31.24 & \cellcolor{secondbest}0.979 & \cellcolor{secondbest}0.135 \\
\midrule[0.15em]
\textbf{VAMamba (Ours)} & - & \cellcolor{best}40.63 & \cellcolor{best}0.996 & \cellcolor{best}0.108 & \cellcolor{best}36.48 & \cellcolor{best}0.988 & \cellcolor{best}0.115 & \cellcolor{best}31.33 & \cellcolor{best}0.980 & \cellcolor{best}0.132 \\
\bottomrule[0.15em]
\end{tabular}}
\end{center}
\vspace{0mm}
\end{table*}

\begin{table*}[!htb]
\begin{center}
\caption{\underline{\textbf{Low-light enhancement}} results on LOL-v1~\cite{we2018deep} and LOL-v2-real~\cite{yang2021sparse} datasets. VAMamba achieves superior performance across all metrics.}
\label{tab:lol_enhancement}
\setlength{\tabcolsep}{12pt}
\scalebox{0.96}{
\begin{tabular}{l | c | ccc | ccc }
\toprule[0.15em]
\multirow{2}{*}{\textbf{Method}} & \multirow{2}{*}{\textbf{Venue}} & \multicolumn{3}{c|}{\textbf{LOL-v1}} & \multicolumn{3}{c}{\textbf{LOL-v2-real}} \\
& & \textbf{PSNR} & \textbf{SSIM} & \textbf{LPIPS} & \textbf{PSNR} & \textbf{SSIM} & \textbf{LPIPS} \\
\midrule[0.15em]
Retinex-Net & BMVC'18 & 16.77 & 0.560 & 0.381 & 18.37 & 0.723 & 0.425 \\
KinD & ACM MM'19 & 20.86 & 0.808 & 0.170 & 19.74 & 0.764 & 0.300 \\
Zero-DCE & CVPR'20 & 14.83 & 0.531 & 0.431 & 15.52 & 0.701 & 0.430 \\
EnlightenGAN & TIP'21 & 17.48 & 0.650 & 0.322 & 17.88 & 0.725 & 0.407 \\
DRBN & TIP'21 & 19.55 & 0.746 & 0.316 & 20.02 & 0.730 & 0.304 \\
Retformer & ICCV'23 & \cellcolor{secondbest}22.43 & \cellcolor{secondbest}0.859 & 0.112 & 21.12 & \cellcolor{secondbest}0.847 & 0.292 \\
MambaIR & ECCV'24 & 24.05 & 0.857 & 0.113 & \cellcolor{secondbest}22.11 & 0.834 & \cellcolor{secondbest}0.262 \\
\midrule[0.15em]
\textbf{VAMamba (Adaptive scanning)} & - & \cellcolor{best}24.30 & \cellcolor{best}0.865 & \cellcolor{best}0.096 & \cellcolor{best}22.50 & \cellcolor{best}0.860 & \cellcolor{best}0.230 \\
\bottomrule[0.15em]
\end{tabular}}
\end{center}
\vspace{0mm}
\end{table*}

\begin{table*}[!htb]
\begin{center}
\caption{ \underline{\textbf{Performance comparison of different scanning methods}} on the dehazing task. VAMamba outperforms traditional scanning methods in PSNR, SSIM, and efficiency.}
\label{tab:scanning_comparison}
\vspace{-1mm}
\setlength{\tabcolsep}{3.5pt}
\scalebox{1.1}{
\begin{tabular}{l|c|c|c|c|c}
\toprule[0.15em]
\textbf{Method} & \textbf{PSNR} & \textbf{SSIM} & \textbf{FLOPs (G)} & \textbf{Params (M)} & \textbf{Inference Time (s)}  \\
\midrule[0.15em]
Vim (Two-way scanning)    & 39.99     & 0.993  & 27.33 & 6.55  & 0.35    \\
Vmamba (Four-way scanning)  & 40.22     & 0.990 & 28.13 & 6.33 & 0.32   \\
PlainMamba (Snake-shaped scanning) & 40.10     & 0.994  & 28.11 & 6.59 & 0.29    \\
LocalMamba (Local scanning) & 39.88     & 0.991  & 26.39 & \textbf{6.42} & \textbf{0.28}    \\
\rowcolor{color4}
\textbf{VAMamba (Ours)} & \textbf{40.63}     & \textbf{0.996}  & 27.23 & 6.43 & 0.29   \\
\bottomrule[0.15em]
\end{tabular}}
\end{center}
\vspace{-2em}
\end{table*}

\subsection{Adaptive Scanning vs. Traditional Scanning}
To validate the effectiveness of our GPS-SS2D adaptive scanning mechanism, we conduct experiments using the dehazing task, comparing our method with traditional scanning approaches in existing Mamba models, including two-way scanning (Vim\cite{visionmamba}), four-way scanning (Vmamba\cite{liu2024vmamba}), snake-shaped scanning (PlainMamba\cite{Yang_2024_BMVC_plainmamba}), and local scanning (LocalMamba\cite{huang2024localmamba}).

\begin{figure*}[!htb]
\centering

\begin{subfigure}[t]{0.095\textwidth}
    \centering
    \includegraphics[width=\textwidth, height=5.8cm, keepaspectratio]{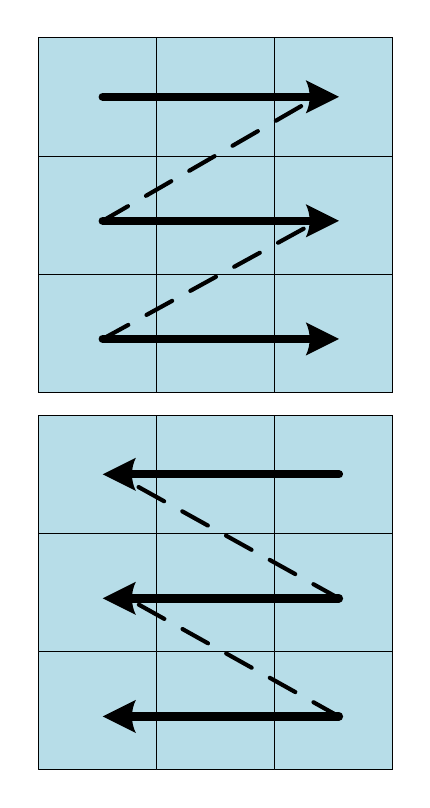}
    \caption*{\hspace{0.5cm}(a)Vim}
    \label{fig:vim_scanning}
\end{subfigure}
\hfill
\begin{subfigure}[t]{0.18\textwidth}
    \centering
    \includegraphics[width=\textwidth]{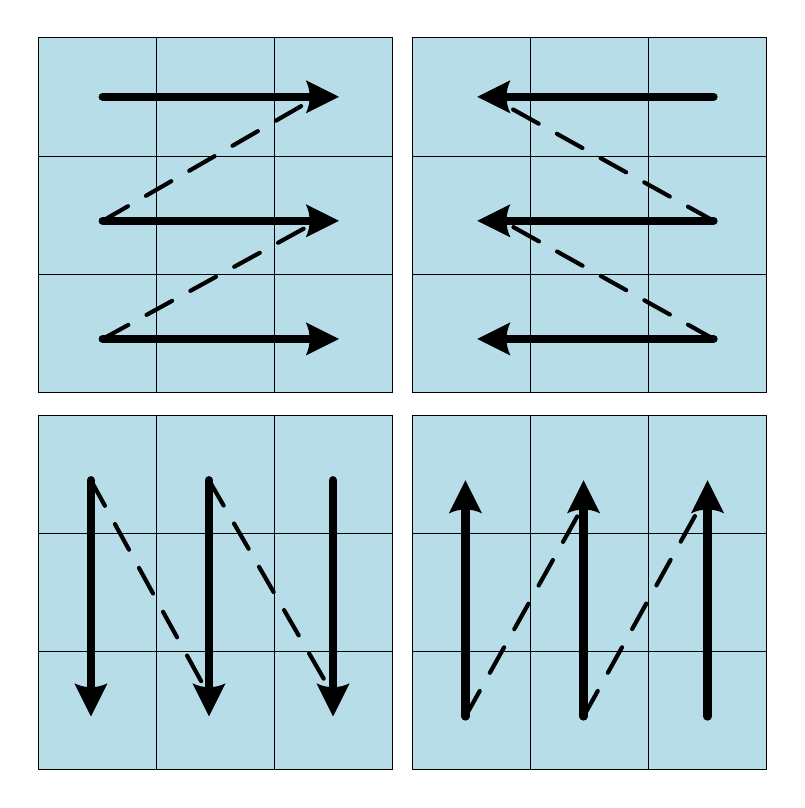}
    \caption*{\hspace{1.1cm}(b)Vmamba}
    \label{fig:vmamba_scanning}
\end{subfigure}
\hfill
\begin{subfigure}[t]{0.18\textwidth}
    \centering
    \includegraphics[width=\textwidth]{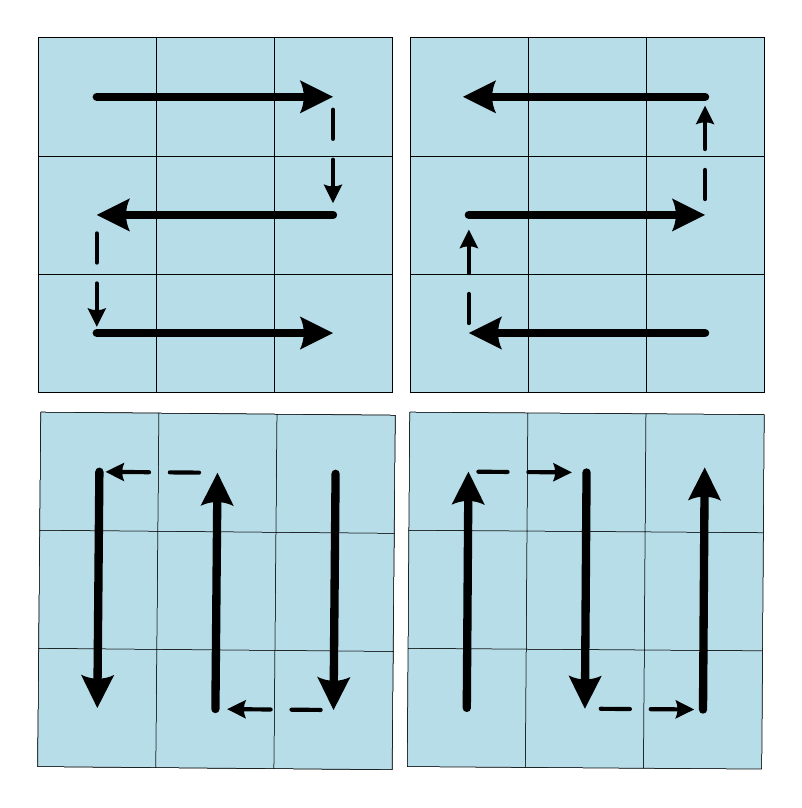}
    \caption*{\hspace{0.7cm}(c)PlainMamba}
    \label{fig:plainmamba_scanning}
\end{subfigure}
\hfill
\begin{subfigure}[t]{0.18\textwidth}
    \centering
    \includegraphics[width=\textwidth]{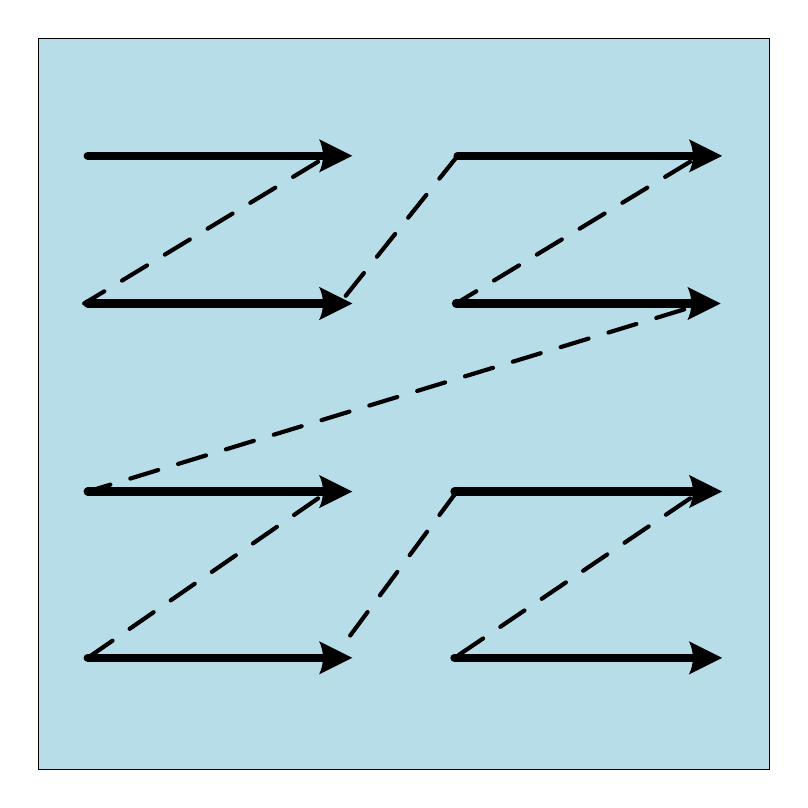}
    \caption*{\hspace{0.6cm}(d)LocalMamba}
    \label{fig:localmamba_scanning}
\end{subfigure}
\hfill
\begin{subfigure}[t]{0.18\textwidth}
    \centering
    \includegraphics[width=\textwidth]{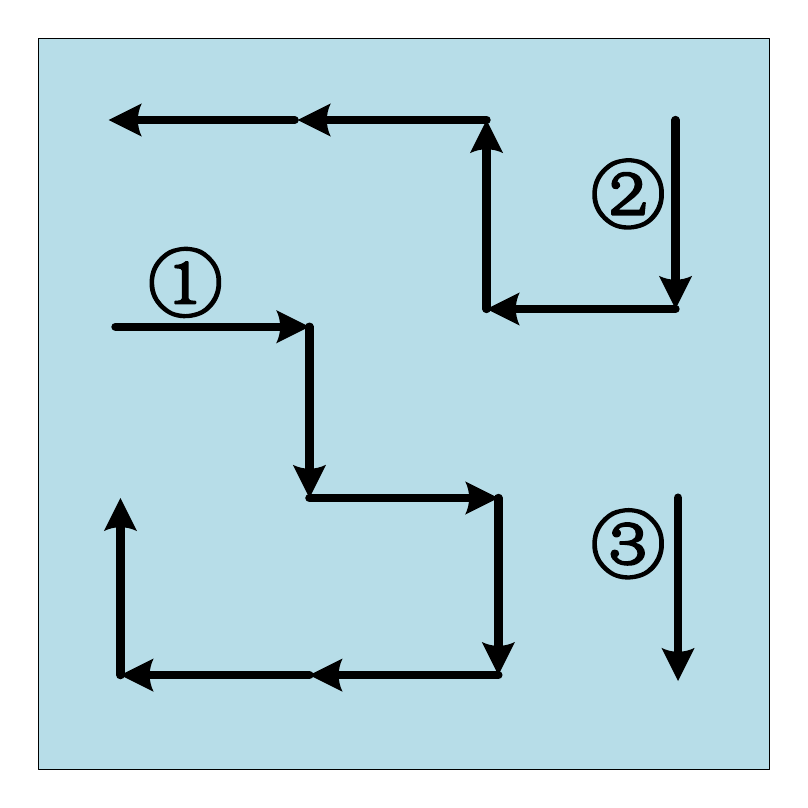}
    \caption*{\hspace{0.5cm}(e)VAMamba (Ours)}
    \label{fig:vamamba_scanning}
\end{subfigure}
\caption{Comparison of different scanning strategies in Mamba-based models. (a) \textbf{Vim}: Two-way scanning with horizontal passes followed by vertical passes. (b) \textbf{Vmamba}: Four-way scanning processing input through left-to-right, top-to-bottom, right-to-left, and bottom-to-top directions. (c) \textbf{PlainMamba}: Snake-shaped scanning traversing in serpentine pattern with alternating row directions. (d) \textbf{LocalMamba}: Localized scanning within predefined regions, processing independent blocks. (e) \textbf{VAMamba (Ours)}: Content-aware adaptive scanning using GPS-SS2D with ViT-generated importance scores to dynamically determine optimal scanning paths, prioritizing degraded regions through greedy path selection.}
\label{fig:scanning_ways}
\vspace{-2mm}
\end{figure*}

\noindent\textbf{Limitations of Traditional Scanning Methods.}As illustrated in Fig.~\ref{fig:scanning_ways}(a-d) Traditional scanning methods in image restoration suffer from inherent design limitations: two-way and four-way scanning use fixed horizontal or vertical paths, leading to redundant computations and lack of flexibility when handling complex degradations; snake-shaped scanning provides better spatial coverage but still cannot adapt to content-specific degradation characteristics; local scanning improves computational efficiency but cannot dynamically adjust scanning paths, potentially missing important global contextual information.

\noindent\textbf{Advantages of GPS-SS2D.}As demonstrated in Fig.~\ref{fig:scanning_ways}(e), Our GPS-SS2D method identifies regions with varying degradation severity using ViT-generated importance score maps and plans optimal scanning trajectories through greedy path selection. The numbered sequence (1→2→3) in the figure illustrates how our method prioritizes high-importance regions while maintaining spatial locality when possible. This adaptive approach offers three key advantages: (1) Content-aware processing: prioritizes regions based on their restoration importance for optimal resource allocation; (2) Dynamic path optimization: ensures sequential processing of neighboring high-importance regions; (3) Global-local balance: maintains global contextual awareness while focusing on locally important regions.

\noindent\textbf{Experimental Results.}As shown in Table.~\ref{tab:scanning_comparison}, GPS-SS2D achieves 40.63 dB PSNR and 0.996 SSIM, outperforming the best traditional method Vmamba (40.22 dB PSNR, 0.990 SSIM). Meanwhile, our method maintains good computational efficiency with only 0.29s inference time and 6.43M parameters, comparable to traditional approaches. This demonstrates that adaptive scanning achieves performance improvements through intelligent resource allocation and reduced redundant computations without increasing model complexity.

\subsection{Path Visualization}
\begin{figure*}[!htb]
    \centering
    \begin{adjustbox}{max width=\textwidth}
    \begin{tabular}{c c c c c}
        \includegraphics[width=0.24\textwidth]{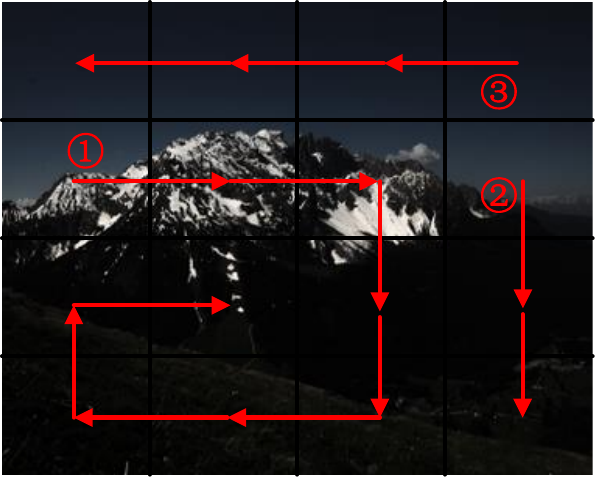} &
        \includegraphics[width=0.24\textwidth]{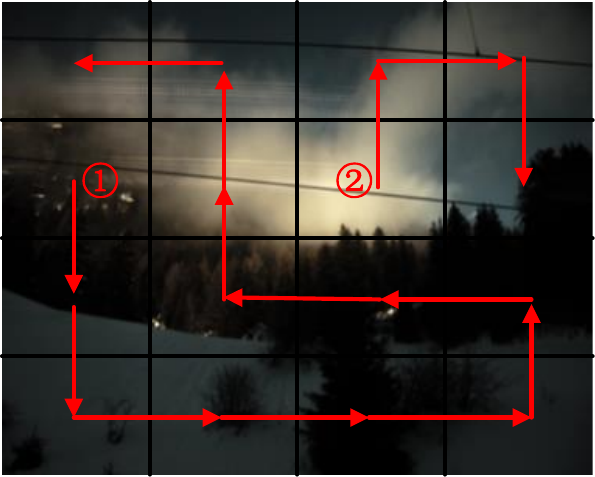} &
        \includegraphics[width=0.24\textwidth]{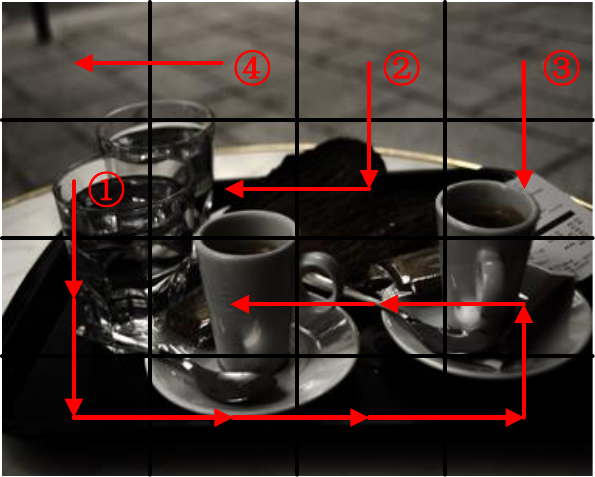} &
        \includegraphics[width=0.24\textwidth]{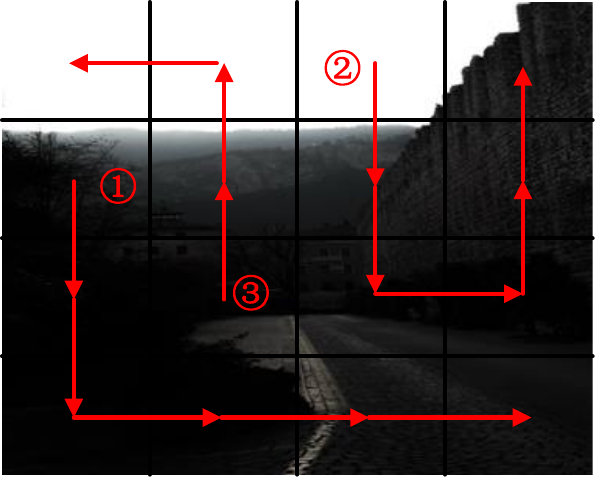} &
        \includegraphics[width=0.24\textwidth]{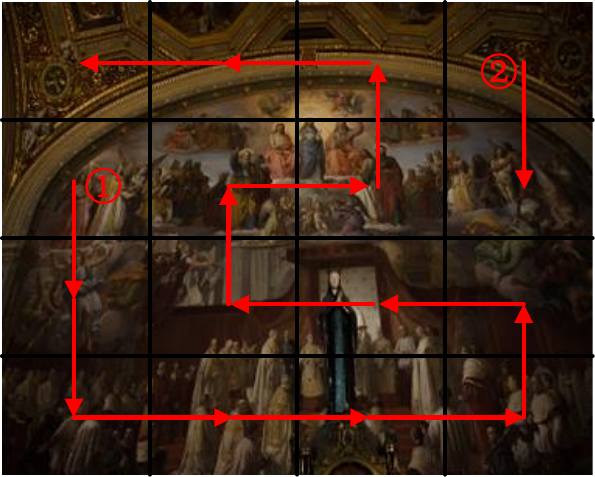}
    \end{tabular}
    \end{adjustbox}
    \vspace{0.5em}

    \begin{adjustbox}{max width=\textwidth}
    \begin{tabular}{c c c c c}
        \includegraphics[width=0.24\textwidth]{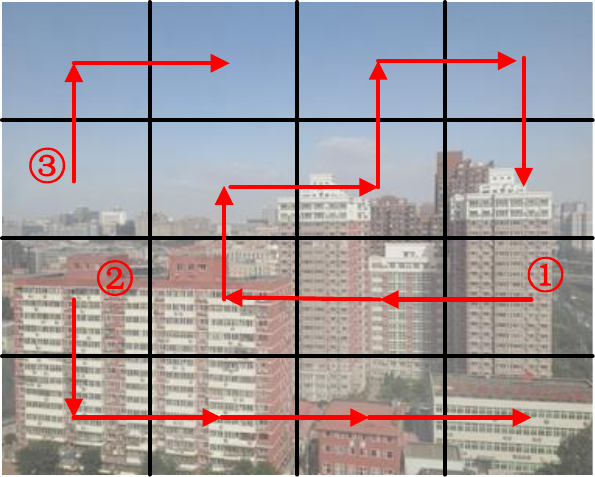} &
        \includegraphics[width=0.24\textwidth]{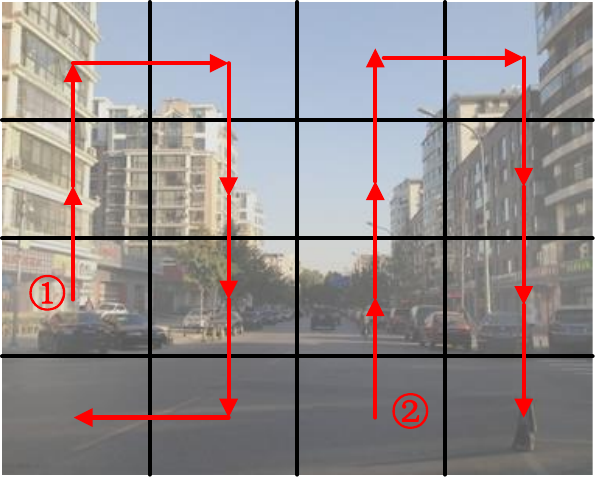} &
        \includegraphics[width=0.24\textwidth]{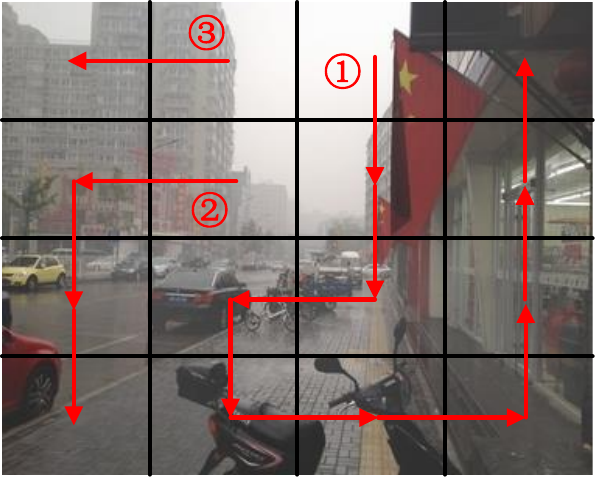} &
        \includegraphics[width=0.24\textwidth]{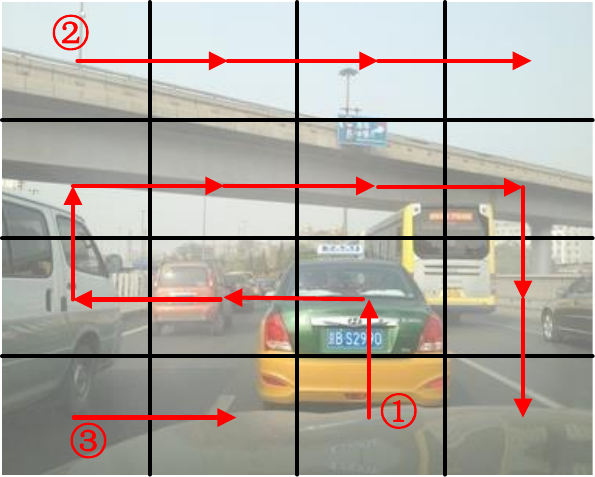} &
        \includegraphics[width=0.24\textwidth]{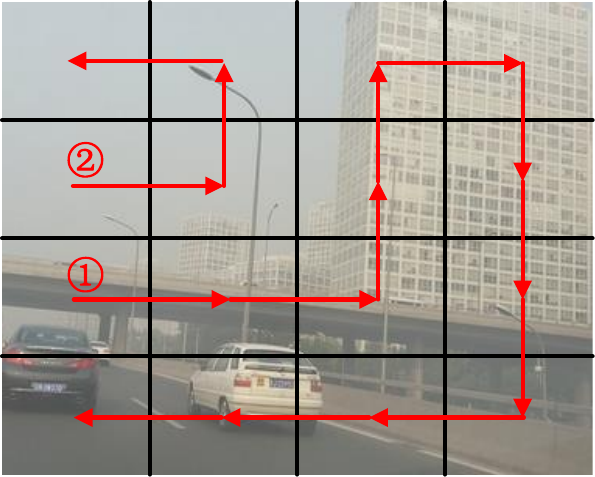}
    \end{tabular}
    \end{adjustbox}
    \vspace{0.5em}

    \begin{adjustbox}{max width=\textwidth}
    \begin{tabular}{c c c c c}
        \includegraphics[width=0.24\textwidth]{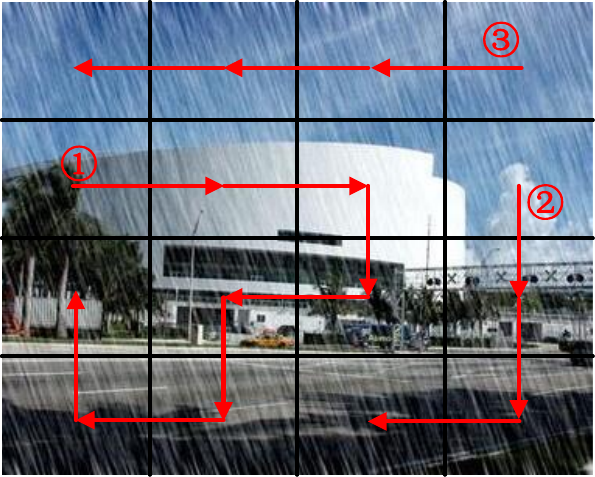} &
        \includegraphics[width=0.24\textwidth]{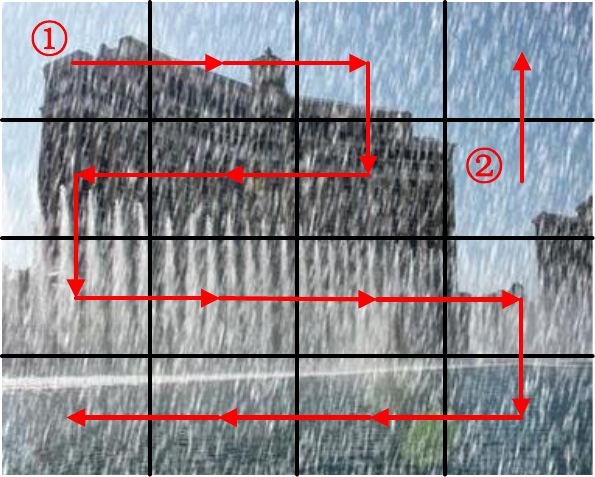} &
        \includegraphics[width=0.24\textwidth]{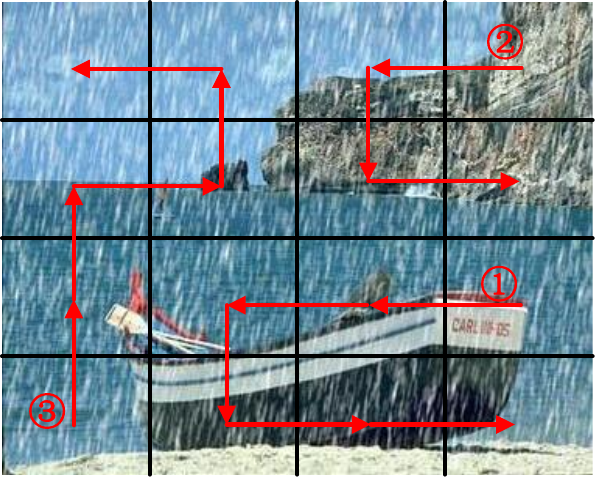} &
        \includegraphics[width=0.24\textwidth]{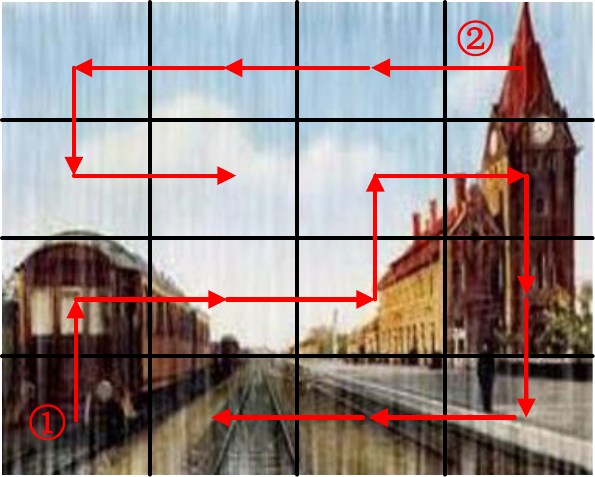} &
        \includegraphics[width=0.24\textwidth]{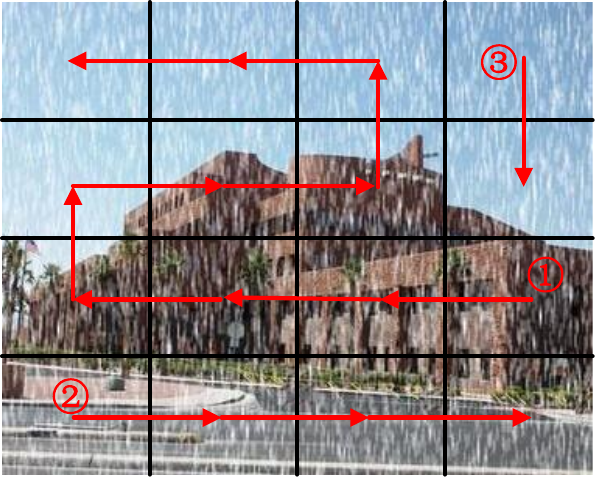}

        \\
        
    \end{tabular}
    \end{adjustbox}
    \caption{Path visualization across three restoration tasks (top to bottom: low-light enhancement, dehazing, and deraining). Red polylines and arrows denote the GPS-SS2D scan trajectories planned from patch-wise importance, with numbers indicating the visiting order (smaller = earlier); the grid shows the patch partition. }
    \label{fig:path}
\end{figure*}

To verify the effectiveness of our dynamic adaptive scanning strategy across different image degradation tasks, we conduct path-visualization experiments on low-light enhancement, dehazing, and deraining. As shown in Fig.~\ref{fig:path} By inspecting the resulting scan trajectories, we can intuitively reveal how the model prioritizes where to scan under different degradation types. Unlike traditional methods with fixed scan patterns, our approach adjusts the trajectory according to the specific degradation characteristics of each image, enabling more efficient and precise feature extraction
The visualizations show a clear prioritization behavior in the learned paths. The model first scans the main subject region, ensuring that critical information—such as foreground objects and fine details—is restored early, while background areas are deferred to later stages of the scan. For example, in low-light images the model starts with dark regions that typically contain recoverable details; in dehazing it first focuses on subject areas heavily affected by haze; and in deraining it prioritizes regions strongly occluded by raindrops. In contrast, background regions are generally handled later, which avoids redundant computation and leads to a more effective allocation of computational resources
The advantage of this strategy is its ability to adapt the scanning path to the image’s degradation pattern, concentrating on informative regions rather than following a fixed template. The path visualizations make it clear that, across low-light, dehazing, and deraining, the model performs effective adaptive scanning that improves restoration quality. This strategy not only enhances fidelity but also reduces redundant computation, highlighting the superiority of dynamic adaptive scanning.
%
\setlength{\tabcolsep}{1pt}
\def \imgl {0.14\linewidth}
\def \imgs {0.14\linewidth}
\begin{figure*}[!htb]
    \small
    \begin{center}
    \begin{tabular}{cccccc}
        \includegraphics[width=\imgl]{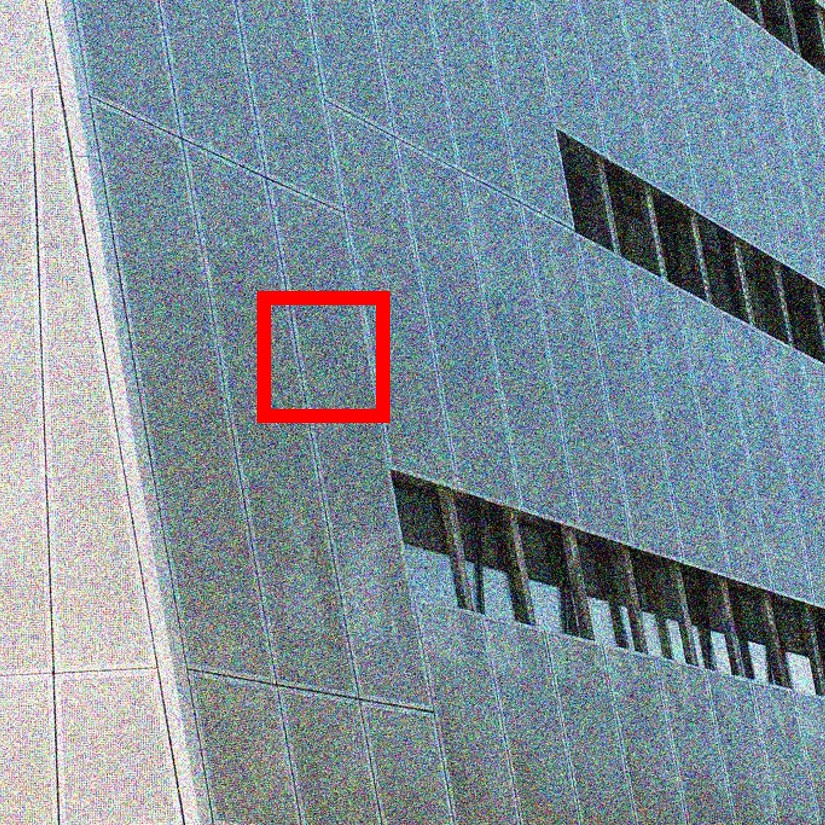} &
        \includegraphics[width=\imgs]{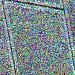} &
        \includegraphics[width=\imgs]{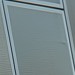} &
        \includegraphics[width=\imgs]{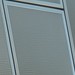} &
        \includegraphics[width=\imgs]{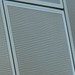} &
        \includegraphics[width=\imgs]{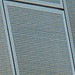} \\
        \includegraphics[width=\imgs,height=0.15\linewidth]{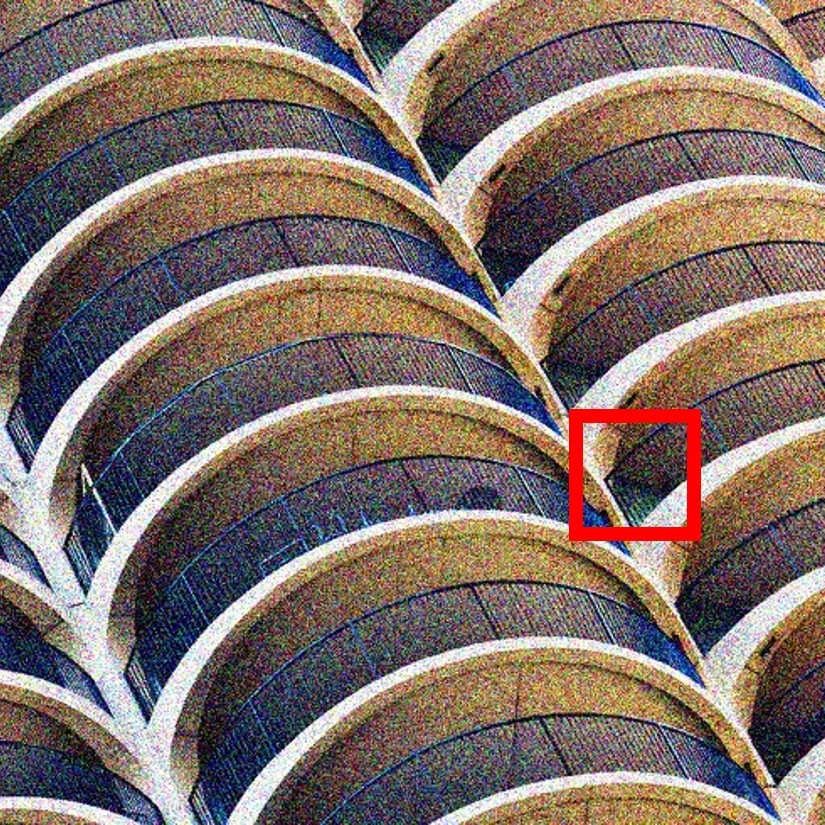} &
        \includegraphics[width=\imgs]{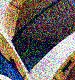} &
        \includegraphics[width=\imgs]{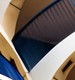} &
        \includegraphics[width=\imgs]{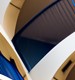} &
        \includegraphics[width=\imgs]{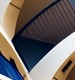} &
        \includegraphics[width=\imgs]{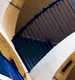} \\
        Full Image & Noisy & MambaIR & MambaIRV2 & Ours & GT \\
    \end{tabular}
    \end{center}\vspace{-2mm}
    \caption{Visualization results on Gaussian denoising of color images.
    }\vspace{-2mm}
    \label{fig: denoise}
\end{figure*}

\begin{figure*}[!htb]
\centering
\scalebox{0.94}{
  \begin{tabular}{ccccccc}
\includegraphics[width=\imgs]{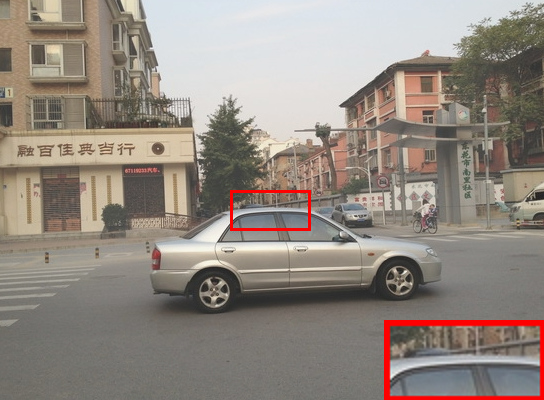} &
\includegraphics[width=\imgs]{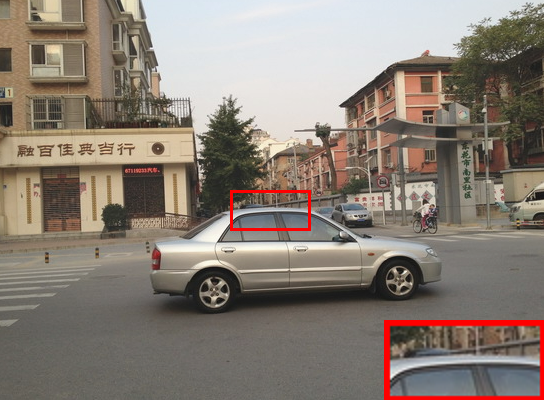} &
\includegraphics[width=\imgs]{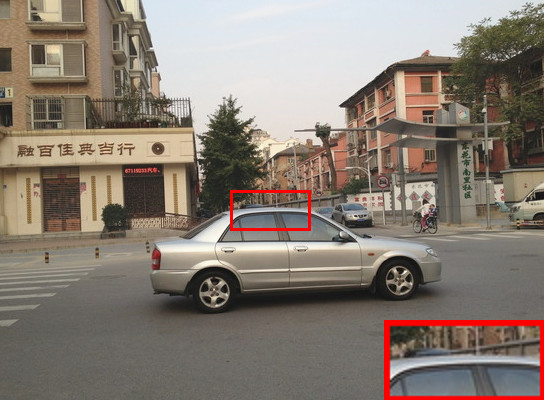} &
\includegraphics[width=\imgs]{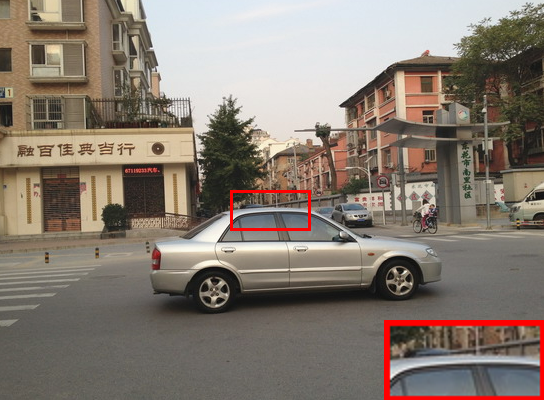} &
\includegraphics[width=\imgs]{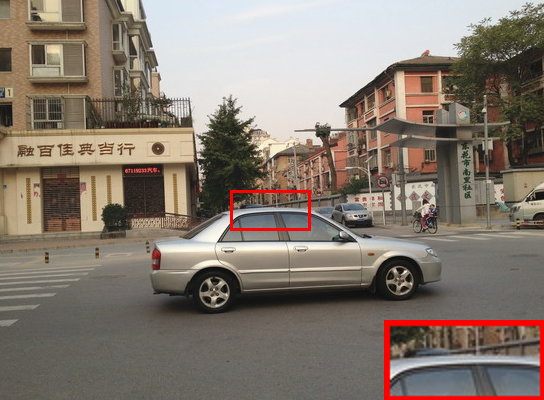} &
\includegraphics[width=\imgs]{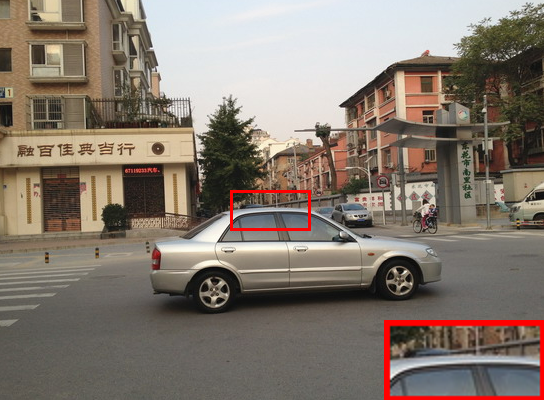} &
\includegraphics[width=\imgs]{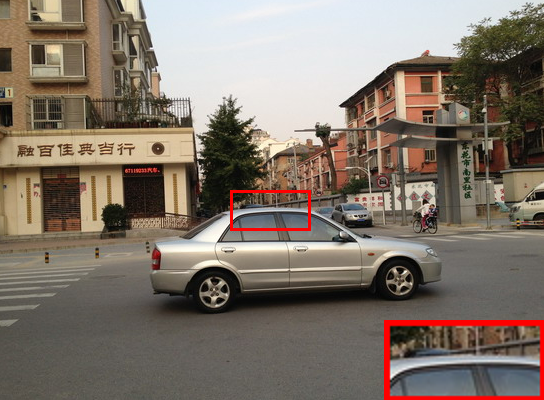}
\\
\includegraphics[width=\imgs]{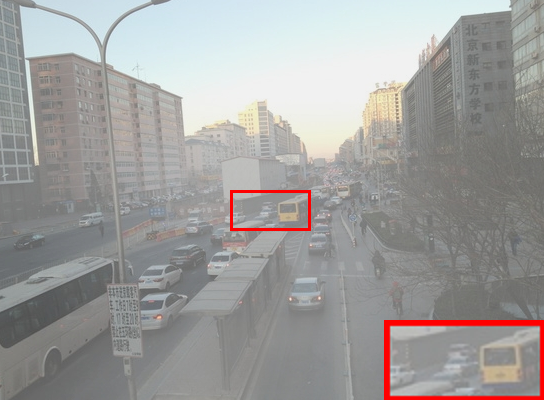} &
\includegraphics[width=\imgs]{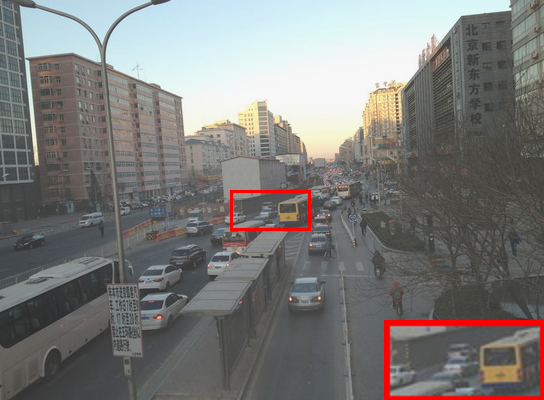} &
\includegraphics[width=\imgs]{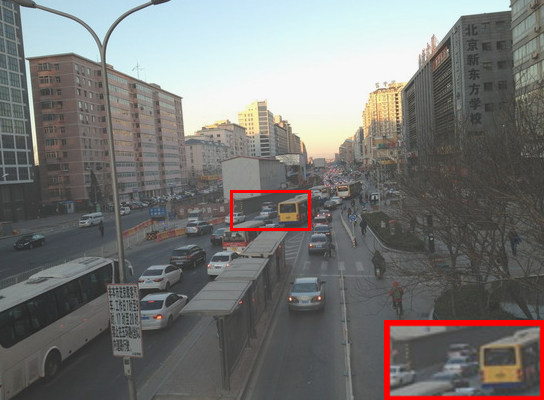} &
\includegraphics[width=\imgs]{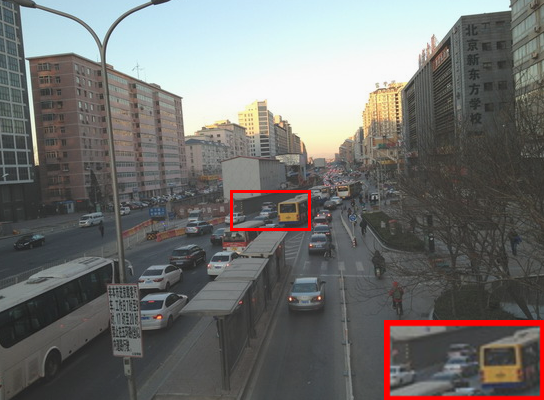} &
\includegraphics[width=\imgs]{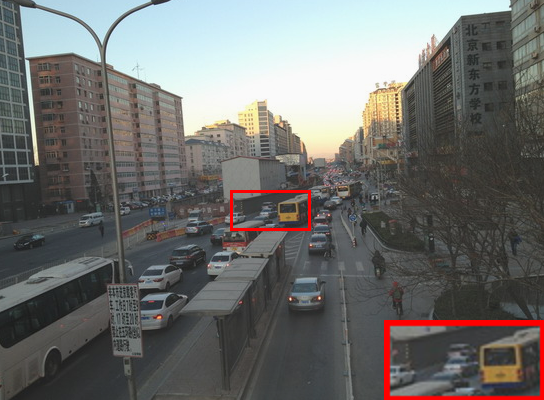} &
\includegraphics[width=\imgs]{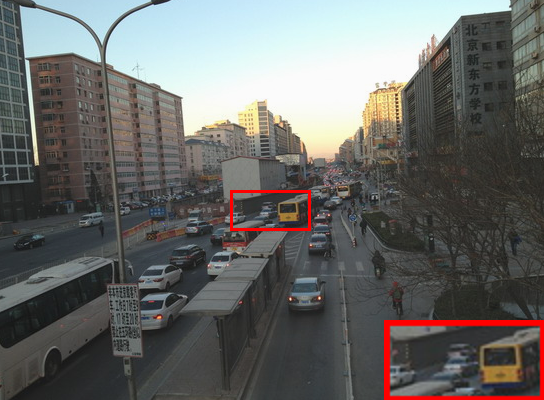} &
\includegraphics[width=\imgs]{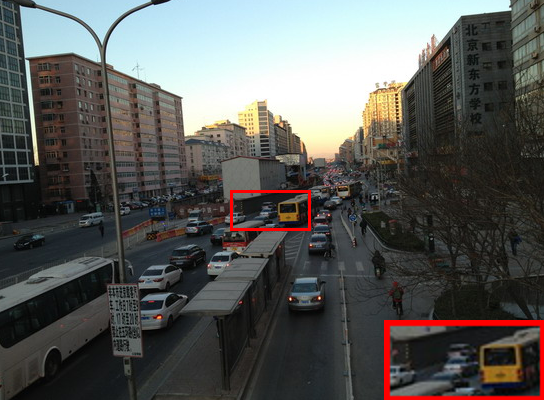} \\
\small Input & GridDehazeNet & FFA-Net & DehazeFormer & MambaIR & Ours & GT \\
\end{tabular}
}
\caption{Visual comparisons on a synthetic hazy image sampled from SOTS-Outdoor dataset. The red boxes highlight specific areas, and zoomed-in details are displayed at the bottom right of each image, showing that our method achieves superior dehazing performance.}
\label{fig:ohaze_comparison_outdoor}
\end{figure*}
%


\begin{figure*}[!htb]
\centering
\setlength{\tabcolsep}{1.5pt}
\renewcommand{\arraystretch}{1.5}
\begin{tabular}{cccccccccccc}
    \includegraphics[width=0.125\textwidth]{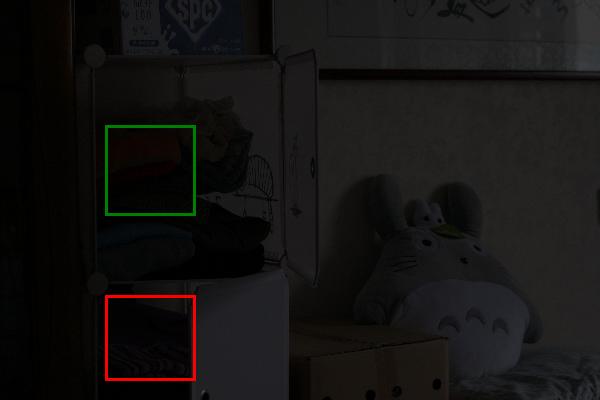} &
    \includegraphics[width=0.125\textwidth]{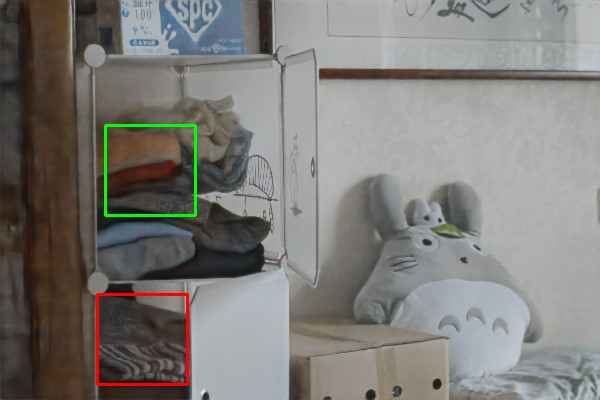} &
    \includegraphics[width=0.125\textwidth]{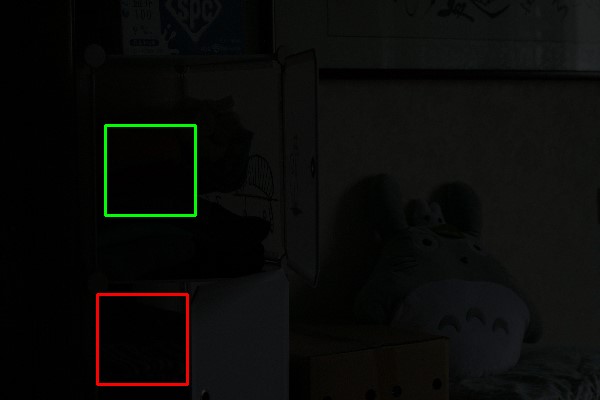} &
    \includegraphics[width=0.125\textwidth]{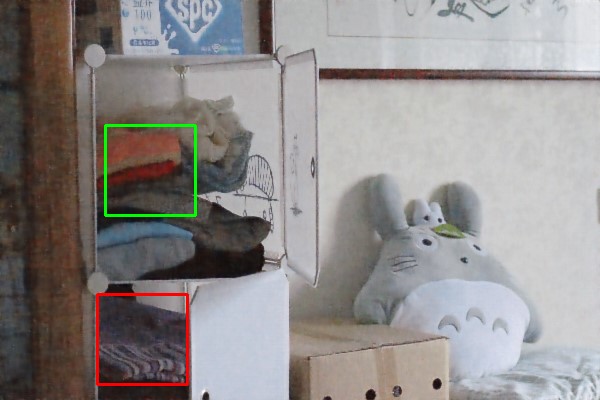} &
    \includegraphics[width=0.125\textwidth]{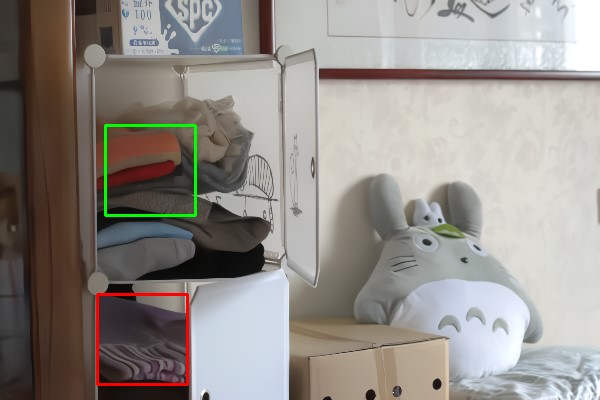} &
    \includegraphics[width=0.125\textwidth]{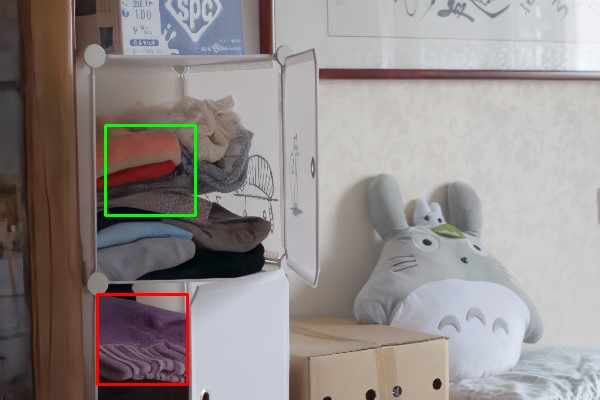} &
    \includegraphics[width=0.125\textwidth]{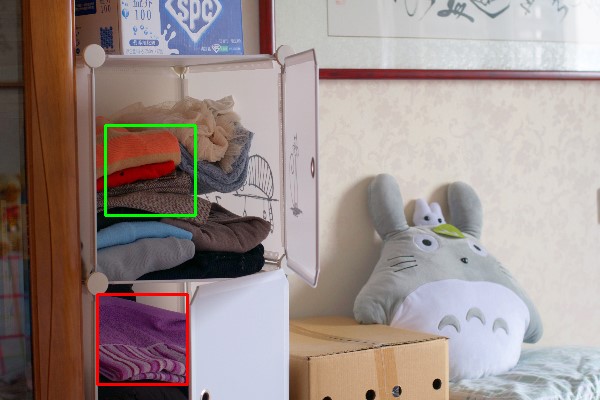} \\
    \includegraphics[width=0.06\textwidth]{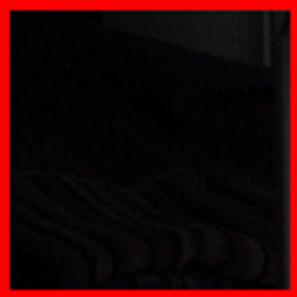} 
    \includegraphics[width=0.06\textwidth]{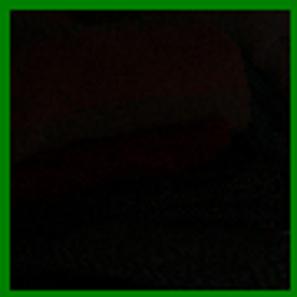} &
    \includegraphics[width=0.06\textwidth]{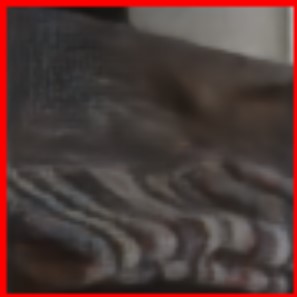} 
    \includegraphics[width=0.06\textwidth]{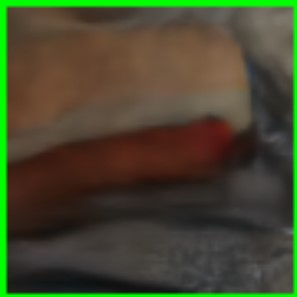} & 
    \includegraphics[width=0.06\textwidth]{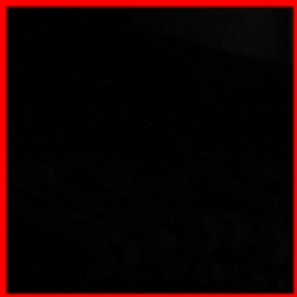} 
    \includegraphics[width=0.06\textwidth]{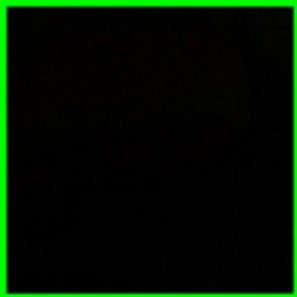} &
    \includegraphics[width=0.06\textwidth]{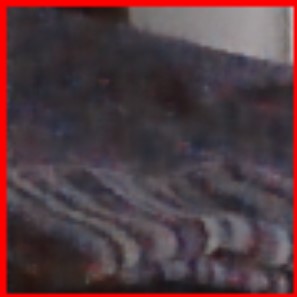} 
    \includegraphics[width=0.06\textwidth]{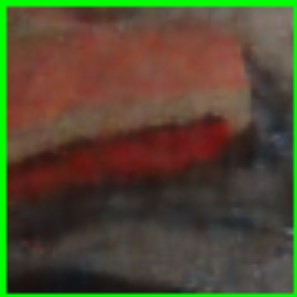} &
    \includegraphics[width=0.06\textwidth]{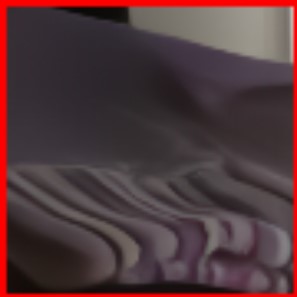}
    \includegraphics[width=0.06\textwidth]{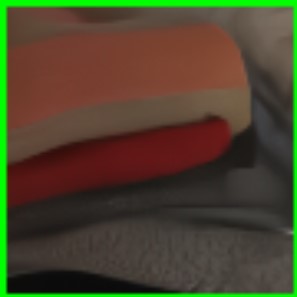}&
    \includegraphics[width=0.06\textwidth]{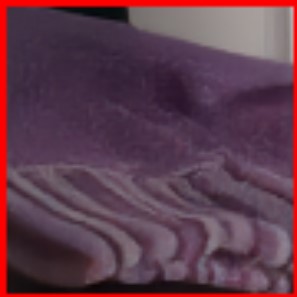}
    \includegraphics[width=0.06\textwidth]{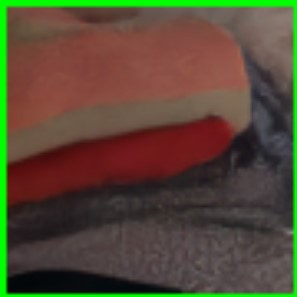}&
    \includegraphics[width=0.06\textwidth]{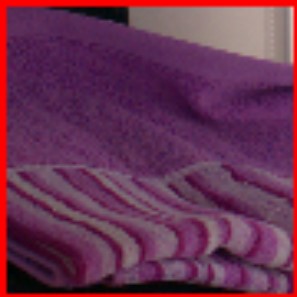}
    \includegraphics[width=0.06\textwidth]{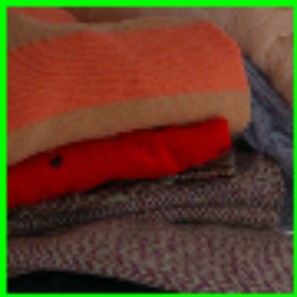}\\
        \includegraphics[width=0.125\textwidth]{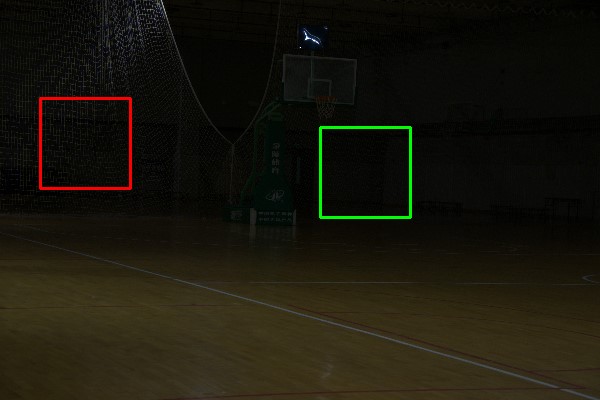} &
    \includegraphics[width=0.125\textwidth]{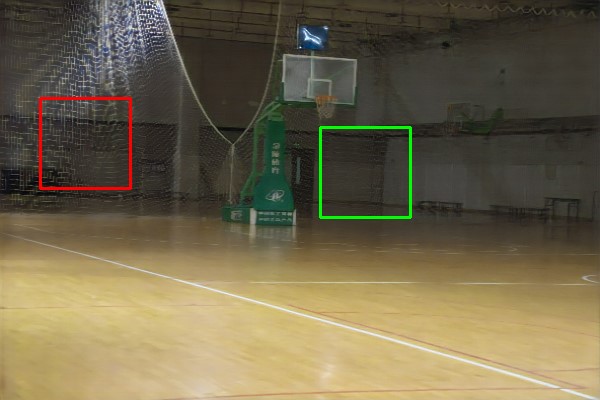} &
    \includegraphics[width=0.125\textwidth]{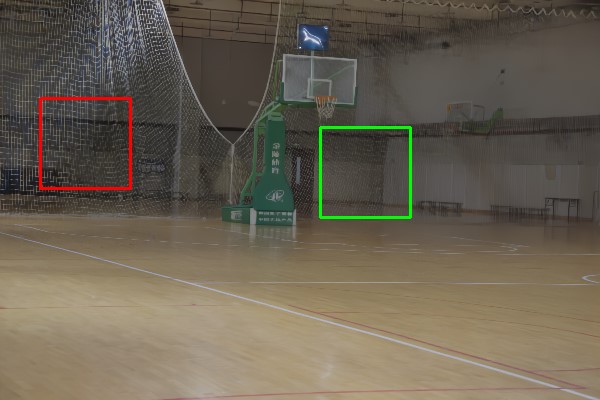} &
    \includegraphics[width=0.125\textwidth]{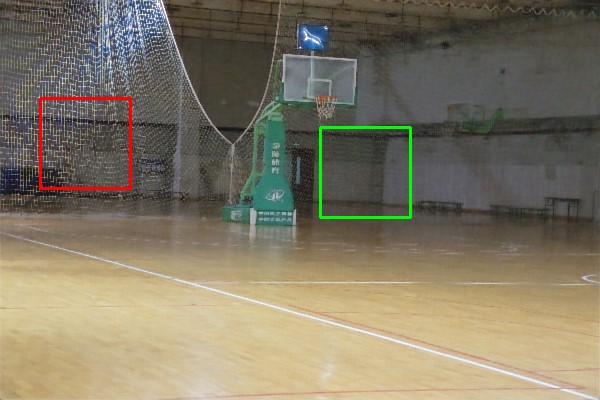} &
    \includegraphics[width=0.125\textwidth]{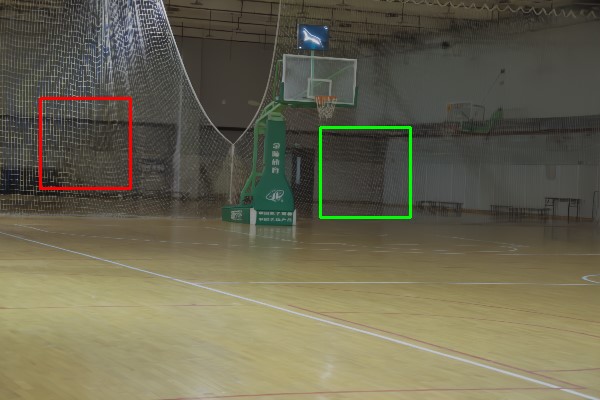} &
    \includegraphics[width=0.125\textwidth]{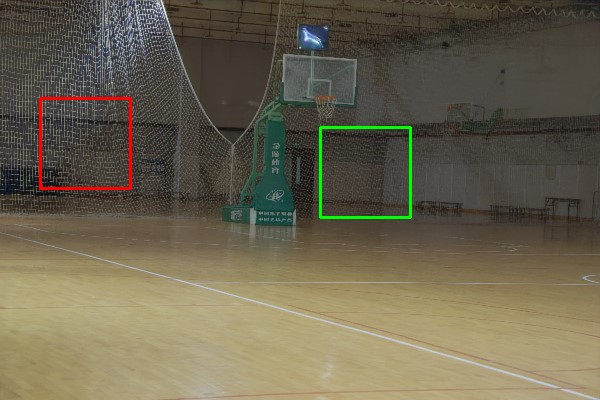} &
    \includegraphics[width=0.125\textwidth]{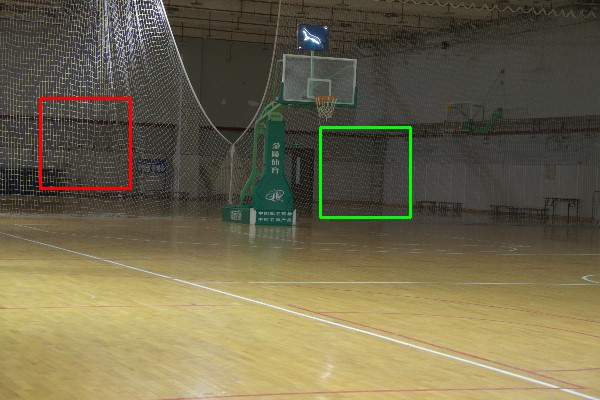} \\
    \includegraphics[width=0.06\textwidth]{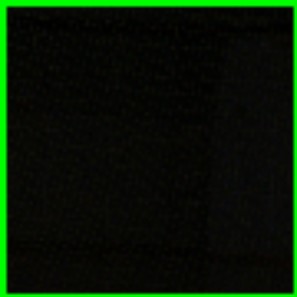} 
    \includegraphics[width=0.06\textwidth]{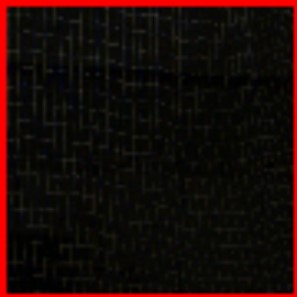} &
    \includegraphics[width=0.06\textwidth]{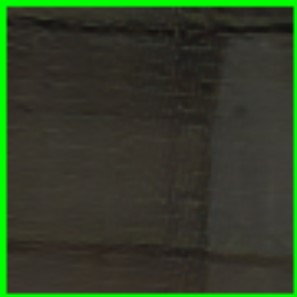} 
    \includegraphics[width=0.06\textwidth]{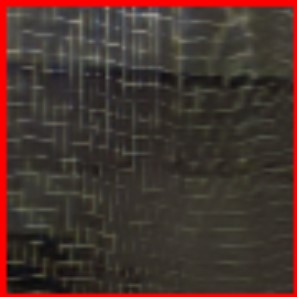} & 
    \includegraphics[width=0.06\textwidth]{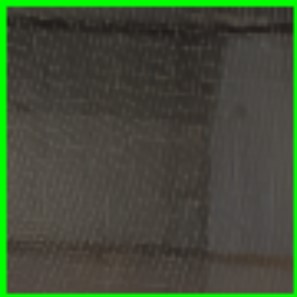} 
    \includegraphics[width=0.06\textwidth]{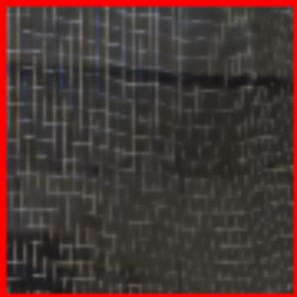} &
    \includegraphics[width=0.06\textwidth]{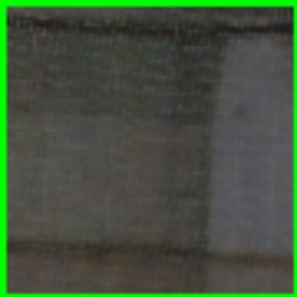} 
    \includegraphics[width=0.06\textwidth]{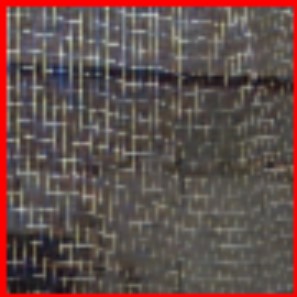} &
    \includegraphics[width=0.06\textwidth]{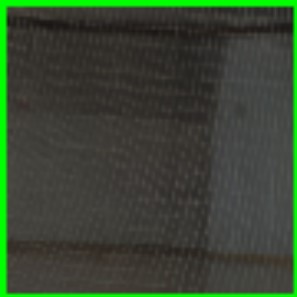}
    \includegraphics[width=0.06\textwidth]{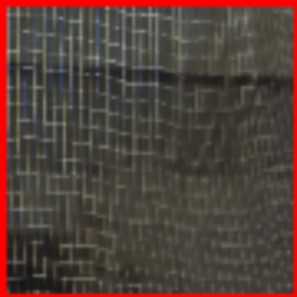}&
    \includegraphics[width=0.06\textwidth]{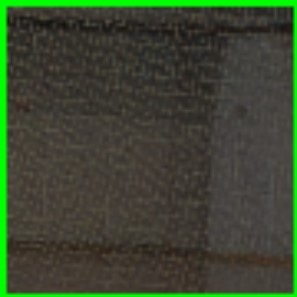}
    \includegraphics[width=0.06\textwidth]{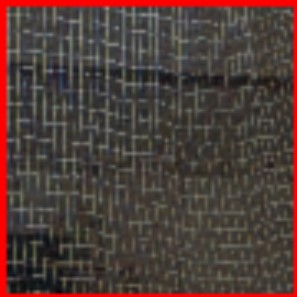}&
    \includegraphics[width=0.06\textwidth]{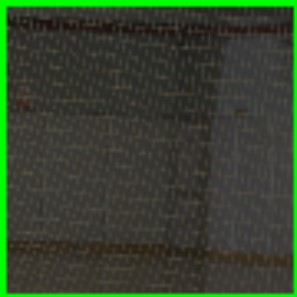}
    \includegraphics[width=0.06\textwidth]{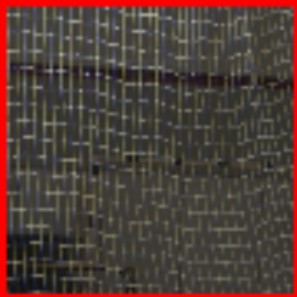}\\
    Input & KinD & LLFlow & Retformer & MambaIR & Ours & GT \\
\end{tabular}
\caption{Visual comparison of low-light image enhancement results across different methods. Red and green boxes highlight zoomed-in regions for qualitative evaluation.}
\label{fig:lowlight_comparison}
\end{figure*}


\section{Ablation Study}
In this section, we analyze the contributions of the core components in our VAMamba, including the QCLAM and GPS-SS2D, along with more detailed discussions and experiments. We demonstrate their effectiveness through quantitative results, runtime analysis, and visual comparisons. All experiments are conducted on the UHD-LL dataset, focusing on the low-light enhancement task.

\subsection{Effectiveness of QCLAM}
The QCLAM module is designed to reduce memory usage while maintaining model performance by reusing low-rank projections and intelligent feature caching during processing. To evaluate its impact, we compare the memory consumption and performance of the model with and without the QCLAM module.

\begin{figure*}[!htb]
    \centering
    \begin{adjustbox}{max width=\textwidth}
    \begin{tabular}{c c c c}
        \includegraphics[width=0.24\textwidth]{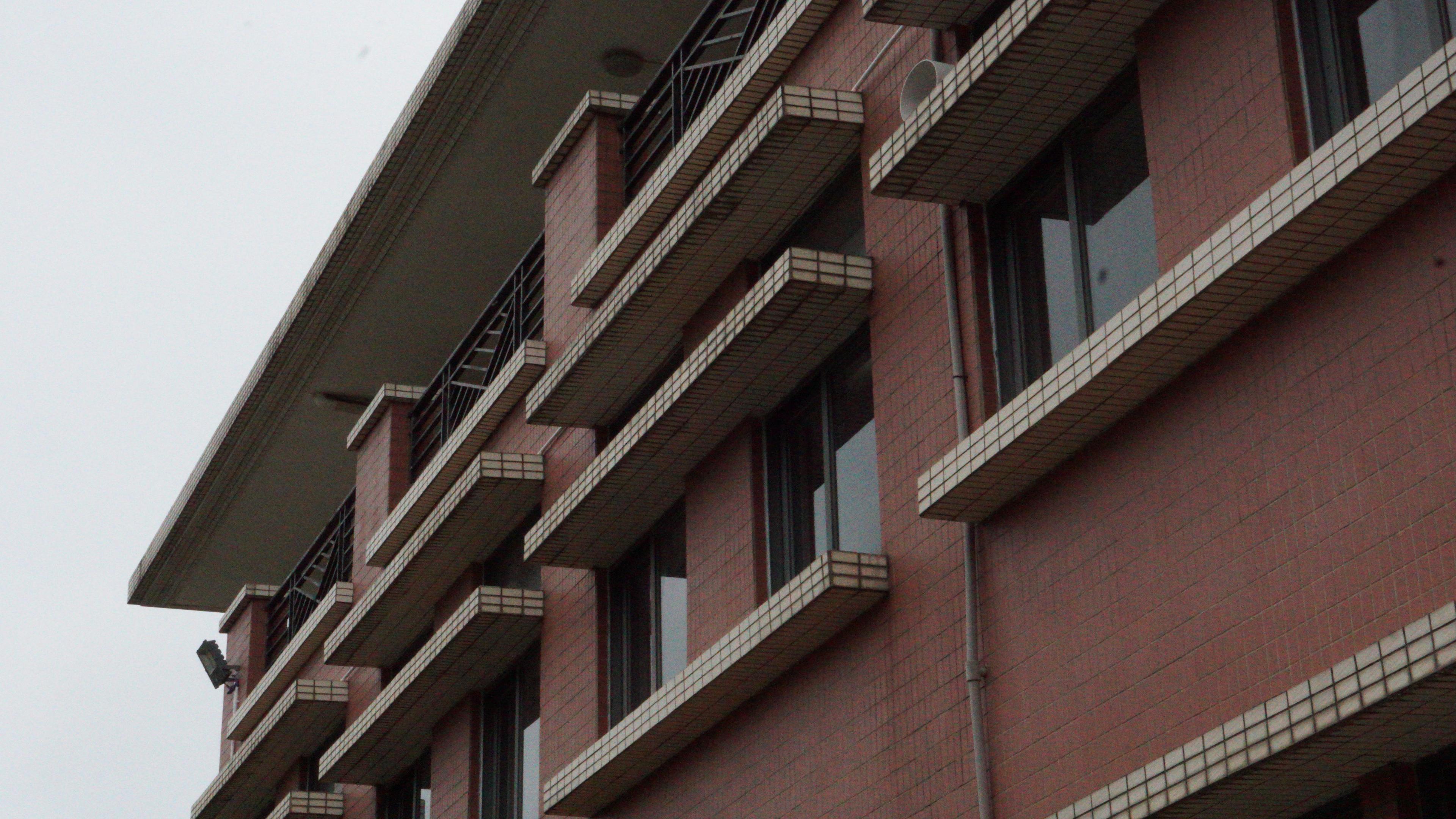} &
        \includegraphics[width=0.24\textwidth]{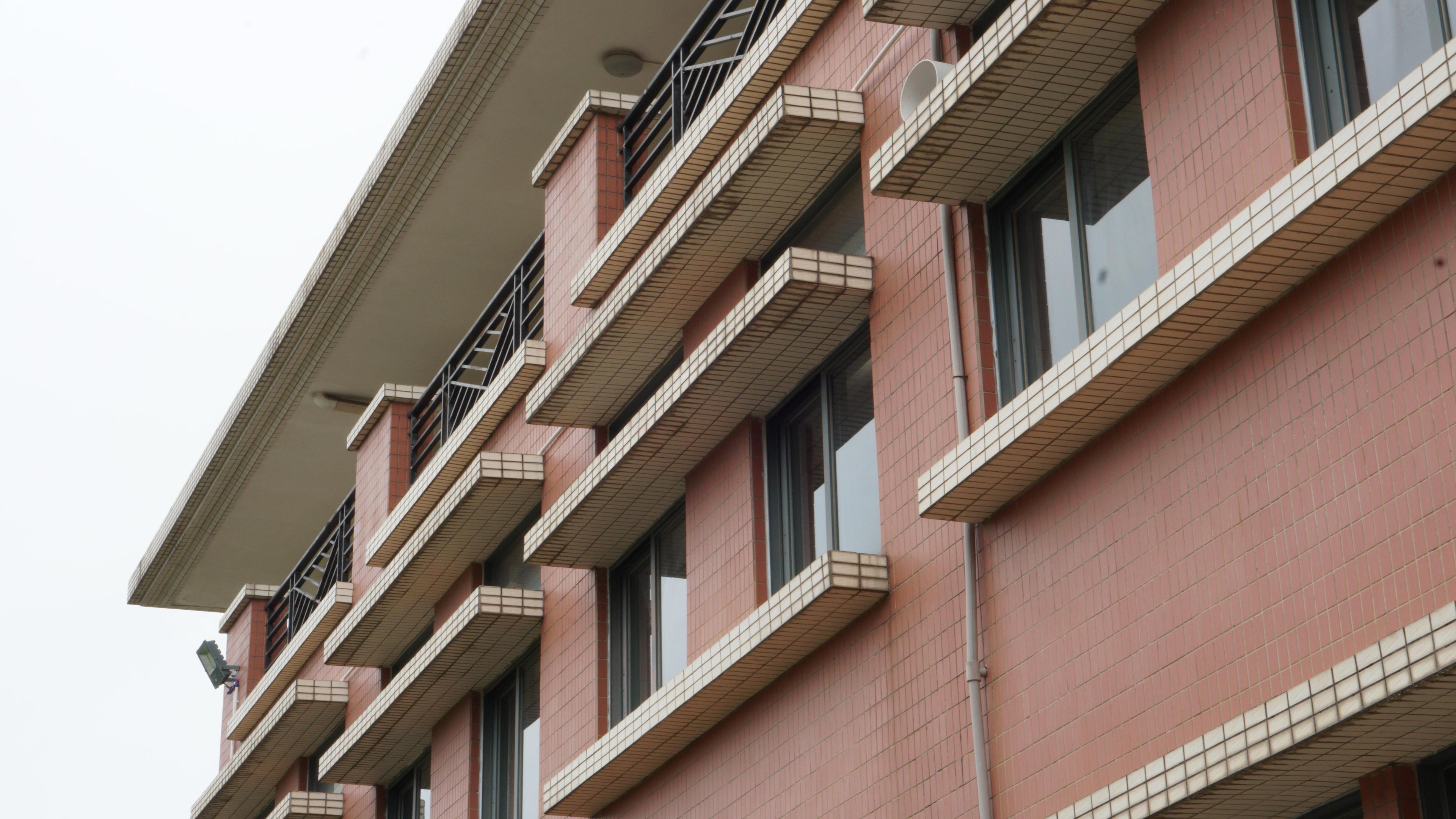} &
        \includegraphics[width=0.24\textwidth]{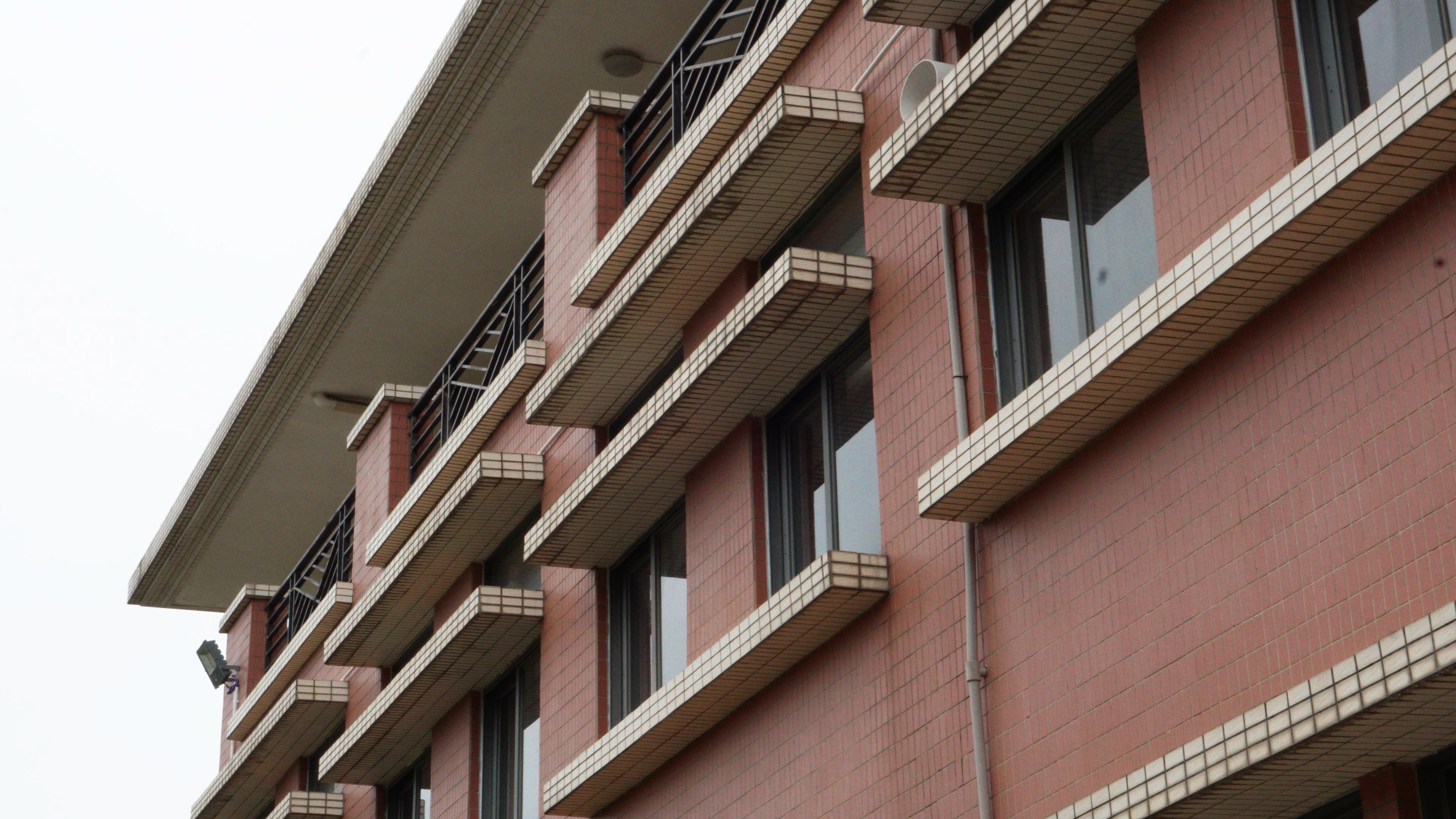} &
        \includegraphics[width=0.24\textwidth]{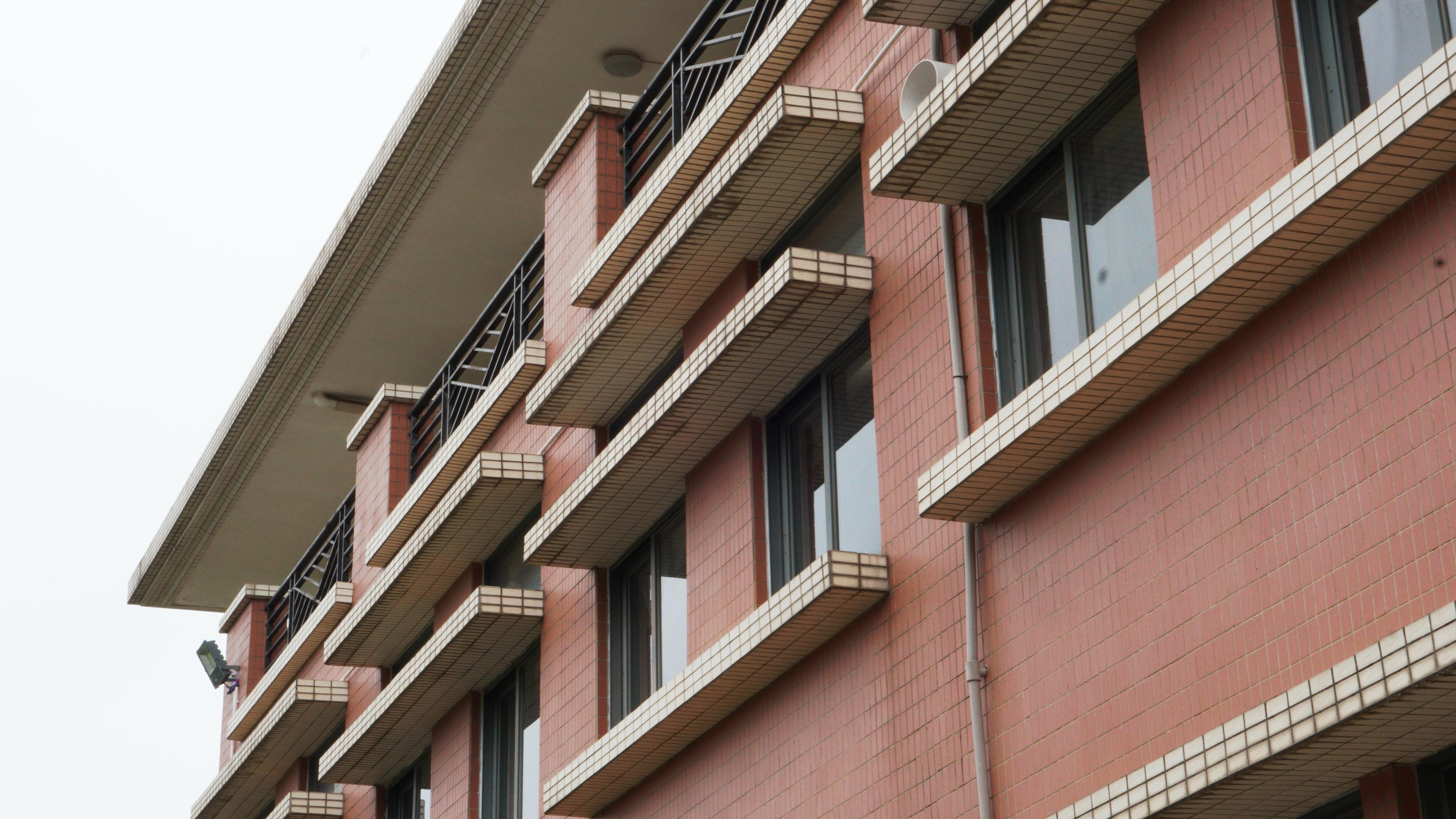} 
    \end{tabular}
    \end{adjustbox}
    \vspace{0.5em}
    \begin{adjustbox}{max width=\textwidth}
    \begin{tabular}{c c c c}
        \includegraphics[width=0.24\textwidth]{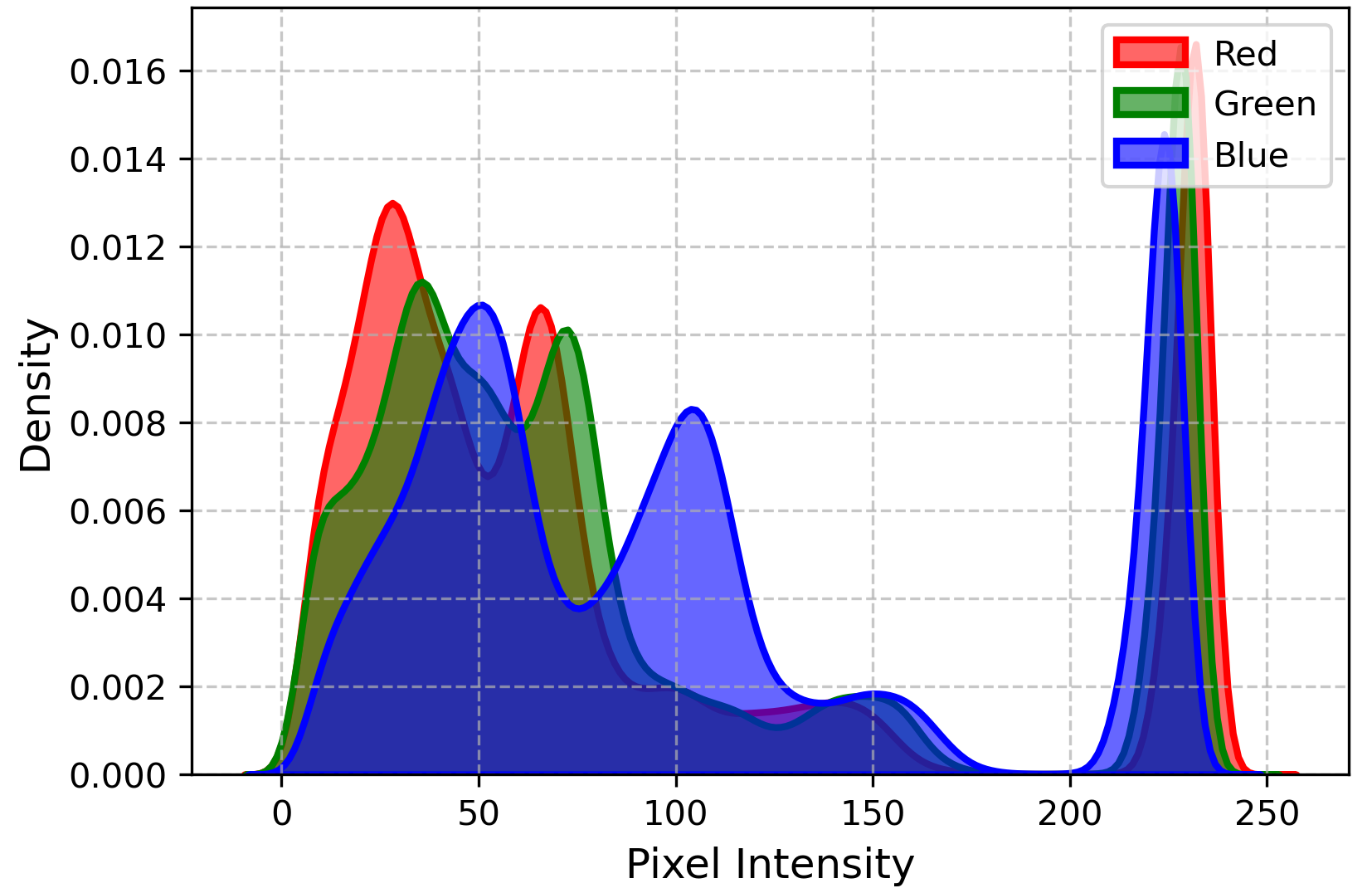} &
        \includegraphics[width=0.24\textwidth]{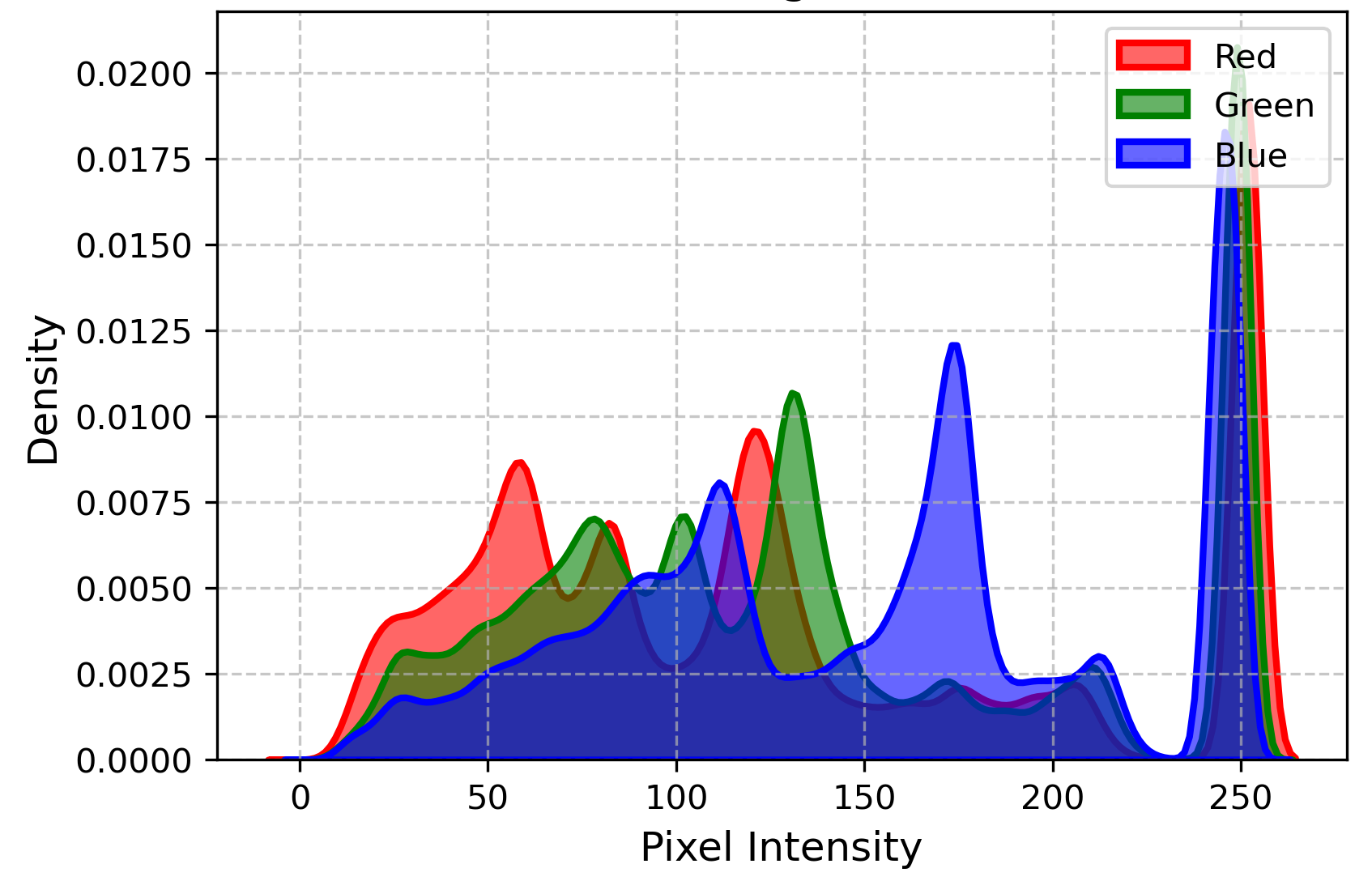} &
        \includegraphics[width=0.24\textwidth]{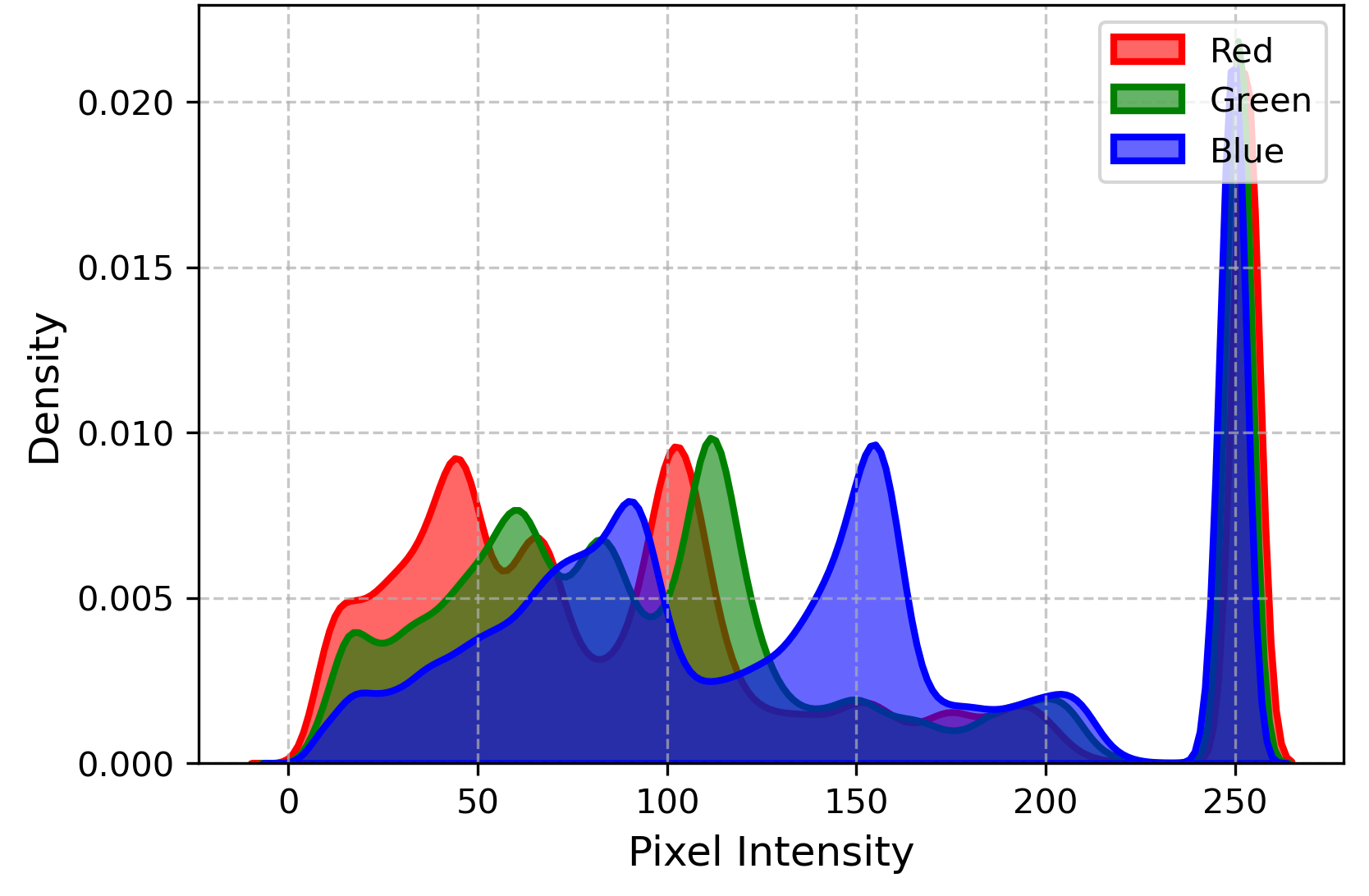} &
        \includegraphics[width=0.24\textwidth]{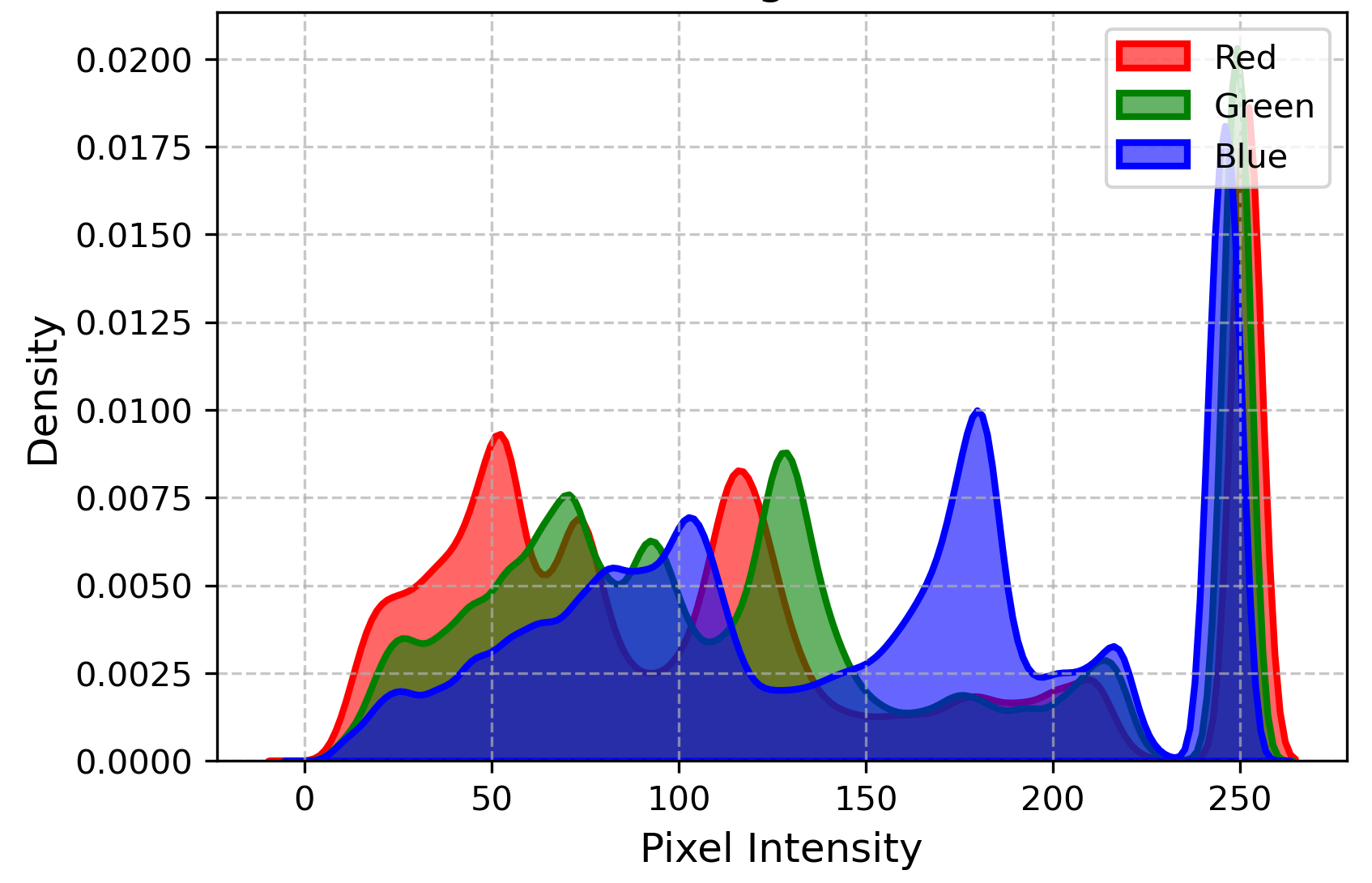} \\
        \small (a) Input & \small (b) Ground Truth & \small (c) W/ LoRA Cache& \small (d) W/O LoRA Cache \\
    \end{tabular}
    \end{adjustbox}
    \caption{Comparison of restoration quality and RGB distribution. Our method with LoRA Cache produces results that are closer to the ground truth in both visual quality and RGB balance.}
    \label{fig:rgb_distribution}
\end{figure*}

\begin{table}[!htb]
    \centering
    \caption{Impact of LoRA Cache on memory consumption and performance, evaluated on the UHD-LL dataset with full-resolution inputs of $3840 \times 2160$. Memory consumption (in GB) was measured using PyTorch's \texttt{torch.cuda.max\_memory\_allocated} utility during the inference stage with full-resolution images.}
    \label{tab:lora_cache}
    \scalebox{1.2}{
    \begin{tabular}{l|c|c|c}
        \toprule[0.15em]
        \textbf{Model Variant} & \textbf{PSNR} & \textbf{SSIM} & \textbf{Memory (GB)}  \\
        \midrule
        w/o QCLAM & 27.07 & 0.926 & 12.4 \\
        w/ QCLAM & \textbf{27.13} & \textbf{0.928} & \textbf{9.6} \\
        \bottomrule[0.15em]
    \end{tabular}}\vspace{-2mm}
\end{table}
As shown in Table.~\ref{tab:lora_cache}, the QCLAM module reduces memory usage by 23.5\% while achieving comparable performance in terms of PSNR and SSIM. Notably, the memory savings enable processing larger batch sizes, which is crucial for high-resolution images. To quantitatively measure the restoration quality beyond traditional metrics, we analyze the RGB distribution of the restored images. Fig.~\ref{fig:rgb_distribution} shows the RGB histograms of images restored with and without QCLAM. As seen in the figure, QCLAM helps to preserve a more balanced RGB distribution, reducing color distortion and improving the overall color fidelity in the restored images. The histograms reveal that the model with QCLAM produces a more natural distribution of pixel intensities across the red, green, and blue channels, which translates to visually more pleasing results.

\begin{figure*}[!htb]
    \centering
    \begin{adjustbox}{max width=\textwidth}
    \begin{tabular}{c c c c}
        \includegraphics[width=0.24\textwidth]{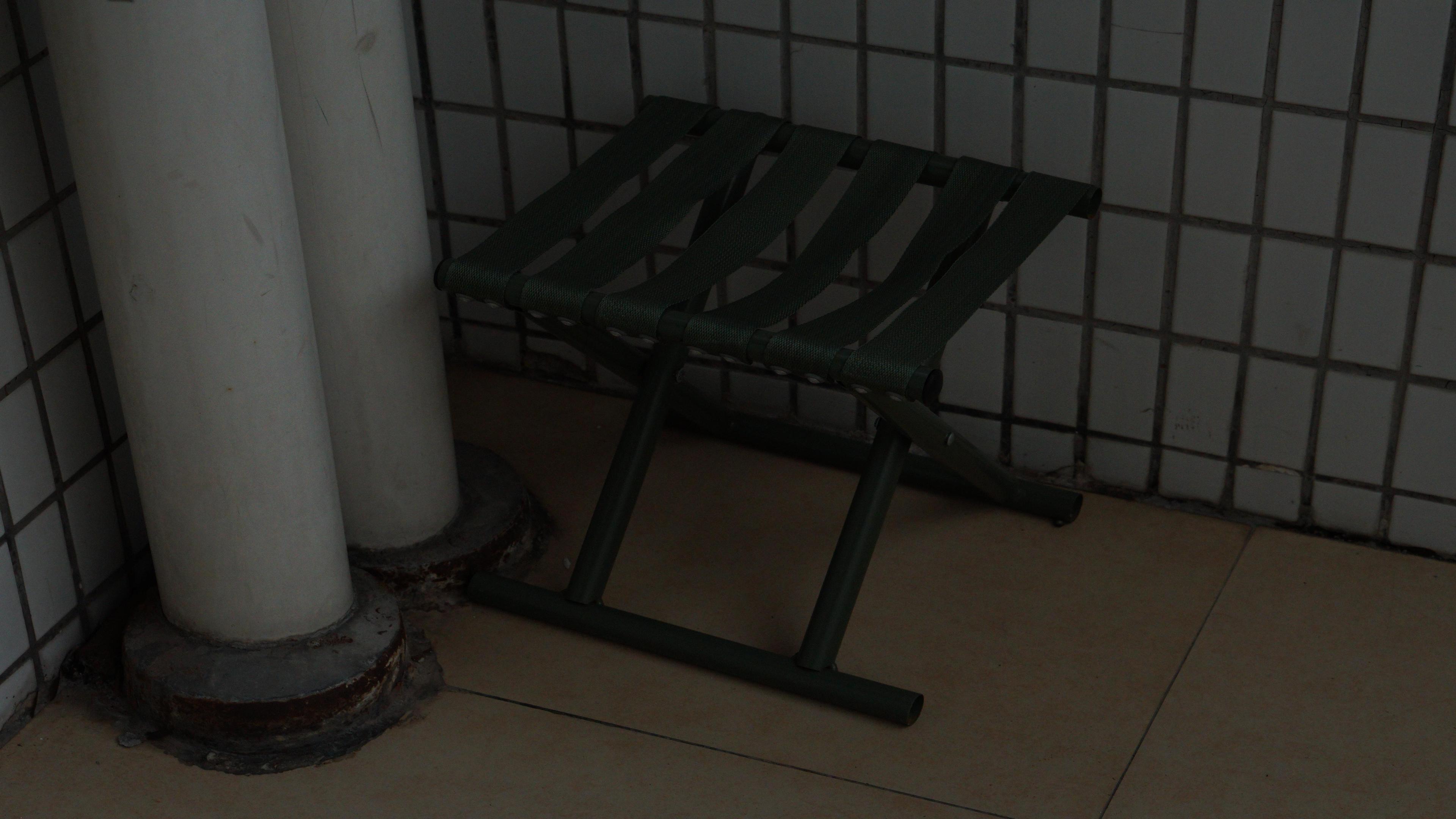} &
        \includegraphics[width=0.24\textwidth]{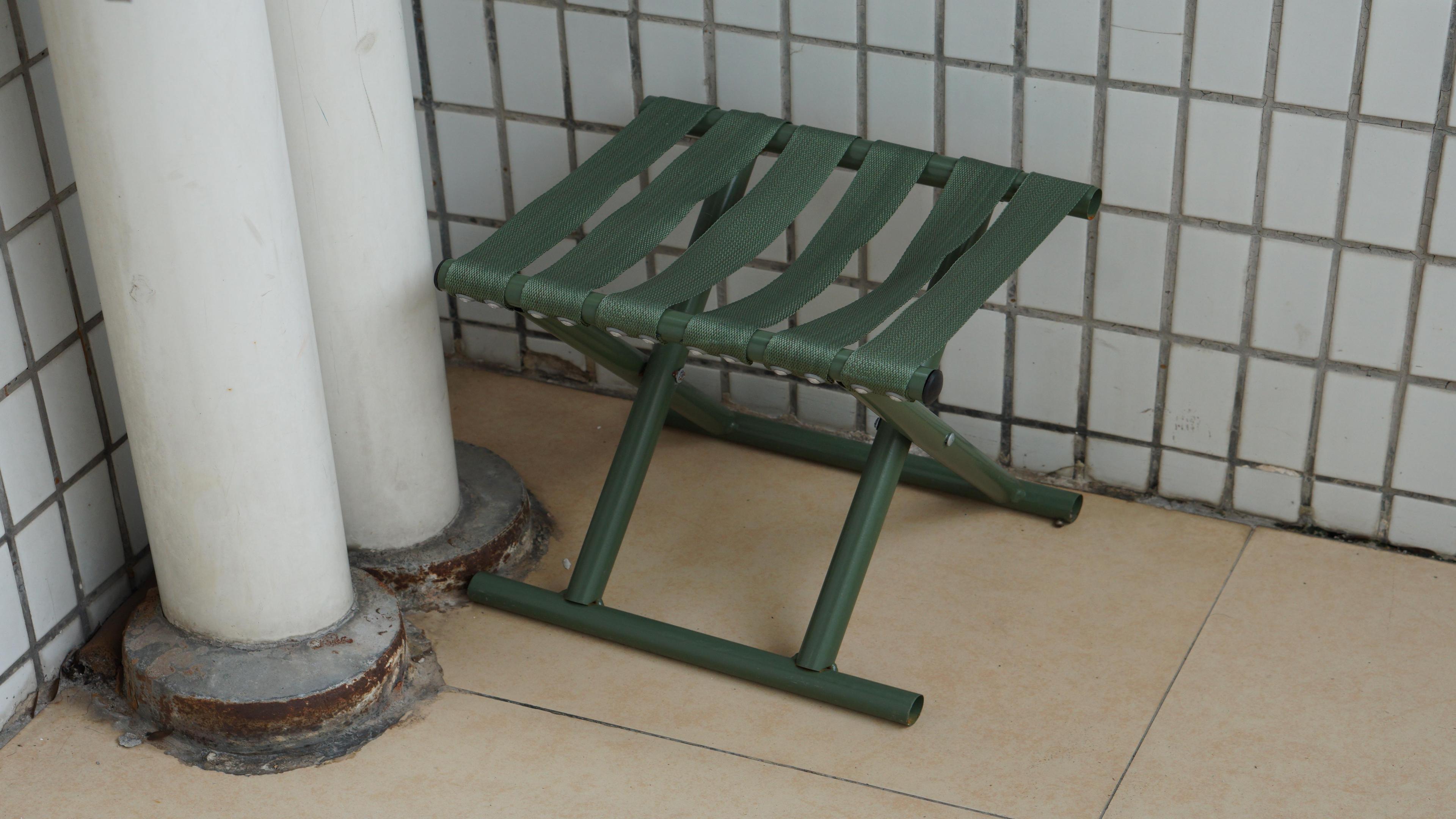} &
        \includegraphics[width=0.24\textwidth]{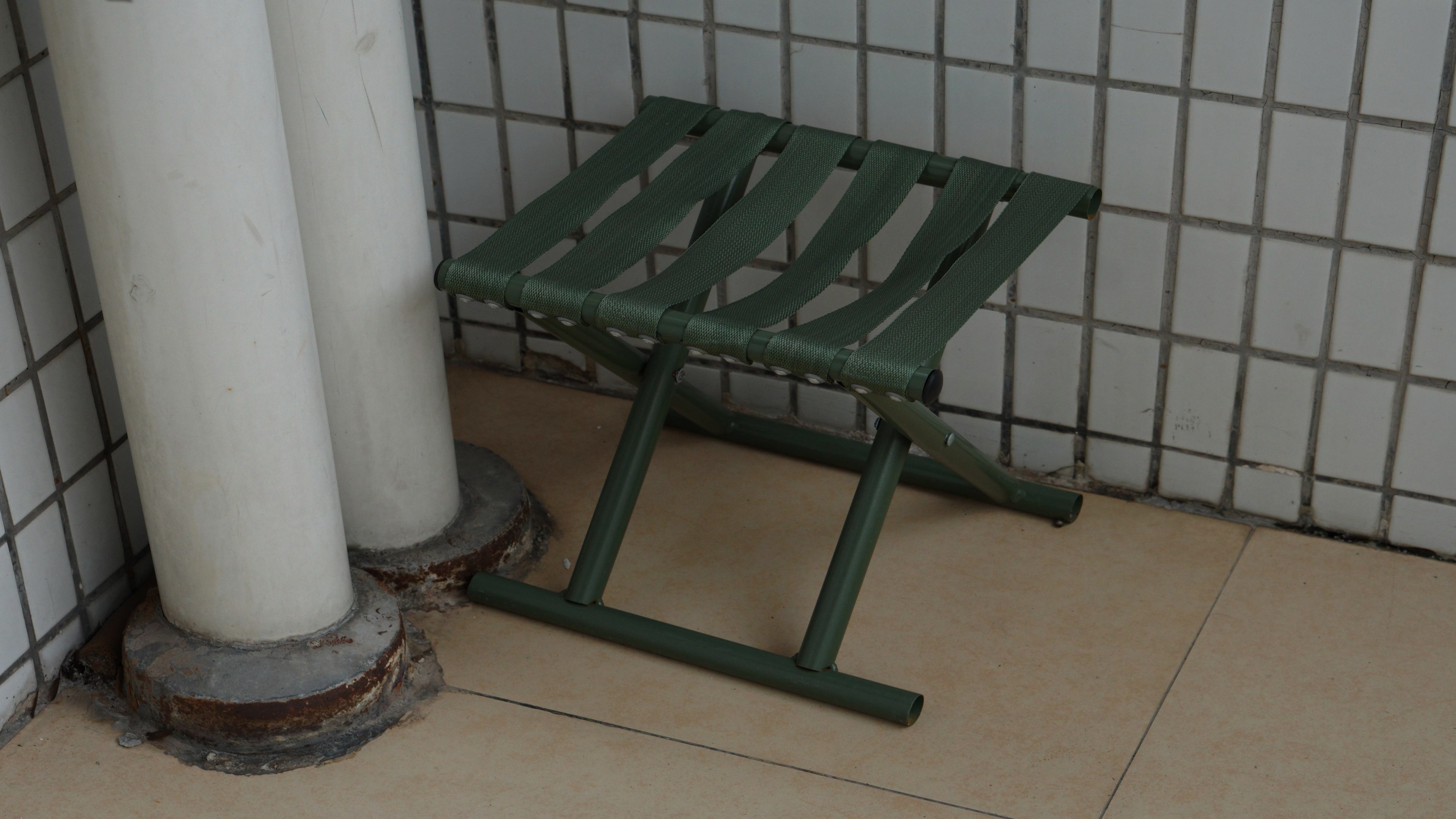} &
        \includegraphics[width=0.24\textwidth]{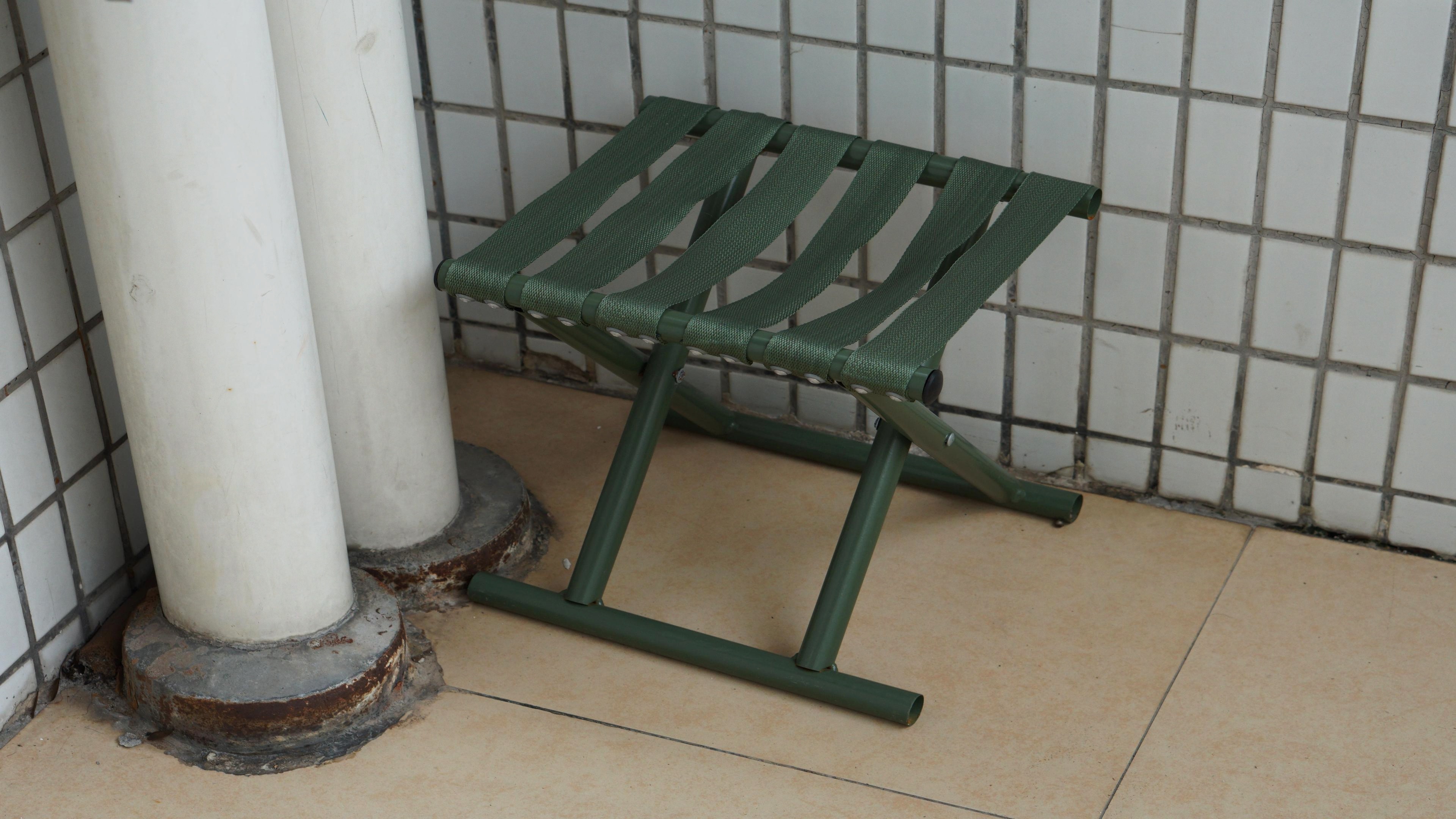} 
    \end{tabular}
    \end{adjustbox}
    \vspace{0.5em}
    \begin{adjustbox}{max width=\textwidth}
    \begin{tabular}{c c c c}
        \includegraphics[width=0.24\textwidth]{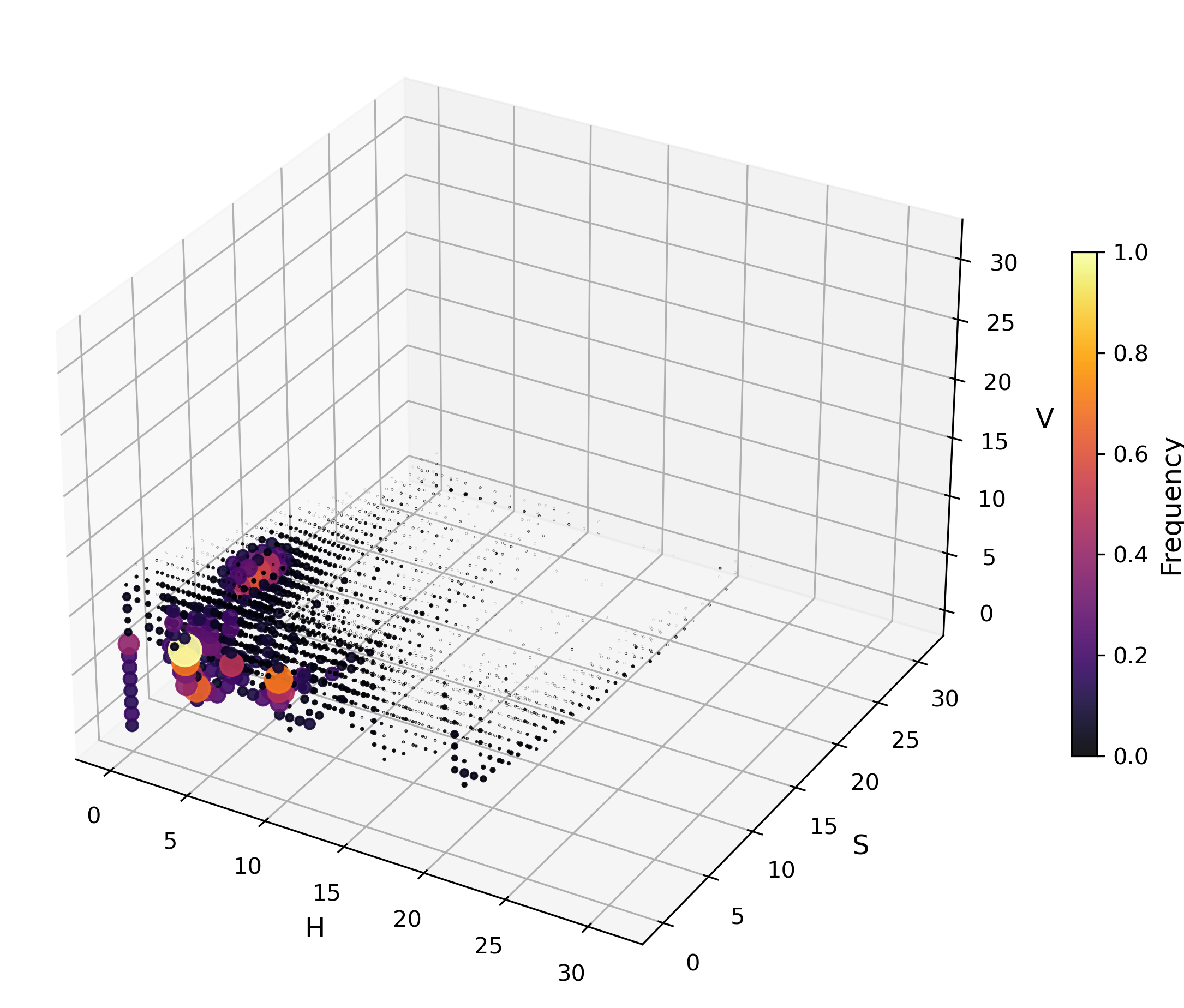} &
        \includegraphics[width=0.24\textwidth]{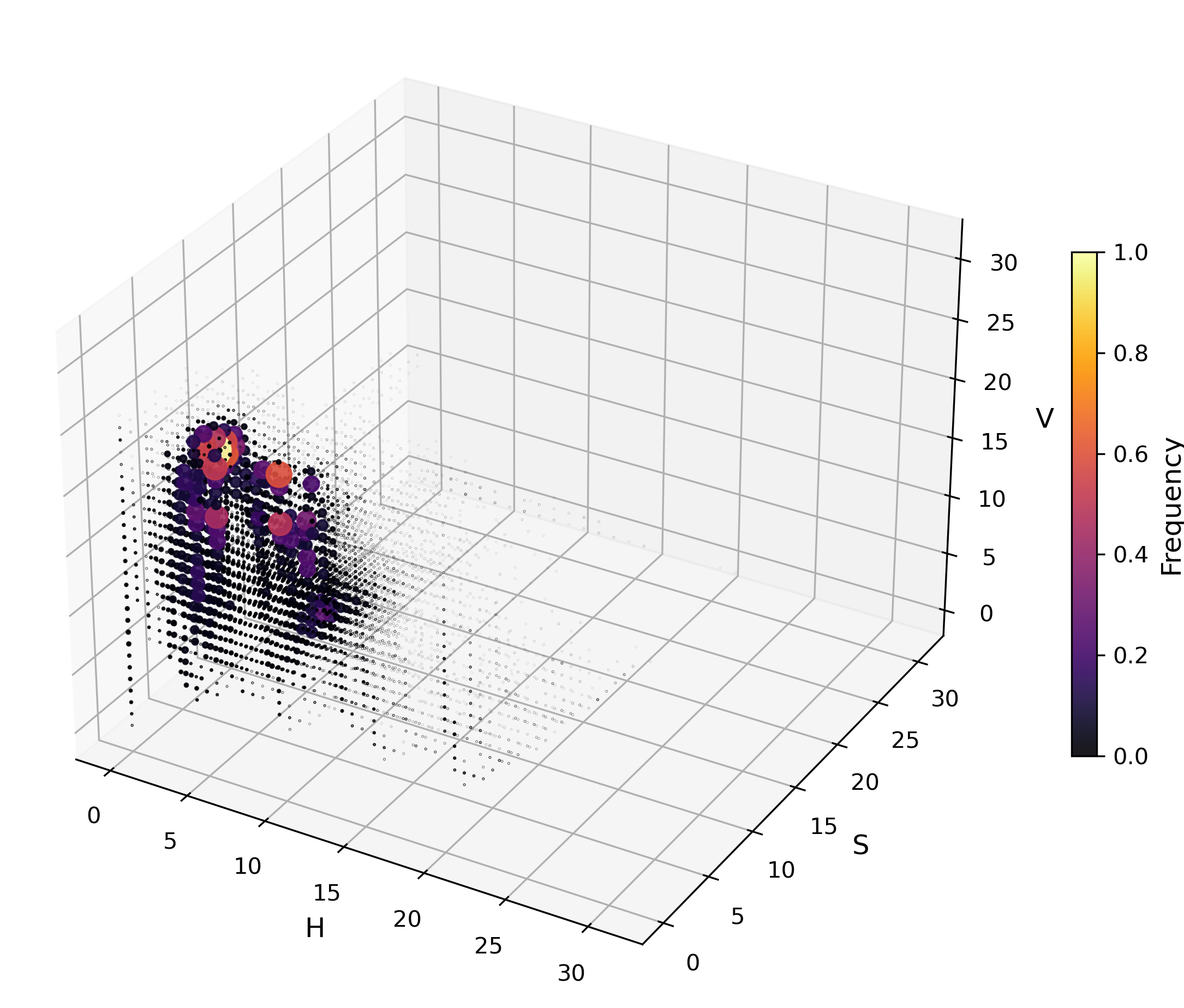} &
        \includegraphics[width=0.24\textwidth]{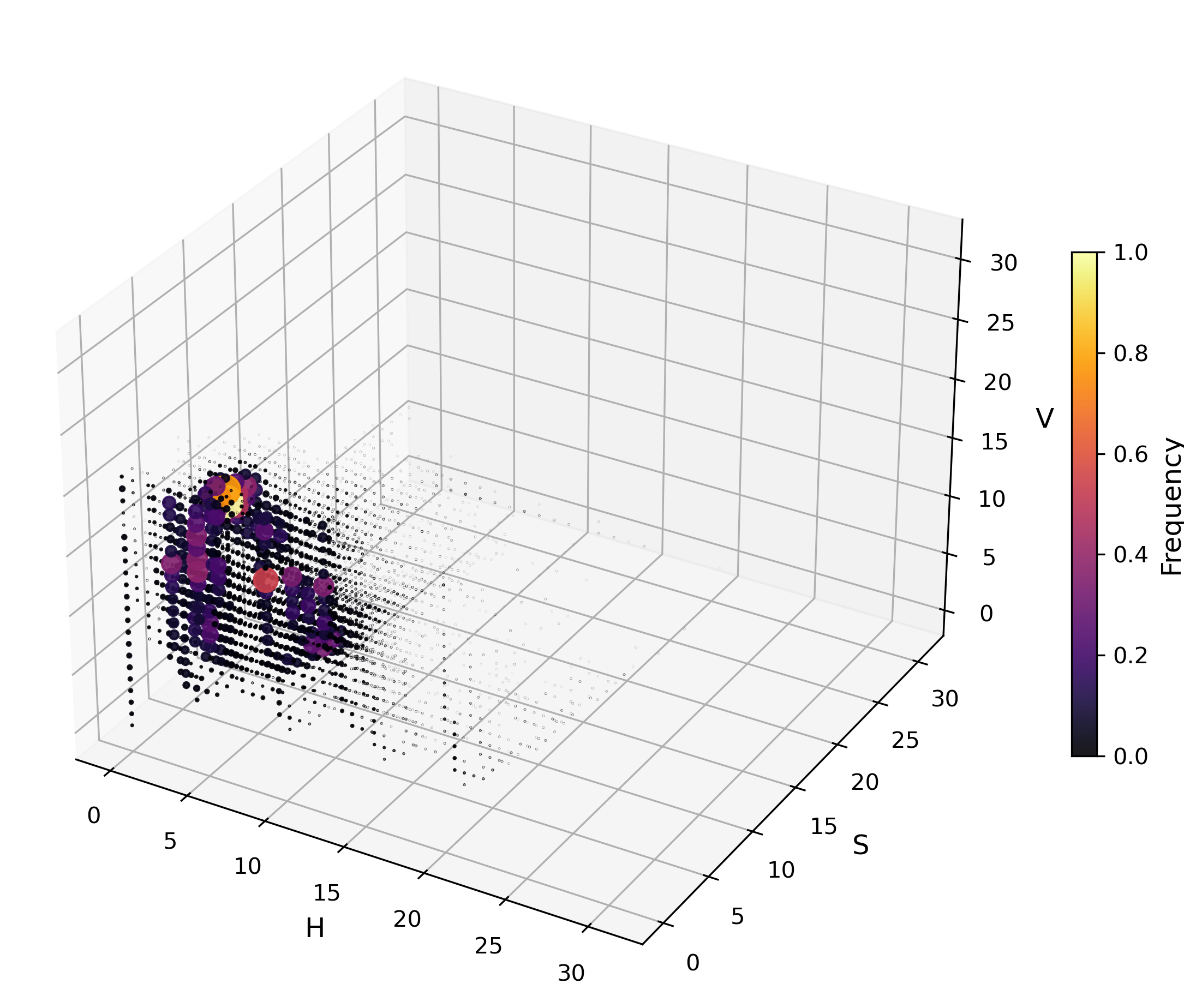} &
        \includegraphics[width=0.24\textwidth]{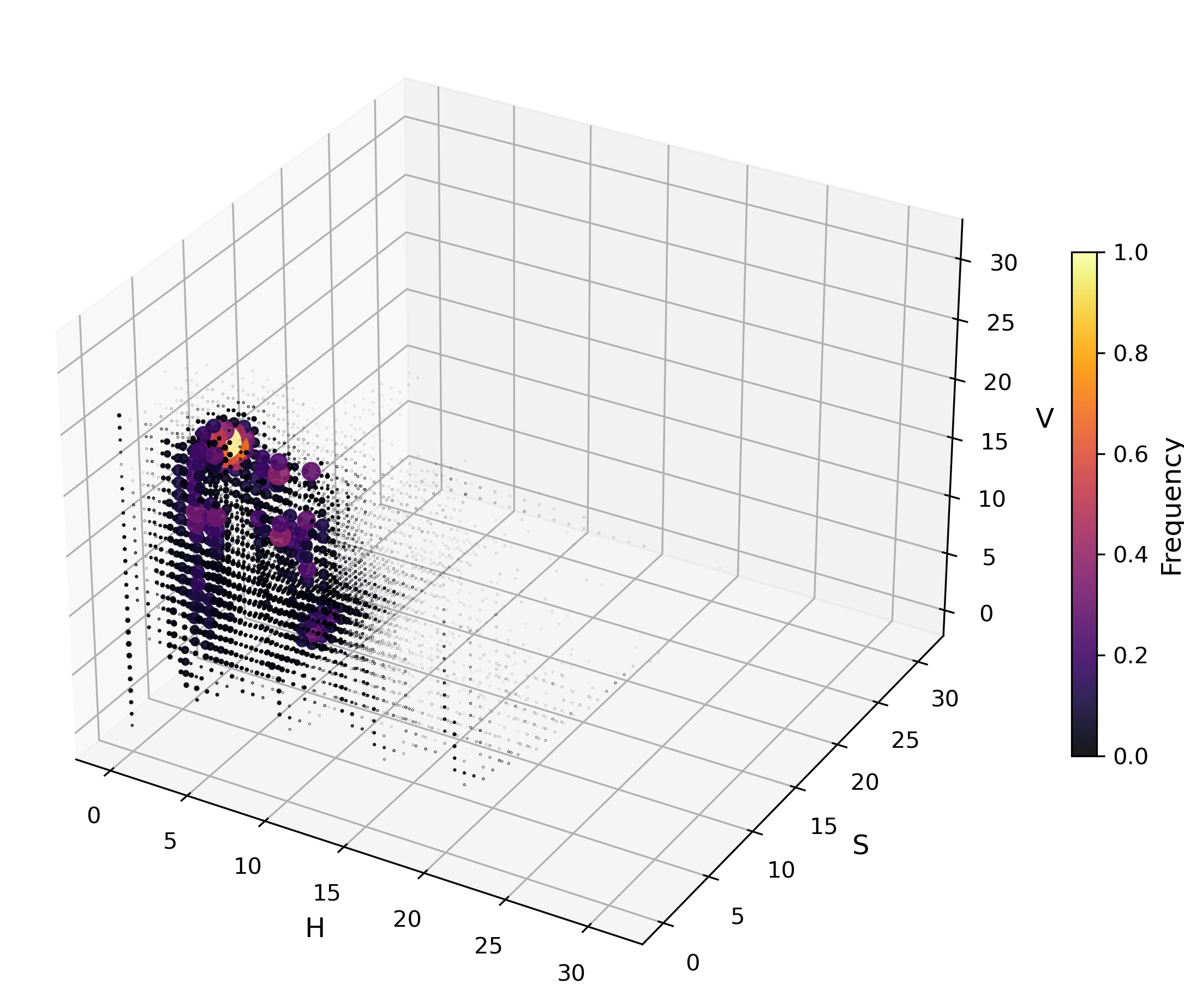} \\
        \small (a) Input & \small (b) Ground Truth & \small (c) W/ Adaptive Scanning& \small (d) W/O Adaptive Scanning \\
    \end{tabular}
    \end{adjustbox}
    \caption{HSV color space comparison of image restoration results with and without dynamic adaptive scanning. The model with adaptive scanning produces a restoration that is closer to the ground truth in terms of hue and saturation, preserving color consistency and enhancing fine details.}
    \label{fig:HSV}
\end{figure*}

\begin{figure*}[!htp]
    \centering
    \begin{adjustbox}{max width=\textwidth}
    \begin{tabular}{c c c c}
        \includegraphics[width=0.24\textwidth]{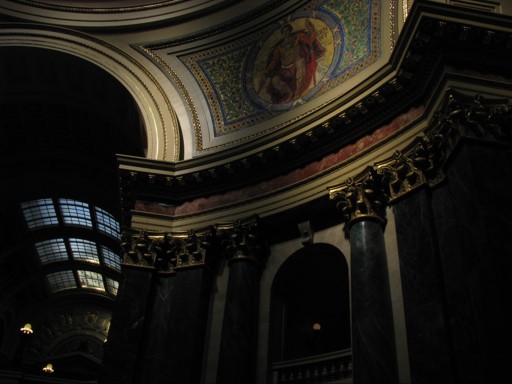} &
        \includegraphics[width=0.24\textwidth]{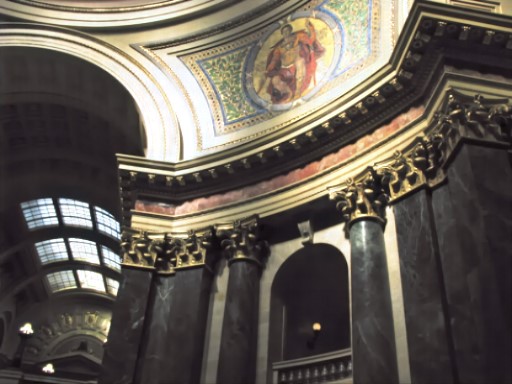} &
        \includegraphics[width=0.24\textwidth]{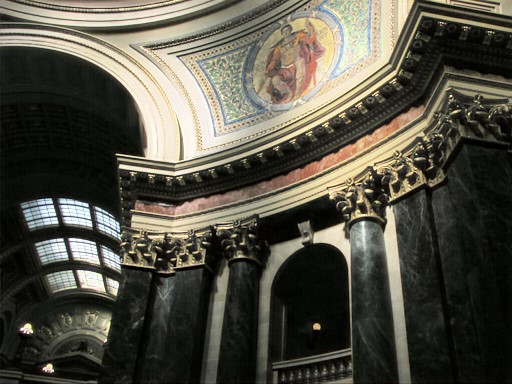} &
        \includegraphics[width=0.24\textwidth]{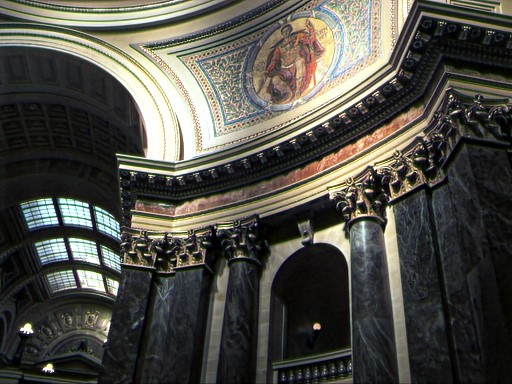} 
    \end{tabular}
    \end{adjustbox}
    \vspace{0.5em}
    \begin{adjustbox}{max width=\textwidth}
    \begin{tabular}{c c c c}
        \includegraphics[width=0.24\textwidth]{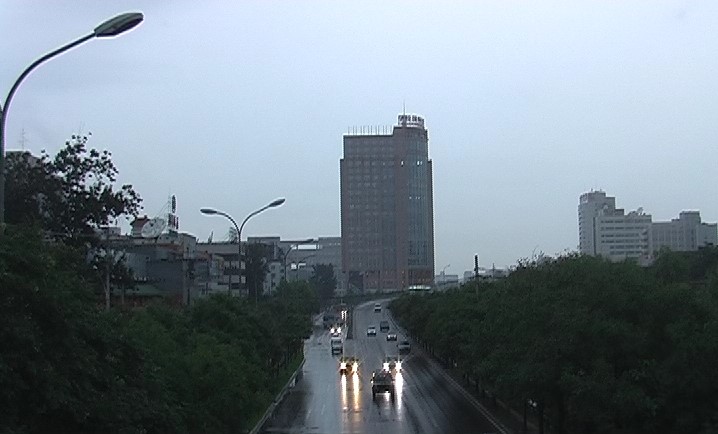} &
        \includegraphics[width=0.24\textwidth]{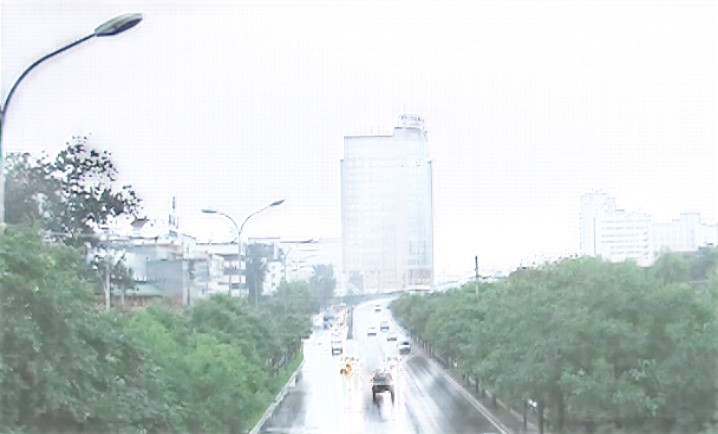} &
        \includegraphics[width=0.24\textwidth]{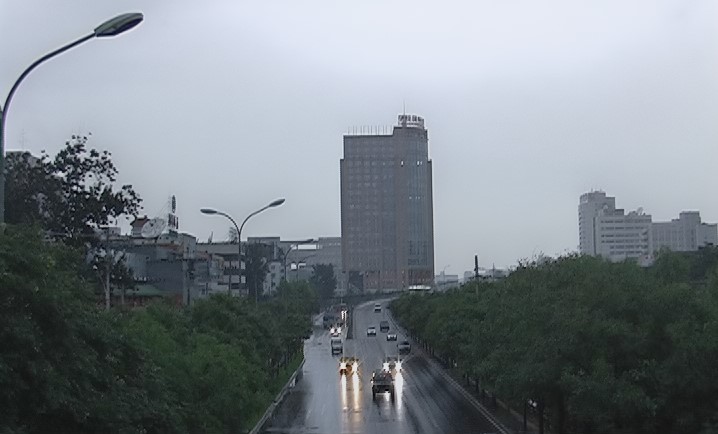} &
        \includegraphics[width=0.24\textwidth]{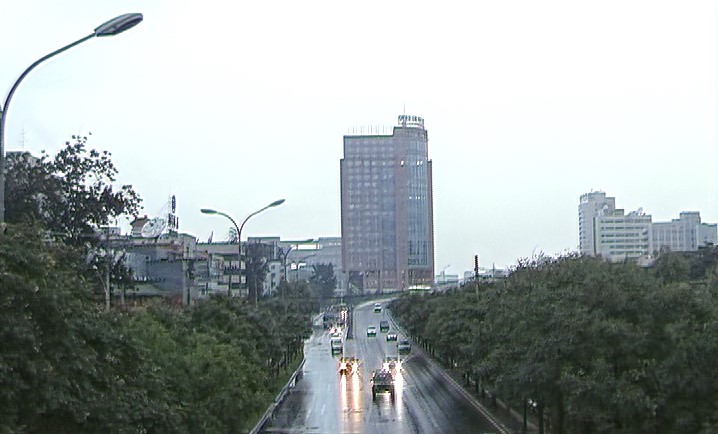}  \\
        \small (a) Input & \small (b) Retformer & \small (c) MambaIR & \small (d) Ours  \\
    \end{tabular}
    \end{adjustbox}
    \caption{Visualization results of cross-dataset evaluation on unpaired real-world low-light datasets.}
    \label{fig:realworld}\vspace{-5mm}
\end{figure*}


\begin{figure*}[!htp]
    \centering
    \begin{adjustbox}{max width=\textwidth}
    \begin{tabular}{c c c c c c}
        \includegraphics[width=0.16\textwidth]{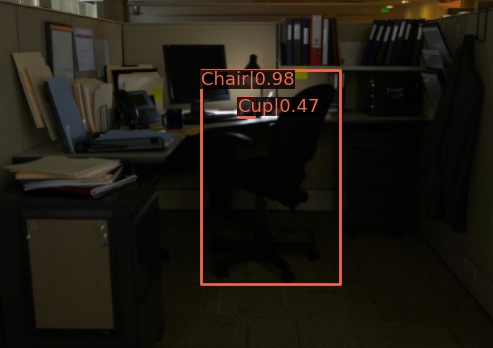} &
        \includegraphics[width=0.16\textwidth]{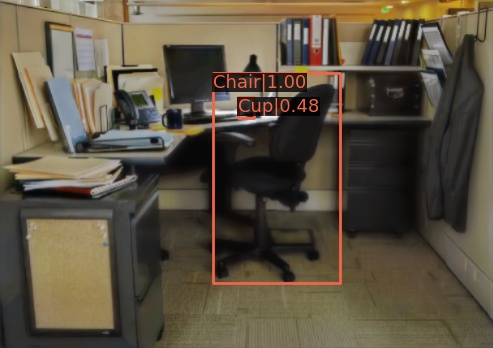} &
        \includegraphics[width=0.16\textwidth]{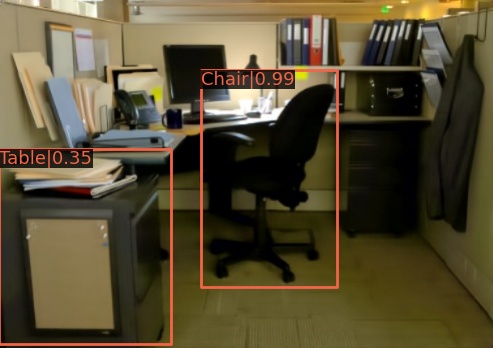} &
        \includegraphics[width=0.16\textwidth]{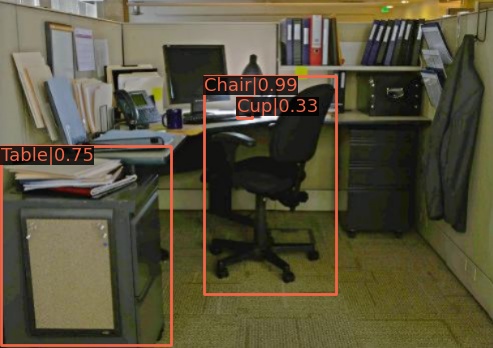}&
        \includegraphics[width=0.16\textwidth]{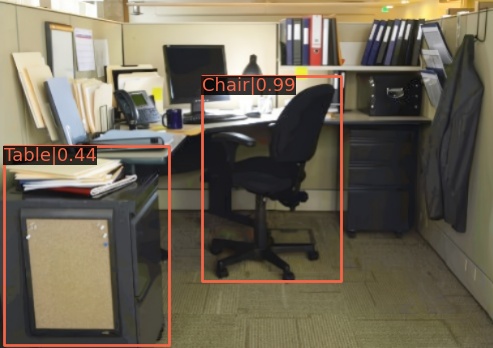} &
        \includegraphics[width=0.16\textwidth]{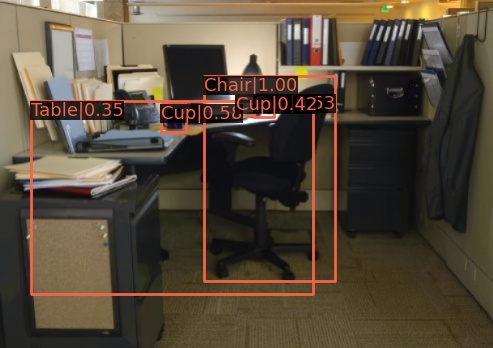} \\
        \small (a) Input & \small (b) KinD& \small (c) DRBN & \small (d) Retformer & \small (e) MambaIR & \small (f) Ours  \\
    \end{tabular}
    \end{adjustbox}
    \vspace{0.5em}
    \begin{adjustbox}{max width=\textwidth}
    \begin{tabular}{c c c c c c}
        \includegraphics[width=0.16\textwidth]{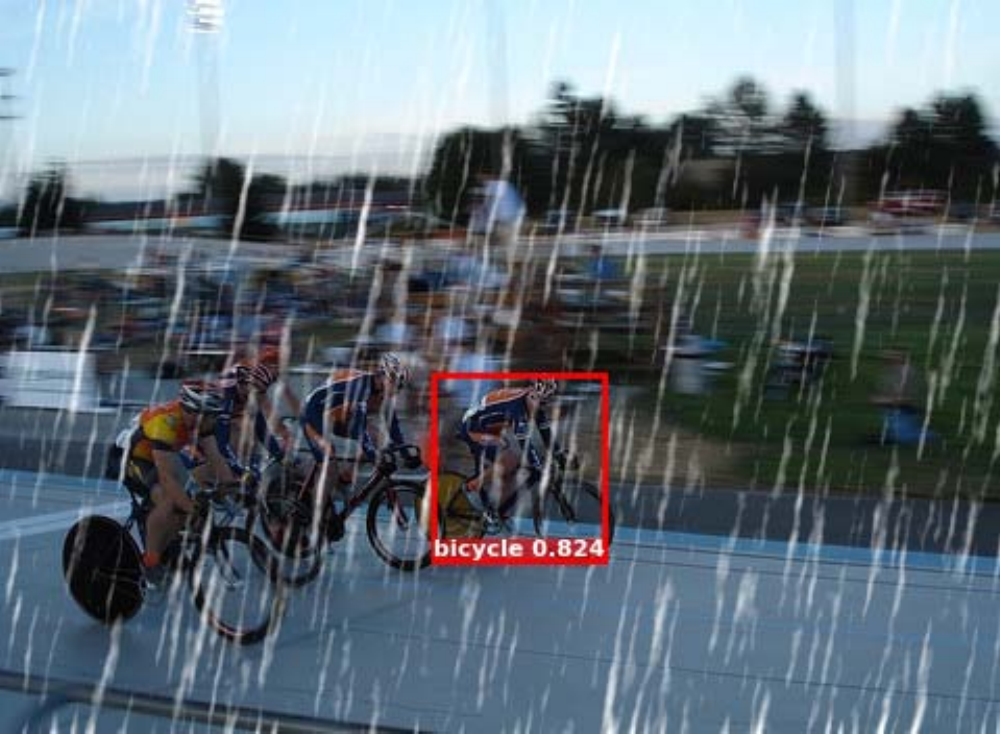} &
        \includegraphics[width=0.16\textwidth]{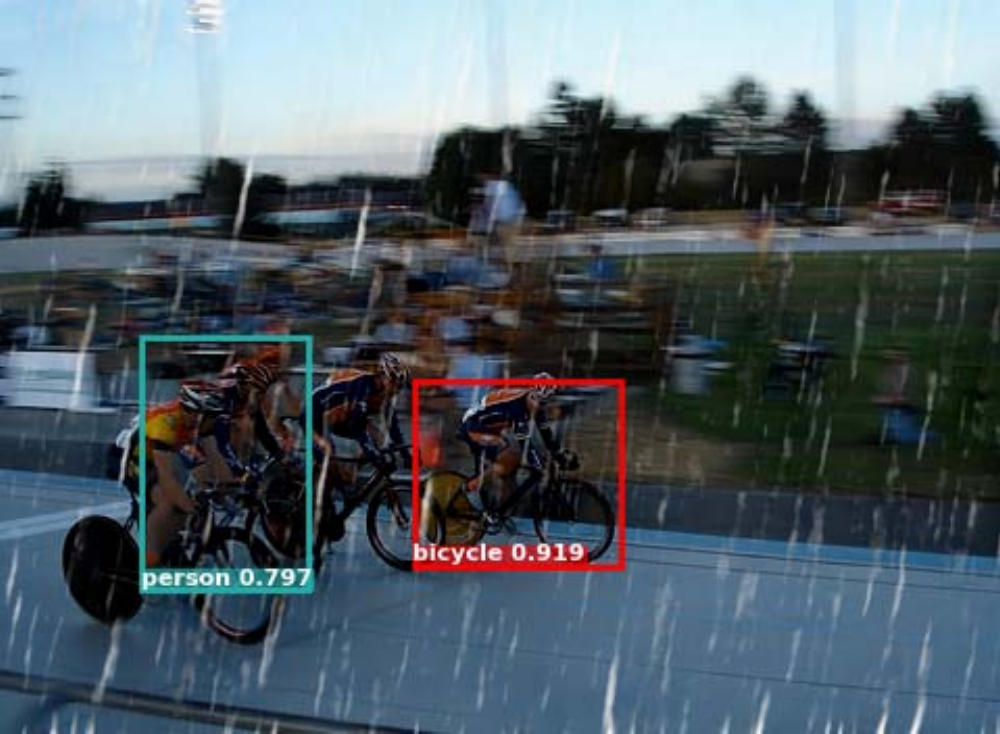} &
        \includegraphics[width=0.16\textwidth]{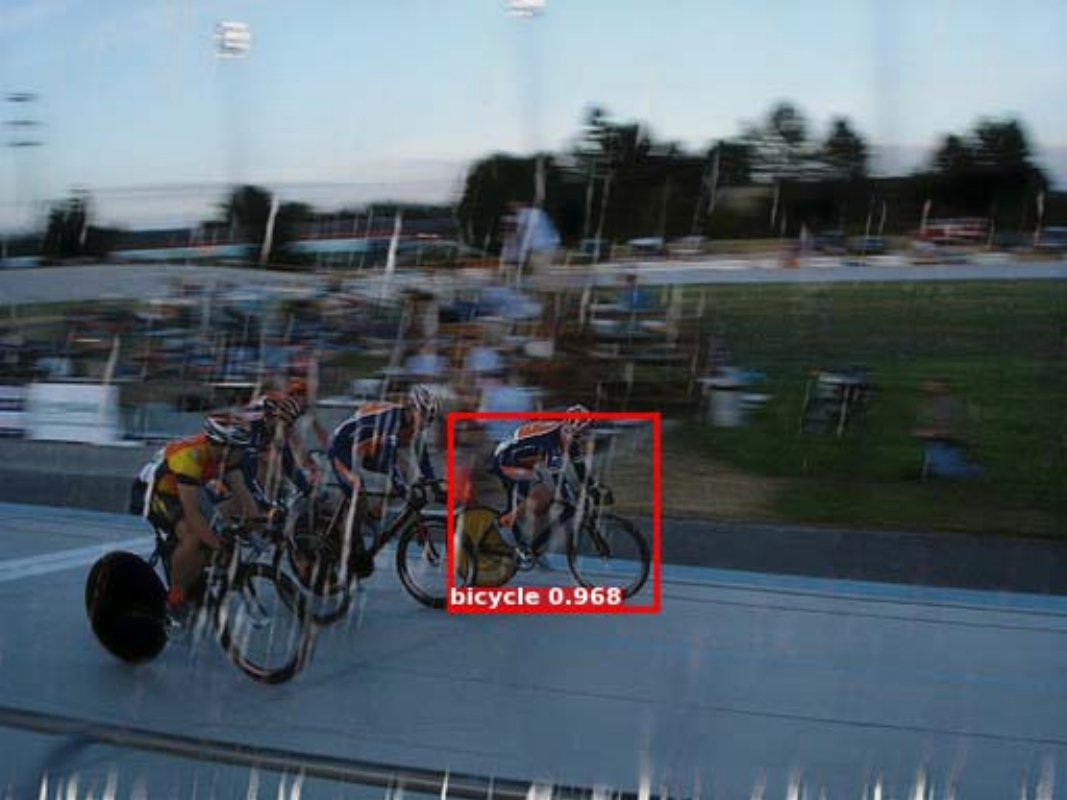} &
        \includegraphics[width=0.16\textwidth]{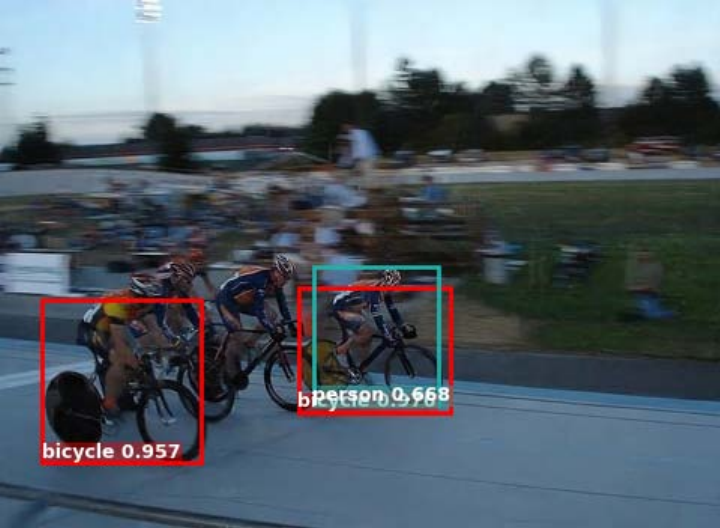}&
        \includegraphics[width=0.16\textwidth]{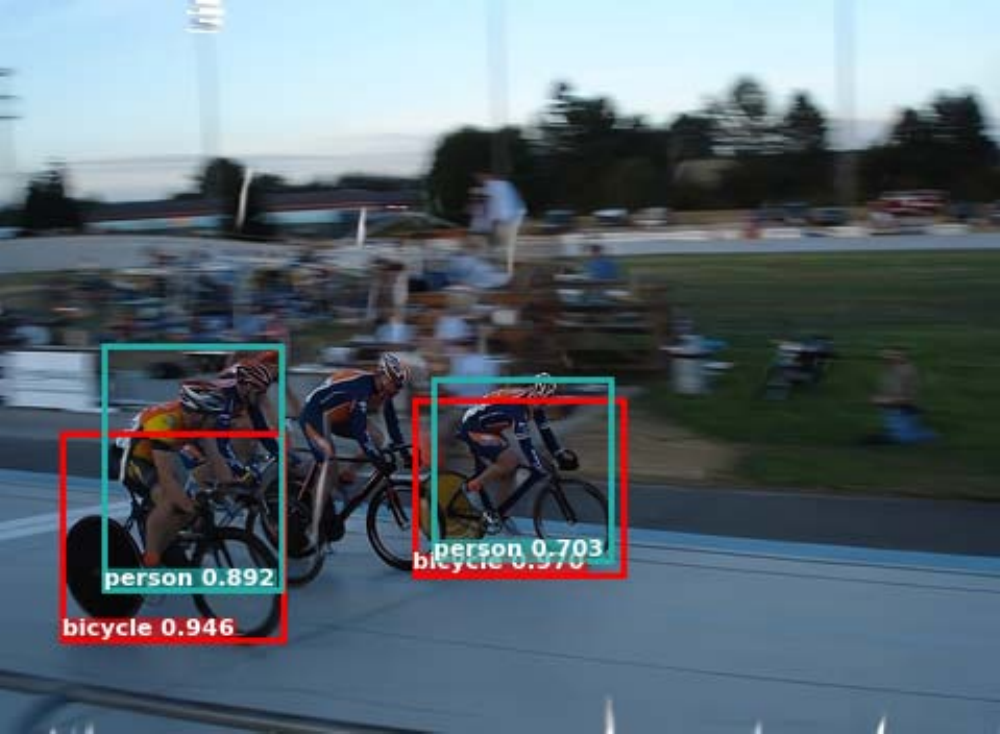} &
        \includegraphics[width=0.16\textwidth]{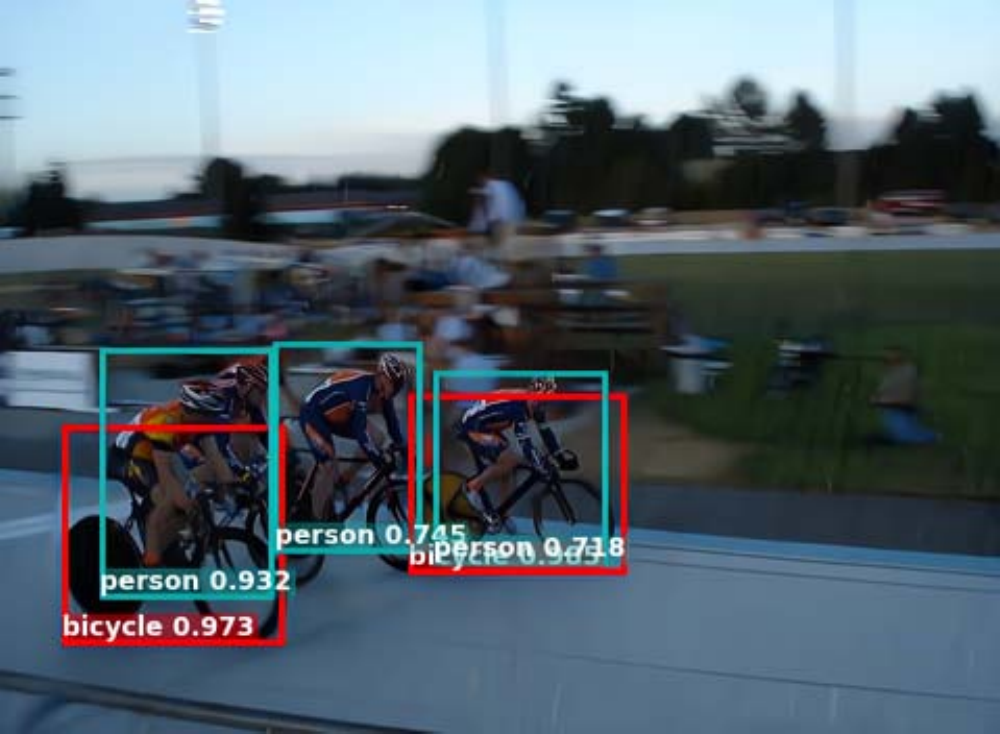} \\
        \small (a) Input & \small (b) VMambaIR & \small (c) MambaIR & \small (d) FreqMamba & \small (e) FourierMamba & \small (f) Ours  \\
    \end{tabular}
    \end{adjustbox}
    \caption{Visualization results of object detection in low-light and rainy scenes.}
    \label{fig:downstream}\vspace{-5mm}
\end{figure*}

\subsection{Effectiveness of GPS-SS2D Adaptive Scanning}

In this experiment, we investigate the contribution of the GPS-SS2D mechanism to the overall performance of the model. By comparing the full model that incorporates adaptive path planning with a variant that uses the original SS2D (without adaptive scanning), the results demonstrate that incorporating GPS-SS2D improves both restoration quality and perceptual fidelity.

As shown in Table.~\ref{tab:ss2d_ablation}, the model with GPS-SS2D adaptive scanning achieves a 0.25 dB improvement in PSNR and a 0.041 improvement in SSIM compared to the model without adaptive scanning. Moreover, the LPIPS score is reduced by 0.033, indicating better perceptual quality. These improvements highlight the effectiveness of content-aware scanning in enhancing the model's performance. In the second row of  Fig.~\ref{fig:HSV}, we visualize the HSV distributions for the input, ground truth, and the two model variants. As shown, the model with GPS-SS2D preserves the color balance better, with more consistent hue and saturation distributions, resulting in a visually more accurate restoration. This aligns with the quantitative improvements.

\subsection{Impact of ViT-based Guidance for GPS-SS2D}
To further assess the importance of the ViT-based guidance module for GPS-SS2D, we conduct an ablation study by comparing different implementations: fixed-path scanning (baseline), MLP-based guidance, CNN-based guidance, and ViT-based guidance. The results are summarized in Table.~\ref{tab:score-generator-ablation}.

The fixed variant corresponds to standard SS2D with a predetermined scanning order. Introducing MLP-based guidance improves performance by 0.74 dB in PSNR and 0.04 in SSIM, demonstrating the value of adaptive scanning. CNN-based guidance further boosts performance, indicating that stronger local feature modeling contributes to better spatial prioritization.

Our full model employing ViT-based guidance achieves the best performance, surpassing the fixed baseline by 1.71 dB in PSNR and 0.04 in SSIM. This validates the superiority of ViT in modeling global context and assigning meaningful importance scores across regions. The modest increase in computation (from 12.5G to 13.3G MACs) and parameters (from 3.2M to 3.6M) is justified by the substantial quality improvement. These results confirm that global attention derived from ViT is more effective in directing GPS-SS2D than purely local (CNN) or shallow (MLP) mechanisms.

\subsection{Comprehensive Component Analysis}
To understand the individual contributions and synergistic effects of VAMamba's core components, we systematically evaluate different combinations of LoRA, Queue-based Cache, and GPS-SS2D.

As shown in  Table.~\ref{tab:component_analysis}, individual components provide distinct contributions: LoRA alone offers minimal improvement (0.13 dB PSNR) with slight computational overhead, while GPS-SS2D alone achieves more substantial gains (0.38 dB PSNR) but introduces higher complexity (22.5 vs 15.2 GFlops).

The analysis reveals strong synergistic effects between components. The combination of LoRA and Queue-based Cache yields significant improvements (24.83 dB PSNR, 0.982 SSIM), demonstrating how parameter-efficient adaptation complements intelligent feature reuse. The Queue-based Cache + GPS-SS2D combination achieves substantial performance gains (25.11 dB PSNR) with the highest computational cost (27.1 GFlops).
The complete VAMamba model integrating all three components achieves optimal performance (25.45 dB PSNR, 0.987 SSIM, 0.162 LPIPS) while maintaining the same computational complexity as the two-component combination. Notably, adding LoRA to the Queue-based Cache + GPS-SS2D combination improves SSIM from 0.944 to 0.987 and LPIPS from 0.185 to 0.162 without additional computational cost, highlighting LoRA's efficiency in enhancing model expressiveness. These results confirm that the three components work synergistically to deliver VAMamba's superior restoration performance.

\subsection{Path Configuration Experiment}
To further validate the effectiveness of the GPS-SS2D adaptive scanning mechanism, we compared different path configurations, as shown in Table.~\ref{tab:different_path_k}. In this setting, $k=1$ denotes a single path (either forward or backward), $k=2$ indicates the combination of forward and backward paths, and $k=4$ represents repeating the two-path combination twice.
For the single-path setting ($k=1$), we observe a noticeable difference between forward and backward scans. The forward path achieves 36.40 dB PSNR and 0.984 SSIM, while the backward path performs slightly better with 36.43 dB PSNR and 0.987 SSIM. This discrepancy can be attributed to the directional bias in sequential scanning: the forward path tends to prioritize salient foreground regions but may lose global consistency when handling complex degradations (e.g., elongated textures or occlusions). In contrast, the backward path compensates better for such structural information, resulting in higher SSIM and lower perceptual error (LPIPS = 0.133).
When both paths are combined ($k=2$), the model achieves the best performance, with PSNR reaching 36.48 dB, SSIM improving to 0.988, and LPIPS decreasing to 0.115. This demonstrates the complementary nature of forward and backward paths, which jointly enhance feature interactions and mitigate the limitations of a single scanning direction.
However, increasing the path redundancy ($k=4$) leads to a slight performance drop (PSNR = 36.37 dB, LPIPS = 0.126), mainly due to redundant computations and feature overlap, which reduce the model’s efficiency and may even introduce noise.

\begin{table}[!htb]
    \centering
    \caption{Impact of Dynamic Adaptive Scanning on image motion deblurring performance, comparing the original SS2D and the dynamic adaptive scanning version. Evaluations were performed on the UHD-LL dataset.}
    \label{tab:ss2d_ablation}
    \scalebox{1.3}{
    \begin{tabular}{l|c|c|c}
        \toprule[0.15em]
        \textbf{Model Variant} & \textbf{PSNR} & \textbf{SSIM} & \textbf{LPIPS}\\
        \midrule
        Original SS2D & 26.88 & 0.887 & 0.237 \\
        w/  Adaptive Scanning (Ours) & \textbf{27.13} & \textbf{0.928} & \textbf{0.204}\\
        \bottomrule[0.15em]
    \end{tabular}}
\end{table}

\begin{table}[!htb]
\centering
\caption{Effectiveness of different guidance mechanisms for SS2D. ViT-based guidance yields the best trade-off between performance and complexity.}
\scalebox{0.95}{
\begin{tabular}{lcccc}
\toprule[0.15em]
\textbf{Method} & \textbf{PSNR (dB) ↑} & \textbf{SSIM ↑} & \textbf{MACs (G) ↓} & \textbf{Params (M)} \\
\midrule
Fixed                & 28.31 & 0.829 & \textbf{12.5} & \textbf{3.2} \\
MLP Guidance         & 29.05 & 0.847 & 12.9 & 3.4 \\
CNN Guidance         & 29.32 & 0.851 & 13.1 & 3.5 \\
ViT Guidance  (Ours) & \textbf{30.02} & \textbf{0.869} & 13.3 & 3.6 \\
\bottomrule[0.15em]
\end{tabular}}
\label{tab:score-generator-ablation}
\end{table}


\begin{table*}[!htb]
\centering
\caption{Comprehensive analysis of core components in VAMamba.}
\label{tab:component_analysis}
\setlength{\tabcolsep}{12pt}
\begin{tabular}{ccc|ccc|cc}
\toprule[0.15em]
\multicolumn{3}{c|}{\textbf{Components}} & \multicolumn{3}{c|}{\textbf{Performance}} & \multicolumn{2}{c}{\textbf{Complexity}} \\
\midrule
\textbf{LoRA} & \textbf{Queue-based Cache} & \textbf{GPS-SS2D} & \textbf{PSNR} & \textbf{SSIM} & \textbf{LPIPS} & \textbf{Params (M)} & \textbf{GFlops} \\
\midrule[0.15em]
$\times$ & $\times$ & $\times$ & 23.09 & 0.950 & 0.285 & 3.8 & 15.2 \\
$\checkmark$ & $\times$ & $\times$ & 23.22 & 0.947 & 0.278 & 4.1 & 16.8 \\
$\checkmark$ & $\checkmark$ & $\times$ & 23.47 & 0.963 & 0.245 & 5.2 & 18.3 \\
$\times$ & $\times$ & $\checkmark$ & 24.83 & 0.982 & 0.198 & 4.8 & 22.5 \\
$\times$ & $\checkmark$ & $\checkmark$ & 25.11 & 0.944 & 0.185 & 6.43 & 27.1 \\
\midrule[0.1em]
$\checkmark$ & $\checkmark$ & $\checkmark$ & \textbf{25.45} & \textbf{0.987} & \textbf{0.162} & 6.58 & 28.5 \\
\bottomrule[0.15em]
\end{tabular}
\vspace{-2mm}
\end{table*}


\begin{table}[htbp]
    \centering
    \caption{This table shows the performance of different path configurations in image restoration tasks. The combination of forward and backward paths (k=2) gives the best results, while repeating the combination (k=4) leads to a decrease in performance.}
    \renewcommand{\arraystretch}{1.2}
    \setlength{\tabcolsep}{1.em}
    \begin{tabular}{lccccc}
        \toprule[0.15em]
        \textbf{k} & \textbf{type} & \textbf{PSNR}$\uparrow$ & \textbf{SSIM}$\uparrow$ & \textbf{LPIPS}$\downarrow$ & \textbf{MAE}$\downarrow$ \\
        \midrule[0.1em]
        $k=1$ & forward   & 36.40 & 0.984 & 0.136 & 0.034 \\
        $k=1$ & backward  & 36.43 & 0.987 & 0.133 & 0.036 \\
        $k=2$ & two path  & \textbf{36.48} & \textbf{0.988} & \textbf{0.115} & \textbf{0.029} \\
        $k=4$ & two path $\times$2 & 36.37 & 0.986 & 0.126 & 0.032 \\
        \bottomrule[0.15em]
    \end{tabular}
    
    \label{tab:different_path_k}
\end{table}

\section{Additional Experiments}
In this section, we present additional experiments to further evaluate the generalization and versatility of our model. Specifically, we focus on two important aspects: performance on real-world low-light enhancement and validation of our method on downstream tasks, such as object detection.

\subsection{Real-World Low-Light Enhancement}
To assess the practical applicability of our method, we conduct experiments on unpaired real-world low-light datasets. 
Unlike synthetic datasets that are carefully curated, real-world images come with diverse lighting conditions, complex noise patterns, and variations in scene composition.
As shown in Fig.~\ref{fig:realworld}, our method outperforms state-of-the-art techniques, providing clearer details in the shadows and more natural color preservation.

\subsection{Computer Vision Applications}
In addition to enhancing image restoration quality, we also assess the impact of our method on downstream tasks, such as object detection. To validate the effectiveness of our restored images in practical applications, we perform object detection experiments on both low-light and raining scenarios. The visual comparisons of object detection results, using images enhanced by our model, are shown in Fig.~\ref{fig:downstream}. These results demonstrate that our method not only improves image quality but also enhances the performance of object detection.

\section{Conclusion}
We propose VAMamba, a visual Adaptive Mamba network for image restoration that addresses the limitations of fixed scanning patterns in existing state space models. Our approach introduces two key innovations: GPS-SS2D, which enables content-aware adaptive scanning through ViT-generated importance scores, and QCLAM, which combines queue-based feature caching with LoRA adaptation for efficient memory usage and parameter learning.
Extensive experiments across five image restoration tasks demonstrate that VAMamba consistently outperforms state-of-the-art methods in both restoration quality and computational efficiency. The comprehensive ablation studies confirm the synergistic effects of our proposed components. VAMamba represents a significant advancement in efficient image restoration, offering a practical solution that sets new benchmarks across diverse restoration scenarios.

\printcredits

\bibliographystyle{cas-model2-names}

\bibliography{adaptivemamba}


\bio{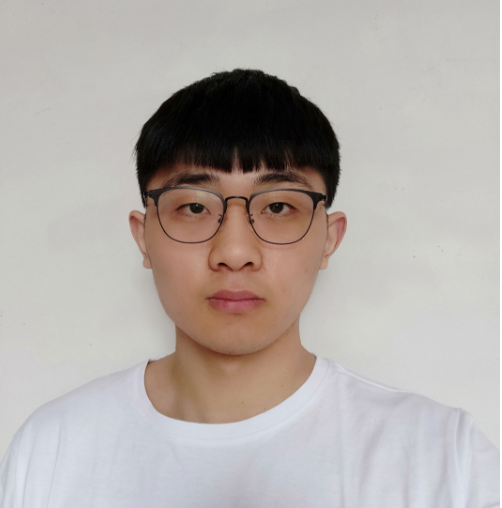}
{Han Hu} is a master's student in Software Engineering at Shandong Normal University, enrolled in 2023. His research interests include computer vision.
\endbio
\vskip5pc

\bio{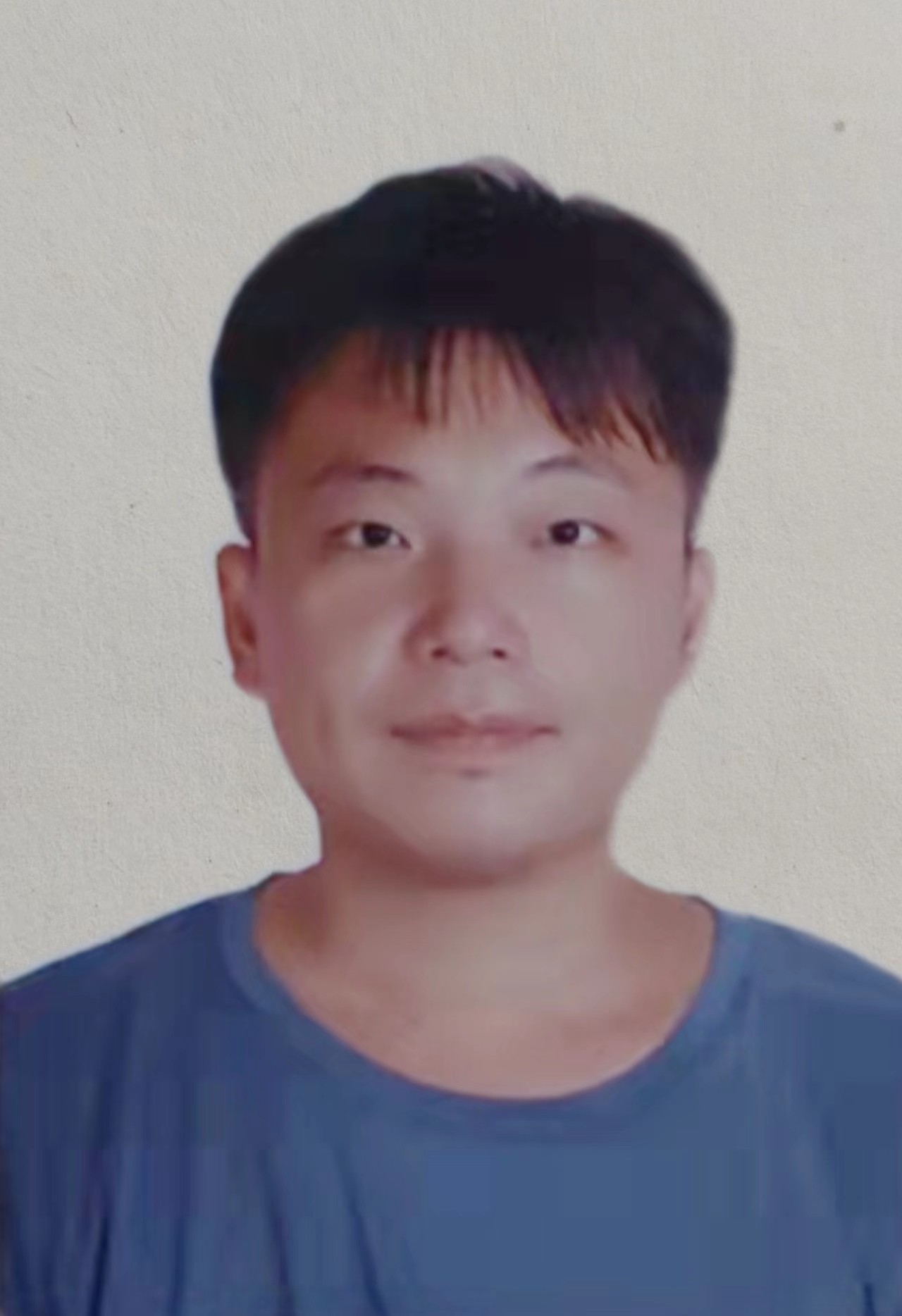}
{Zhuoran Zheng} received his Ph.D. from Nanjing University of Science and Technology(2019-2023).Since 2023, he has been working as a Postdoctora Researcher at Sun Yat-sen University. His primary research focuses on image processing, with particular emphasis on image enhancement,restoration, and reconstruction. Dr. Zheng has published multiple papers in top-tier computer vision conferences including CVPR, ICCV, and ECCV.
\endbio

\vskip5pc

\bio{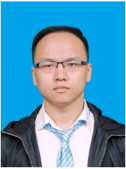}
{Liang Li} received the Ph.D. degree in computer science and technology from the School of Information Science and Engineering, Shandong Normal University, Jinan, China, in 2020. His research interests include computer simulation, crowd dynamics models, and evacuation management.
\endbio

\vskip5pc

\bio{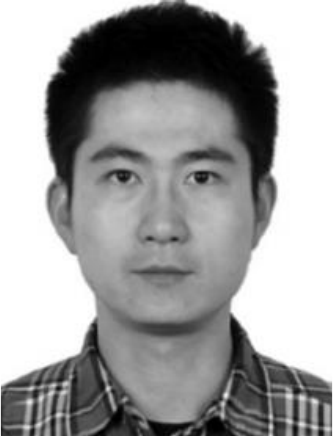}
{Chen Lyu} received his Ph.D. degree from the Institute of Computing Technology, Chinese Academy of Sciences, Beijing, China, in 2015. He is currently an associate professor with the School of Information Science and Engineering, Shandong Normal University, Jinan, China. His research interests include computer vision, multimedia information processing and artificial intelligence.
\endbio

\end{document}